\documentclass[lettersize,journal]{IEEEtran}
\usepackage{amsmath,amsfonts}
\usepackage{algorithmic}
\usepackage{algorithm}
\usepackage{array}
\usepackage[caption=false,font=normalsize,labelfont=sf,textfont=sf]{subfig}
\usepackage{textcomp}
\usepackage{stfloats}
\usepackage{url}
\usepackage{verbatim}
\usepackage{graphicx}
\usepackage{cite}
\usepackage{CJKutf8}
\usepackage{color}
\usepackage{booktabs}
\usepackage{hyperref}
\usepackage{pdflscape}
\usepackage{longtable}
\usepackage{makecell}
\usepackage{multirow}
\usepackage{xurl}

\newcommand{\etal}{\textit{et al}.}

\begin{document}


\bstctlcite{IEEEexample:BSTcontrol}

\title{Radar-Camera Fusion for Object Detection and Semantic Segmentation in Autonomous Driving: \\A Comprehensive Review}

\author{Shanliang Yao$^{\text{1}}$, Runwei Guan$^{\text{1}}$, Xiaoyu Huang$^{\text{1}}$, Zhuoxiao Li$^{\text{1}}$, Xiangyu Sha$^{\text{1}}$, \\Yong Yue$^{\text{2}}$, Eng Gee Lim$^{\text{2}}$, Senior Member, IEEE, Hyungjoon Seo$^{\text{1}}$, Ka Lok Man$^{\text{2}}$, Xiaohui Zhu$^{\text{2}, \dagger}$, Yutao Yue$^{\text{3}, \dagger}$

\thanks{$^{\text{1}}$ Shanliang Yao, Runwei Guan, Xiaoyu Huang, Zhuoxiao Li, Xiangyu Sha and Hyungjoon Seo are with Faculty of Science and Engineering, University of Liverpool, Liverpool, UK. (email: \{shanliang.yao, runwei.guan, x.huang42, zhuoxiao.li, sgxsha2, hyungjoon.seo\}@liverpool.ac.uk).}
\thanks{$^{\text{2}}$ Yong Yue, Eng Gee Lim, Ka Lok Man and Xiaohui Zhu are with School of Advanced Technology, Xi'an Jiaotong-Liverpool University, Suzhou, China. (email: \{yong.yue, enggee.lim, ka.man, xiaohui.zhu\}@xjtlu.edu.cn).}
\thanks{$^{\text{3}}$ Yutao Yue is with Institute of Deep Perception Technology, JITRI, Wuxi, China; XJTLU-JITRI Academy of Industrial Technology, Xi'an Jiaotong-Liverpool University, Suzhou, China; and Department of Mathematical Sciences, University of Liverpool, Liverpool, UK. (email: yueyutao@idpt.org).}
\thanks{$^{\dagger}$ Corresponding author: xiaohui.zhu@xjtlu.edu.cn, yueyutao@idpt.org}
}


\maketitle

\begin{abstract}
Driven by deep learning techniques, perception technology in autonomous driving has developed rapidly in recent years, enabling vehicles to accurately detect and interpret surrounding environment for safe and efficient navigation. To achieve accurate and robust perception capabilities, autonomous vehicles are often equipped with multiple sensors, making sensor fusion a crucial part of the perception system.
Among these fused sensors, radars and cameras enable a complementary and cost-effective perception of the surrounding environment regardless of lighting and weather conditions.
This review aims to provide a comprehensive guideline for radar-camera fusion, particularly concentrating on perception tasks related to object detection and semantic segmentation.
Based on the principles of the radar and camera sensors, we delve into the data processing process and  representations, followed by an in-depth analysis and summary of radar-camera fusion datasets.
In the review of methodologies in radar-camera fusion, we address interrogative questions, including ``why to fuse'', ``what to fuse'', ``where to fuse'', ``when to fuse'', and ``how to fuse'', subsequently discussing various challenges and potential research directions within this domain. To ease the retrieval and comparison of datasets and fusion methods, we also provide an interactive website: \url{https://radar-camera-fusion.github.io}.
\end{abstract}


\section{Introduction} \label{Introduction}

\IEEEPARstart{A}{utonomous} driving has excellent potential in mitigating traffic congestion and improving driving safety. Perception, akin to eyes in autonomous driving, constitutes the foundation for successive functions, such as motion prediction, path planning and maneuver control \cite{paden2016survey, xiao2020multimodal}. To achieve optimal accuracy and robustness of the perception system, various sensors are integrated into autonomous vehicles, allowing for the utilization of their complementary and redundant characteristics \cite{feng2020deep, liang2019multi}. However, which sensors to choose and how to fuse the data between different sensors have emerged as challenging issues requiring further exploration.

\begin{figure}[!t]
\begin{center}
\includegraphics[width=1\linewidth]{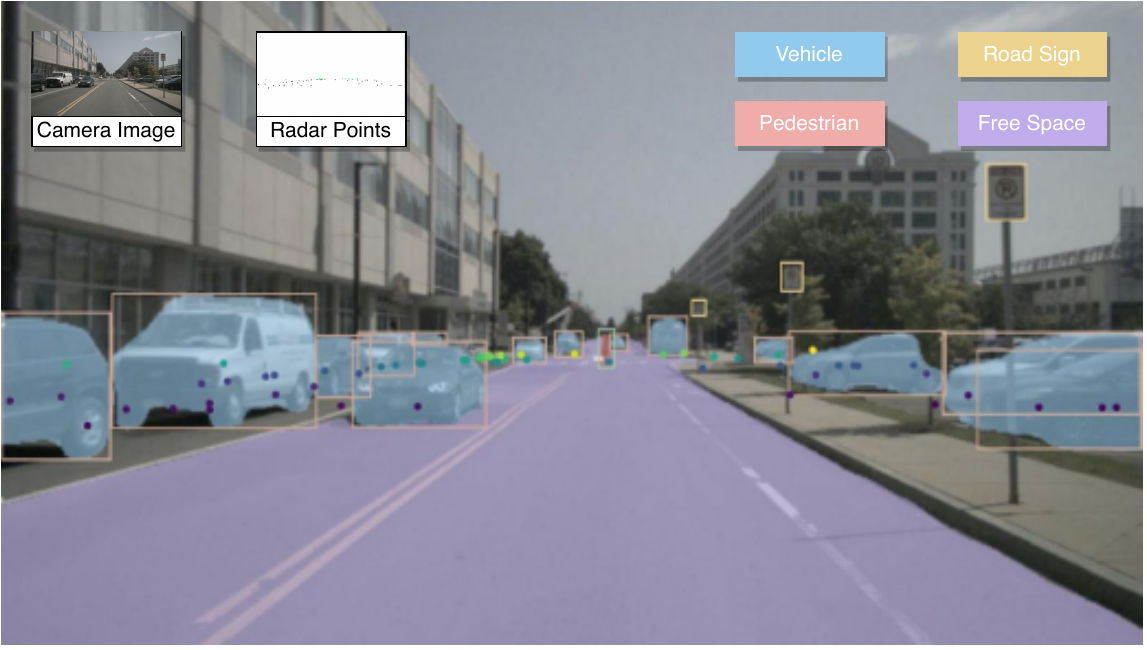}
\end{center}
\caption{Common scenario of object detection and semantic segmentation in autonomous driving. Boxes and masks represent the results of detection and segmentation, respectively. Dots indicate the location of each radar point, and the darker the dot, the closer the distance to the ego-vehicle. Image is generated from the nuScenes \cite{caesar2020nuscenes} dataset.}
\label{fig:cover}
\end{figure}

Given the rich semantic information that can be perceived, cameras are widely utilized in autonomous driving for object detection, segmentation and tracking. LiDARs calculate the distance to surrounding objects by measuring the time difference of the laser beam from emission to reception via the objects. The denser the laser layers emitted by a LiDAR sensor, the clearer an object's three-dimensional (3D) contour. 
These complementary features provided by cameras and LiDARs have made LiDAR-camera sensor fusion a hot topic in recent years, and achieved high accuracy in two-dimensional (2D) and 3D object detection \cite{chen2017multi, ku2018joint, li2016vehicle, nabati2019rrpn}, semantic segmentation \cite{zhang2015sensor, yin2020fusionlane} and object tracking \cite{asvadi20163d, fang2019camera}.
Despite their strengths, both LiDARs and cameras suffer from the same defect of being sensitive to adverse weather conditions (e.g., rain, fog, snow) that can significantly diminish their field of view and object recognition capabilities \cite{yoneda2019automated}.
Moreover, the high cost of LiDAR products has brought certain difficulties in promoting their widespread adoption \cite{li2022exploiting}.

Compared to LiDARs and cameras, radars exhibit superior effectiveness under challenging lighting and weather conditions \cite{major2019vehicle, wang2021rodnet}. Radars can also deliver accurate velocity estimation for all detected objects depending on the Doppler effect without requiring any temporal information \cite{nabati2021centerfusion}. With these characteristics, radars are widely used in Advanced Driving Assistance Systems (ADAS) applications, including collision avoidance, Adaptive Cruise Control (ACC), Lane Change Assist (LCA) and Automatic Emergency Braking (AEB). 
As depicted in Figure \ref{fig:cover}, the integration of radar and camera data in sensor fusion enables a comprehensive perception of the surrounding environment in terms of outlines, colors, textures, ranges, and velocities. Moreover, the fusion system can operate continuously throughout the day regardless of weather and lighting conditions.

Although radar sensors are popularly applied to vehicles, few studies focus on data fusion from radars and cameras. One reason for this is the limitations of radar output data, such as low resolution, sparse point clouds, uncertainty in elevation and clutter effects. Another reason is that up to now, the datasets containing both radar and camera data for autonomous driving applications are insufficient, making it challenging for researchers to conduct in-depth analysis.
Additionally, applying or adapting existing LiDAR-based algorithms to radar point clouds yields poor results due to inherent differences of point clouds between the LiDAR sensor and radar sensor \cite{nabati2021centerfusion}. Radar point clouds are significantly sparser than their LiDAR counterparts, making it inefficient to extract objects' geometry information using LiDAR-based algorithms. 
Although Radar Cross Section (RCS) values in the radar sensor indicate the reflective intensity from the surface of an object, they are easily affected by numerous factors and cannot be used singularly to determine the classification of the target. 
In addition, though aggregating multiple radar frames enhances the density of the point clouds, it also causes a delay to the whole system. 
In summary, radar-camera fusion perception is significant in autonomous driving as well as challenging in implementation.
   
\subsection{Related Surveys}

Most sensor fusion surveys focus on LiDAR-camera \cite{arnold2019survey, cui2021deep, xiao2020multimodal, huang2022multi}, or the broader field of multi-sensor fusion, including LiDAR, camera, radar and other sensors \cite{feng2020deep, huang2022multi, jusoh2020systematic, wang2019multi}. 
Specifically, in multi-sensor fusion surveys, LiDARs and cameras are still the main research objectives. For example, Feng~\etal~\cite{feng2020deep} conducted a comprehensive survey on deep multi-modal object detection and semantic segmentation for autonomous driving. However, this survey mainly concentrates on fusion methods based on LiDARs and cameras, and briefly mentions some studies combining camera images and radar data.
 
 To the best of our knowledge, \cite{wei2022mmwave} is the only survey that primarily focuses on radar-camera fusion for object detection in autonomous driving. However, it does not cover the radar-camera fusion dataset or the semantic segmentation task.

\subsection{Contributions}


With the limited focus on radar-camera fusion in existing surveys, it is challenging for researchers to gain an overview of this emerging research field. Our survey attempts to narrow this gap by providing a comprehensive review of radar-camera fusion in autonomous driving. 
The contributions of our review are summarized as follows:
\begin{itemize}
\item To the best of our knowledge, this is the first survey focusing on two fundamental perception problems for radar-camera fusion, namely, object detection and semantic segmentation.
\item We present an up-to-date (2019 - 2023) overview of radar-camera fusion datasets and algorithms, and conduct in-depth research on ``why to fuse", ``what to fuse", ``where to fuse", ``when to fuse", and ``how to fuse". 
\item We analyze the critical challenges and open questions in radar-camera fusion, and put forward potential research directions.
\item We provide an interactive and updated website for better retrieving and comparing the fusion datasets and methods.
\end{itemize}

\section{Background} \label{Background}
This section provides background information on radar-camera fusion in autonomous driving.
We first introduce the working principles, sensor characteristics and data representations of the radar and camera sensors. By comparing the characteristics of the two sensors, we aim to demonstrate the importance of radar-camera fusion.
Subsequently, as the perception module leverages data from specific sensors to understand the surroundings, we present basic concepts and highlight representative algorithms for two fundamental and crucial perception tasks: object detection and semantic segmentation.

\subsection{Radar Sensors}

\subsubsection{Working Principles}
Radar is the abbreviation of Radio Detection And Ranging, which calculates the range and velocity of the target by transmitting radio waves and receiving the reflected waves from the target \cite{iovescu2017fundamentals}. 
In autonomous driving applications, radar typically refers to the MilliMeter-Wave (MMW) radar that works in the millimeter wave band with a wavelength of 1-10mm and frequency of 76-81GHz.
Specifically, the radar equipped in the forward direction, as well as the four corner directions, is usually a Multiple-Input Multiple-Output (MIMO) radar, while the radar on the roof is typically a mechanical rotating radar.
A MIMO radar utilizes multiple antennas and transmitters to simultaneously transmit and receive multiple signals with different frequencies. In contrast, a mechanical rotating radar operates with a single antenna that physically rotates to emit radar signals in different directions. 
With multiple antennas and beamforming capabilities, a MIMO radar achieves higher spatial resolution and interference reduction compared to mechanical rotating radar. While a mechanical rotating radar provides better coverage and is simpler in implementation. 

Based on the Time of Flight (TOF) principle, the radar sensor calculates the range from the object by the time difference between the transmitted and reflected signals. Based on the Doppler principle, when there is a relative movement between the emitted electromagnetic wave and the detected target, the frequency of the returned wave differs from that of the emitted wave. Thus, the target's relative velocity to the radar can be measured using this frequency difference. 
Leveraging the array signal processing method, the azimuth angle is calculated using the signal's phase difference between parallel antennas. Since the receivers of traditional 3D (range, Doppler velocity and azimuth angle) radar sensors are only lined up in a 2D direction, targets are only detected in 2D horizontal coordinates without vertical height information. Recently, with advancements in radar technologies, 4D (range, Doppler velocity, azimuth angle and elevation angle) radar sensors have been developed with antennas arranged horizontally and vertically, enabling the measurement of elevation information. In addition, 4D is often represented as x, y, z coordinates and Doppler velocity.



\subsubsection{Sensor Characteristics}
In addition to the ability to measure range, Doppler velocity and azimuth angle, electromagnetic waves in the millimeter wave band have low atmospheric attenuation and better penetration of rain, smoke and dust \cite{appleby2007millimeter}. These characteristics make the radar sensor work all day regardless of severe weather conditions.
However, radar sensors still have certain limitations. They exhibit low angular resolution and cannot distinguish between closely located objects. The point clouds generated by radars are sparsely distributed, with only a few points on a pedestrian and a dozen points on a car. These points cannot adequately outline an object's contours, making it challenging to extract the geometric information \cite{fritsche2016radar, john2019rvnet}. 
Doppler radar measurements have a limitation in that they only provide the radial component of velocity. The lack of tangential velocity makes it difficult to estimate the accurate velocity of an object in dynamic scenes \cite{long2021full, li2022mathsf}.
Besides, data produced by radars are noisy, which may arise from diverse sources such as multipath interference, electrical interference and equipment imperfections \cite{griffiths2014radar, kopp2023tackling}. Such noise reduces the precision and reliability of radar data, while also  increasing the probability of false detections.
Furthermore, radars are weak in the perception of stationary obstacles. Moving targets can be distinguished from the surrounding scene in one dimension of range and velocity. However, radars are highly sensitive to metal, often resulting in strong reflections from stationary objects such as manhole covers on the ground. Strong reflections from stationary objects are not filtered, resulting in a lack of detecting stationary obstacles.

\subsubsection{Data Representations}
\label{Radar Data Representations}

\begin{figure}[ht]
\begin{center}
   \includegraphics[width=1\linewidth]{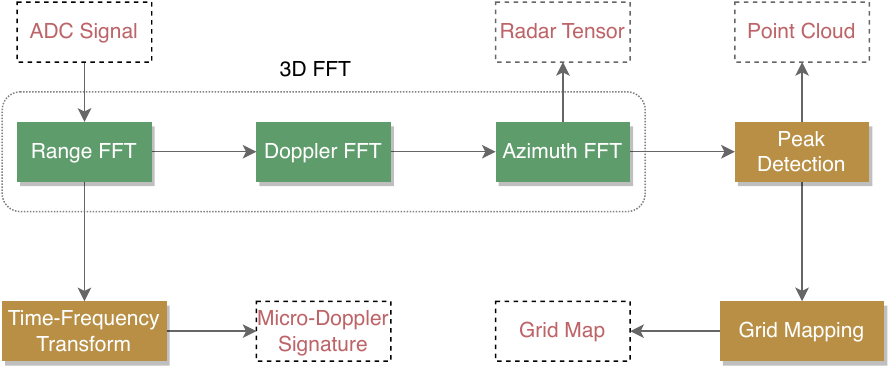}
\end{center}
   \caption{Generation process of radar data representations.}
\label{fig:radar-signal-processing-chain}
\end{figure}

\begin{figure}[ht]
\begin{center}
   \includegraphics[width=1\linewidth]{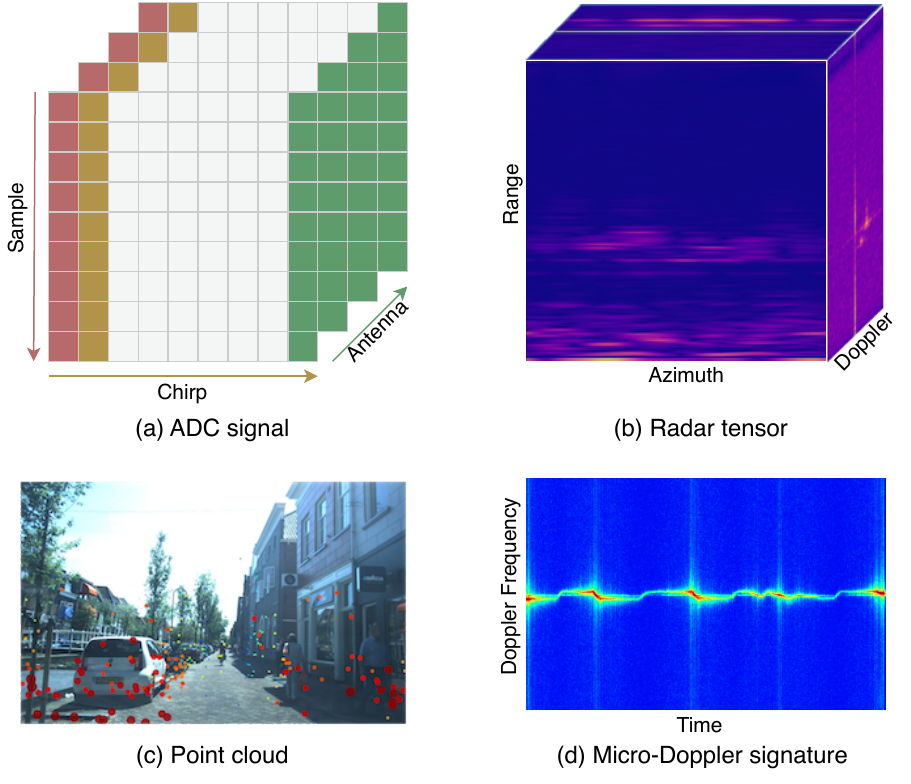}
\end{center}
   \caption{Radar data representations. (a) ADC signal in the format of Simple-Chirp-Antenna tensor. (b) Radar tensor represented by a 3D Range-Azimuth-Doppler tensor. Image is generated from the CARRADA \cite{ouaknine2021carrada} dataset. (c) Point cloud projected on a 2D image plane. Image is generated from the View-of-Delft \cite{palffy2022multi} dataset. (d) Micro-Doppler signature showing a pedestrian walking. Image is generated from the Open Radar Datasets \cite{gusland2021open}.}
\label{fig:radar-representations}
\end{figure}

As depicted in Fig. \ref{fig:radar-signal-processing-chain}, the raw output of a radar sensor is the \textbf{ADC signal}, which refers to the output signal of an Analog-to-Digital Converter (ADC).
At this stage, the signal lacks spatial coherence between the values as all the information exists in the time domain \cite{nowruzi2020deep}. Besides, the signal is represented in complex value which contains real part and imaginary part \cite{yang2023adcnet}.
To represent the ADC signal in a more structured form, it is usually transformed into a 3D Sample-Chirp-Antenna (SCA) tensor, as illustrated in Fig. \ref{fig:radar-representations}(a).
Some researchers apply 3D Fast Fourier Transform (FFT) along the sample, chirp and antenna dimensions to get an image-like representation named \textbf{radar tensor} (Fig. \ref{fig:radar-representations}(b)), describing the spatial pattern of the received echo \cite{major2019vehicle, wang2021rodnet}. 
At this stage, the non-coherent combination (e.g., norm calculation) converts ADC signals composed of complex values to radar tensors that consists of real values.
With these three features in Range-Azimuth-Doppler (RAD) coordinates, two forms of radar tensors are formed: one is in 2D including the Range-Azimuth (RA) tensor, Range-Doppler (RD) tensor and Azimuth-Doppler (AD) tensor; the other is the whole 3D RAD tensor, with each side consisting of a 2D tensor. 
Furthermore, peak detection is carried out on the radar tensor to filter out clutter, resulting in a sparse point-like representation called the \textbf{point cloud}, as depicted in Fig. \ref{fig:radar-representations}(a). The point cloud provides a spatially intuitive representation better suited for visualization and interpretation, yet it can not accurately indicate the outline information \cite{schumann2018semantic, fent2023radargnn}. Constant False Alarm Rate (CFAR) is the most commonly used method for peak detection, which enables the radar system to automatically adjust its sensitivity level to changes in the strength of external interference, thereby maintaining a steady false alarm rate \cite{rohling1983radar, gandhi1988analysis}.
By applying grid mapping methods to point clouds accumulated over a given period, a \textbf{grid map} for identifying static objects is generated. There are two main grid maps: one is the occupancy-based grid map \cite{werber2015automotive, sless2019road, schumann2019scene}, which represents the obstacles and free-space derived from the radar data; the other is the amplitude-based grid map \cite{werber2015automotive, lombacher2015detection}, which displays the RCS values for each cell. However, it is essential to note that the sparsity of the point clouds still influences the accuracy of detection and segmentation performed on the grid map. 
In addition, some researchers perform Time-Frequency transform after the Range-FFT to obtain the \textbf{micro-Doppler signature}, which is utilized to recognize objects with tiny motion features \cite{gusland2021open, chen2022attention}. As exemplified in Fig. \ref{fig:radar-representations}(d), the Doppler frequency of a pedestrian walking shows a periodic variation. This representation enables not only the distinction of different object categories (e.g., pedestrians, bicycles and vehicles), but also the recognition of complex object behaviors, such as gait and gesture recognition.

\subsection{Camera Sensors}


\subsubsection{Working Principles}
The camera sensor usually consists of a lens, an image sensor, an Image Signal Processor (ISP) and an Input/Output (I/O) interface \cite{enwiki:1146373856}. The lens collects the light reflected from the target and converges it to the image sensor. Then, the image sensor converts light waves into electrical signals and converts electrical signals to digital values via an on-chip ADC. After that, the ISP performs post-processing (e.g., noise reduction) and converts the digital values into a format of RGB data for images or videos. Finally, the image data is transferred and displayed via the I/O interface.


\subsubsection{Sensor Characteristics}
Cameras capture the rich appearance features of the objects, including colors, shapes and textures. After learning from neural networks, these features can be utilized to identify obstacles, including vehicles, pedestrians, bicycles and various traffic lights.
However, cameras are passive sensors, indicating that the formation of an image requires incident light intake. When the light intake is adversely affected, such as insufficient light at night, extreme weather, water droplets or dust sticking to the lens, the imaging results will be unclear, and object detection performance may be significantly affected \cite{michaelis2019benchmarking}. Besides, in autonomous driving, it is crucial to identify the distances of obstacles ahead. However, a target in three dimensions in the world coordinate system becomes a 2D target in the image coordinate system after being imaged by the camera sensor, resulting in a loss of distance information.

\subsubsection{Data Representations}
\textbf{Raw data} representation is the uncompressed and unprocessed format captured by the camera sensor. It contains all the radiance information that hits each pixel on the camera sensor during image exposure \cite{xu2019towards,liu2021exploit}. 
After post-processing, a data representation named \textbf{RGB image} is generated, which illustrates an image as a grid of pixels, with each pixel containing a value for each of the red, green and blue color channels. 
In addition, some modern cameras applied in autonomous driving are able to generate specific data representations. For instance, a depth camera produces the \textbf{depth map} representation, providing information about the distance to each pixel in the scene \cite{song2016robust}.
Relying on flood-light flash laser sources, infrared cameras output the \textbf{infrared image} representation, which is able to render perception results in adverse weather and low-light conditions \cite{inbar2008laser, bijelic2020seeing}.
Event cameras are bio-inspired vision sensors that generate an \textbf{event image} representation pertaining to pixel-level changes in brightness. With sub-millisecond latency, high-dynamic range, and robustness to motion blur, event cameras present considerable potential for real-time detection and tracking of objects in time-critical scenarios \cite{gallego2020event}.

\subsection{Comparison of Radar and Camera Sensors}
\label{sec:Comparison of Radar and Camera}

\begin{figure}[h]
\begin{center}
   \includegraphics[width=1\linewidth]{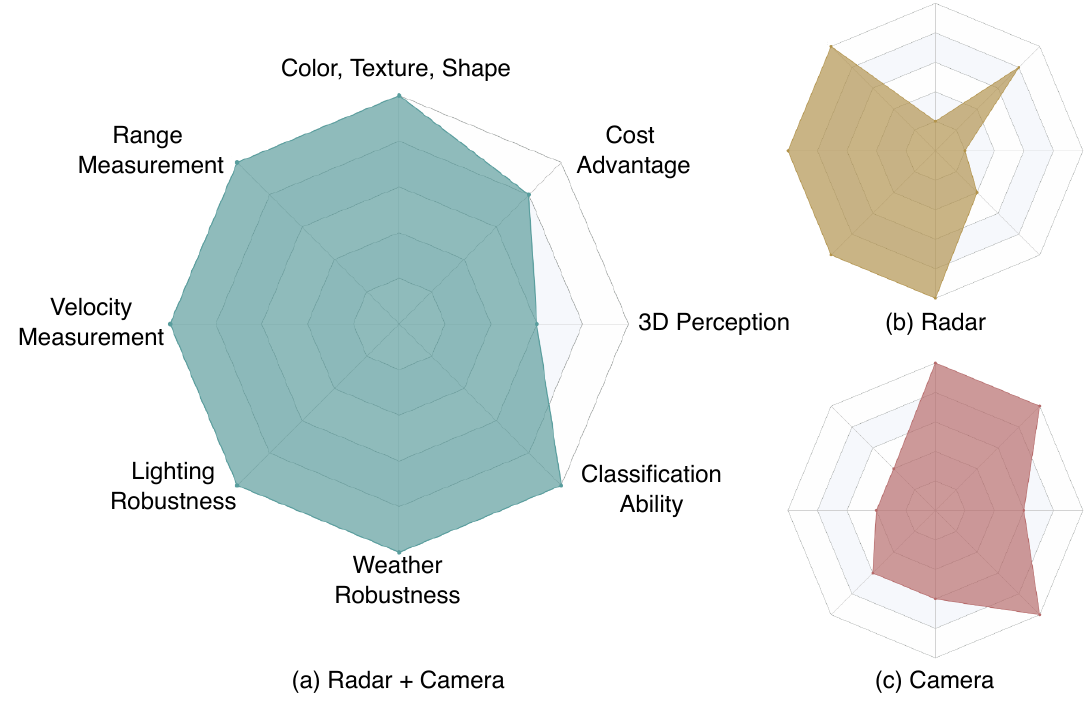}
\end{center}
\caption{Comparison of radar and camera characteristics. In these charts, each characteristic is plotted along one of the line segments radiating from the central point, with closer proximity to the vertex representing higher quality for that characteristic. (a) Fusion characteristics of radar and camera. (b) Radar characteristics. (c) Camera characteristics. }
\label{fig:Comparison of characteristics of radar and camera}
\end{figure}

Through our extensive review, a clear and concise figure is designed to compare the characteristics of the two sensors, shown in Fig. \ref{fig:Comparison of characteristics of radar and camera}. Specifically, the radar sensor is an active sensor and measures various information, including ranges, velocities and azimuth angles \cite{rosique2019systematic}. 
Nowadays, radars equipped in driver-assistance systems can detect up to 300 meters, with a 140$^{\circ}$ horizontal field of view and a less than 1$^{\circ}$ angular resolution \cite{saponara2017radar, sun2020mimo}.
In addition, the radar sensor is robust to darkness and extreme weather conditions, allowing it to work throughout the day. 
The camera sensor is a passive sensor that provides colors, textures and shapes of objects. With a resolution of up to 2K, the camera sensor performs much better in classification than the radar sensor. 
As far as system cost goes, both radars and cameras are relatively cost-effective and are mass-applied in vehicles.

To sum up, both the radar and camera have their strengths and weaknesses, and they cannot be substituted for each other. The most effective way to ensure adequate information acquisition is mutual integration. Based on their respective characteristics, complementary advantages can improve scenario understanding performance. In addition, when one of the sensors fails, the remaining one can continue working, thus increasing the reliability of the autonomous driving system. Hence, the fusion of radar and camera sensors is critical for perception accuracy and robustness in autonomous driving.


\subsection{Perception Tasks}

\subsubsection{Object Detection}

The object detection task involves identifying a particular object in a camera image or a radar scan, locating its position and determining its category. Generally speaking, researchers use a rectangular or cubic bounding box to encompass the object. As there is no depth channel in 2D object detection, the rectangular bounding box is expressed as $(x, y, h, w, c)$, where $(x, y)$ is the bounding box center, $h$ and $w$ are the height and width of the bounding box, and $c$ is the class of the object. While the cubic bounding box for 3D object detection is described as $(x, y, z, h, w, l, \theta, c)$, where $(x, y, z)$ represents the center of the 3D bounding box, $h$, $w$ and $l$ are the height, width and length of the bounding box, $\theta$ is the object's orientation, and $c$ is its class. Bird's Eye View (BEV) object detection is a specialized form of 3D object detection focusing on detecting objects from a top-down perspective. In this approach, height information is typically discarded, and objects are represented as 2D bounding boxes on the ground plane.

\paragraph{Camera-based Object Detection}
In autonomous driving, camera-based object detection approaches have been widely used in detecting vehicles \cite{li20173d, chen2016monocular}, pedestrians \cite{zhang2017towards, zhang2016faster, chen20173d}, traffic lights \cite{weber2016deeptlr, muller2018detecting, bach2017multi, behrendt2017deep}, and traffic signs \cite{zhu2016traffic, lee2018simultaneous, luo2017traffic}. 
According to different training steps, CNN-based object detection algorithms can be classified into two-stage and one-stage. Two-stage detection algorithms (e.g., R-CNN \cite{girshick2014rich}, SPPNet \cite{he2015spatial}, Fast R-CNN \cite{girshick2015fast}, Faster R-CNN \cite{ren2015faster}, FPN \cite{lin2017feature}) segregate the detection problem into two stages: the first step is to generate region proposals, and the second step is to refine the position and predict the classification of each object. Experimental results of these algorithms are high in precision and recall, but relatively slow in time.
Without the region proposal generation phase, one-stage detection algorithms simultaneously predict the bounding boxes and the probability of classes within these boxes. Thus, one-stage detection algorithms are commonly faster than two-stage detection algorithms, but lower in accuracy. Some highly representative one-stage object detectors include the YOLO series \cite{redmon2016you, redmon2017yolo9000, redmon2018yolov3, bochkovskiy2020yolov4, Jocher2020YOLOv5, li2022yolov6, wang2023yolov7, Jocher_YOLO_by_Ultralytics_2023}, SSD \cite{liu2016ssd} and RetinaNet \cite{lin2017focal}.


Exploiting the self-attention mechanism that enables the model to model the contextual features and their correlation, transformer-based methods have emerged as a recent breakthrough compared to CNN-based detectors. Some representative pure transformer detectors include DETR \cite{carion2020end}, Deformable DETR \cite{zhu2020deformable}, RT-DETR \cite{lv2023detrs}, WB-DETR \cite{liu2021wb}, Swin \cite{liu2021swin} and YOLOS \cite{fang2021you}.
Additionally, plenty of studies have endeavored to accelerate the conventional transformer block by the combination of convolution and self-attention, aggregating the advantages of both CNN and transformer, as exemplified by Conformer \cite{peng2021conformer}, EdgeViTs \cite{pan2022edgevits}, MobileViT \cite{mehta2021mobilevit}, ViTAE \cite{xu2021vitae} and Visformer \cite{chen2021visformer}.
 
\paragraph{Radar-based Object Detection}

Radar-based object detection approaches have been widely used in detecting vehicles \cite{major2019vehicle, wang2021rodnet}, pedestrians \cite{palffy2020cnn, zhang2021raddet} and static objects \cite{svenningsson2021radar}. 
As radar tensors are image-like representations, researchers generally utilize image-based networks (e.g., ResNet \cite{he2016deep}, Faster R-CNN \cite{ren2015faster}, YOLOv4 \cite{bochkovskiy2020yolov4}) to perform object detection on 2D RA tensors \cite{gao2019experiments, dong2020probabilistic, wang2021rodnet}, 2D RD tensors \cite{ng2020range, decourt2022darod} and 3D RAD tensors \cite{major2019vehicle, gao2020ramp, palffy2020cnn, zhang2021raddet, rebut2022raw}. Unlike images, radar tensors lack a physical interpretation, thereby presenting difficulties in translating the learned features from image-based algorithms to radar data. Furthermore, applying algorithms to radar tensors in real-time applications poses challenges due to the high-dimensional nature of radar tensors and the presence of noise, interference and clutter.

For radar data in the format of point clouds, various types of point-based networks are utilized to detect objects.
Point-wise methods \cite{danzer20192d, tilly2020detection, scheiner2020seeing, dubey2022haradnet, tan20223d} directly operate on the raw point clouds and leverage LiDAR-based algorithms, such as PointNet \cite{qi2017pointnet}, PointNet++ \cite{qi2017pointnet++} and Frustum PointNets \cite{qi2018frustum}, to classify the points into distinct object classes. 
Grid-based methods \cite{dreher2020radar, scheiner2020seeing, kohler2023improved, liu2023smurf} map the 3D point clouds into grid-like structures, such as 2D image planes or 3D voxel grids. Subsequently, object detection algorithms (e.g., YOLOv3 \cite{redmon2018yolov3}, VoxelNet \cite{zhou2018voxelnet}) are applied to the grid representation to identify objects. The grid-based approaches demonstrate efficiency in handling large datasets and are frequently employed in real-time applications.
Graph-based methods (e.g., RadarGNN \cite{fent2023radargnn}, Radar-PointGNN \cite{svenningsson2021radar}) in radar point cloud object detection employ the Graph Neural Network (GNN), where the points serve as nodes, and the relationships between the points are modeled as edges in the graph. Leveraging graph structures and algorithms, these methods effectively capture the spatial relationships and contextual information among the points, leading to improved detection performance compared to traditional point-wise methods. However, the construction and feature extraction of graphs are complex and computationally intensive, especially when handling large-scale point clouds.

%
%



\subsubsection{Semantic Segmentation}

Semantic segmentation involves clustering the basic components of input data into different semantically relevant regions.
Essentially, it refers to assigning selected labels from a pre-defined set $Y = \{y_1, y_2, ..., y_k \}$ to each pixel in an image-based dataset $D_i = \{d_1, d_2, ..., d_m\}$ or each point in a point-based dataset $D_p = \{d_1, d_2, ..., d_n \}$.

\paragraph{Camera-based Semantic Segmentation}
The technique of camera-based semantic segmentation finds widespread application in the fields of free-space segmentation \cite{tsutsui2017distantly, tsutsui2018minimizing, chan2019lane, qian2019dlt, wang2021sne, shao2021semantic}, lane segmentation \cite{kim2017end, neven2018towards, wang2018lanenet, zou2019robust,chan2019lane,qian2019dlt, shao2021semantic} and obstacle segmentation \cite{chen2021wodis, bovcon2021wasr, bovcon2021mods} in autonomous driving.
Fully Convolutional Network (FCN) \cite{long2015fully} is a milestone in semantic segmentation as it enables end-to-end training of deep networks for this task. However, due to its failure to account for global contextual information, the obtained segmentation results tend to be coarse. Therefore, the encoder-decoder architecture has emerged to address this shortcoming, represented by SegNet \cite{badrinarayanan2017segnet}, U-Net \cite{ronneberger2015u} and HRNet \cite{yuan2020object}. The encoder-decoder architecture typically uses an image classification network as its encoder, gradually reducing the spatial dimensions of the pooling layer. Meanwhile, the decoder gradually restores the details and spatial dimensions for segmentation purposes.

However, for encoder-decoder architectures, high-resolution representations are lost during the encoding process, reducing fine-grained information within the image.
Dilated (or ``atrous") convolution structure is created to avoid decimating the input's resolution by adding a dilation rate to standard convolutions. This architecture enlarges the receptive field without increasing the parameters and avoids the loss of information caused by repeated pooling. Some notable examples of representative networks which apply dilated convolution structure include DeepLab series \cite{chen2014semantic, chen2017deeplab, chen2017rethinking, chen2018encoder}, ENet \cite{paszke2016enet}, PSPNet \cite{zhao2017pyramid}, DUC-HDC \cite{wang2018understanding} and DenseASPP \cite{yang2018denseaspp}.

CNNs require multiple decoder stacks to map high-level features to the original spatial resolution. In contrast, transformer-based models can be graciously combined with a lightweight transformer decoder for segmentation mask prediction due to their global modeling capability and resolution invariance. Recently, transformer-based segmentation models (e.g., SETR \cite{zheng2021rethinking}, Segmenter \cite{strudel2021segmenter}, SegFormer \cite{xie2021segformer}, Lawin \cite{yan2022lawin} and MaskFormer \cite{cheng2021per}) extract global contextual features based on self-attention and achieve remarkable results.

\paragraph{Radar-based Semantic Segmentation}

Radar-based semantic segmentation is applied in the fields of vehicle segmentation \cite{schumann2018semantic, ouaknine2021multi}, pedestrian segmentation \cite{schumann2019scene, isele2021learning}, free-space segmentation \cite{nowruzi2020deep, nowruzi2021polarnet} and static object segmentation \cite{lombacher2017semantic, prophet2019semantic} in autonomous driving.
Similar to object detection methods in radar-based applications, network architectures vary depending on specific radar representations. These architectures also incorporate algorithms adapted from the image and point cloud domains to enable efficient processing and analysis of radar data.
Segmentation on radar tensors refers to the process of dividing the tensor into discrete regions or segments based on specific criteria or properties. The goal is to identify and label different parts or objects in the radar RA tensor \cite{kaul2020rss}, RD tensor \cite{orr2021high} and RAD semantic segmentation \cite{nowruzi2020deep, nowruzi2021polarnet, ouaknine2021multi, zhang2023peakconv, rebut2022raw}, enabling a more comprehensive understanding of the scene. CNN architectures like DeepLabv3+ \cite{chen2018encoder} and U-Net \cite{ronneberger2015u} possess the ability to extract intricate features and relationships directly from the radar tensor data, thereby facilitating effective segmentation tasks.

For segmentation on radar point clouds, conventional CNN algorithms (e.g., PointNet and PointNet++) can effectively capture the spatial relationships and semantics of individual radar points to classify them into different categories or segments. These algorithms are widely utilized in point-wise semantic segmentation \cite{schumann2018semantic, danzer20192d, feng2019point, schumann2018supervised, nobis2021kernel, liu2022deep} and grid-based methods \cite{lombacher2017semantic, schumann2019scene, prophet2019semantic, prophet2020semantic, kaul2020rss}. However, the initial data transformation involved in these approaches may result in information loss and sparsity in the data representation.
Recent point transformer networks (e.g., Gaussian Radar Transformer (GRT) \cite{zeller2022gaussian}) enhance performance for 3D point cloud understanding by elaborating the attention mechanism, which is able to capture complex structures in sparse point clouds.
In addition to basic semantic segmentation, instance-based segmentation methods \cite{liu2022deep, xiong2022contrastive} not only classify each point in the radar point cloud but also group nearby points together into instances.






\section{Fusion Datasets} \label{Datasets}

\begin{figure*}[ht]
\begin{center}
\includegraphics[width=1\linewidth]{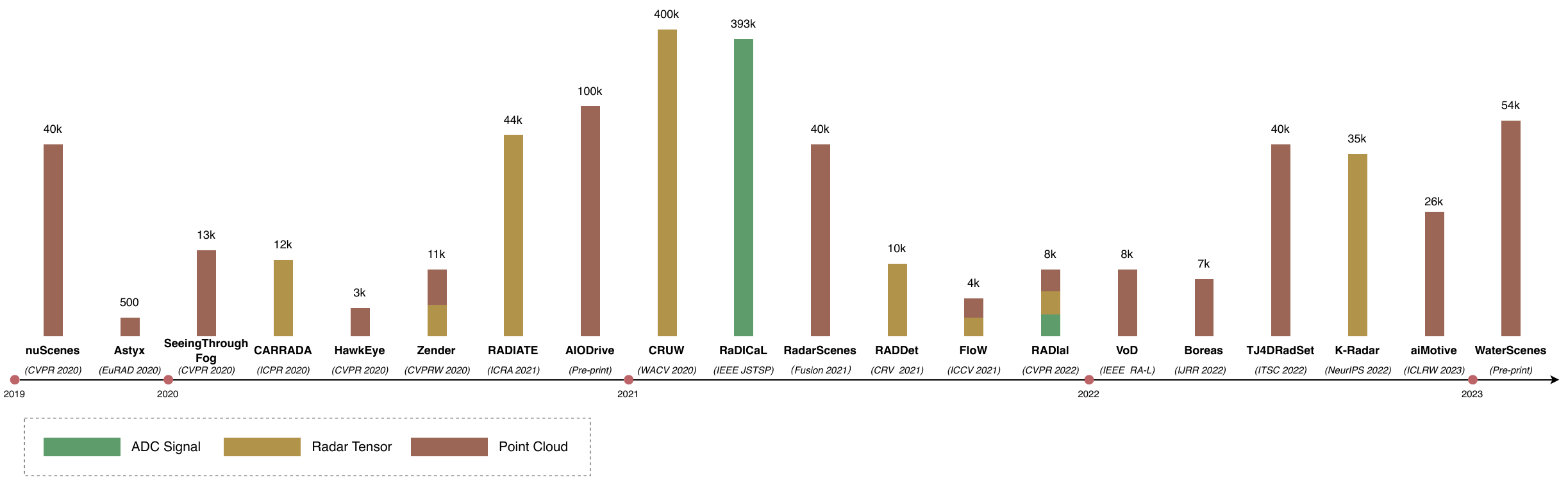}
\end{center}
\vspace{-2mm}
   \caption{Overview of radar-camera fusion datasets. Color of the bar donates different radar representations, and height of the bar represents the frame of each dataset.}
   \label{fig:datasets}
\end{figure*}

High-quality and large-scale data are fundamental for deep learning-based perception algorithms in autonomous driving. Datasets containing data from LiDARs and cameras, such as KITTI \cite{geiger2012we}, Oxford RobotCar \cite{maddern20171}, ApolloScape \cite{huang2019apolloscape}, and Waymo \cite{sun2020scalability}, have been widely used for LiDAR-camera fusion in autonomous driving. As radar research continues in-depth, dozens of radar and camera datasets have been released in recent years. In this section, we analyze and summarize these datasets designed explicitly for tasks related to object detection and semantic segmentation. Fig. \ref{fig:datasets} presents clear statistics of radar-camera fusion datasets with their radar representations and dataset sizes. We also provide a table for retrieval and comparison of different datasets (see Table \ref{tab:datasets}).

\subsection{Dataset Tasks}
\label{Dataset Tasks}

According to the dimension of bounding boxes and masks, datasets that incorporate radar and camera modalities in object detection and semantic segmentation are categorized into four groups: 
\begin{itemize}
  \item \textit{2D object detection:} SeeingThroughFog \cite{bijelic2020seeing}, CARRADA \cite{ouaknine2021carrada}, Zendar \cite{mostajabi2020high}, RADIATE \cite{sheeny2021radiate}, AIODrive \cite{weng2023all}, CRUW \cite{wang2021rethinking}, RaDICaL \cite{lim2021radical}, RADDet \cite{zhang2021raddet}, FloW \cite{cheng2021flow}, RADIal \cite{rebut2022raw}, VoD \cite{palffy2022multi}, Boreas \cite{burnett2023boreas} and WaterScenes \cite{yao2023waterscenes};
  \item \textit{3D object detection:} nuScenes \cite{caesar2020nuscenes}, Astyx \cite{meyer2019deep}, SeeingThroughFog \cite{bijelic2020seeing}, AIODrive \cite{weng2023all}, VoD \cite{palffy2022multi}, TJ4DRadSet \cite{zheng2022tj4dradset}, K-Radar \cite{paek2022k} and aiMotive \cite{matuszka2022aimotive};
  \item \textit{2D semantic segmentation:} CARRADA \cite{ouaknine2021carrada}, RadarScenes \cite{schumann2021radarscenes} and RADIal \cite{rebut2022raw};
  \item \textit{3D semantic segmentation:} HawkEye \cite{guan2020through}.
\end{itemize}


Regarding dataset tasks, most datasets are oriented toward object detection, whereas comparatively fewer datasets are employed for semantic segmentation tasks. Notably, CARRADA \cite{ouaknine2021carrada}, RadarScenes \cite{schumann2021radarscenes} and RADIal \cite{rebut2022raw} can be applied to both object detection and semantic segmentation tasks. As to those datasets that contribute to multiple tasks, nuScenes \cite{caesar2020nuscenes} is the most widely used dataset in radar-camera fusion algorithms, supporting tasks of detection, tracking, prediction and localization. In addition to object detection, RADIATE \cite{sheeny2021radiate} involves object tracking, scene understanding and SLAM tasks. Moreover, datasets like Zender \cite{mostajabi2020high} and Boreas \cite{burnett2023boreas} are available for localization and odometry.

%
%

\subsection{Sensing Modalities}


For radar-camera fusion datasets in object detection and semantic segmentation tasks, the data produced by the camera sensor is either a single image or a video over a while, both of which are essentially 2D images. In comparison, data produced by the radar sensor is rich in representations, which can be grouped into ADC signal, radar tensor, and point cloud according to the stages of data processing.

\subsubsection{ADC Signal}

As the raw data produced by radar sensors, ADC signals retain all semantic information and can be highly valuable in deep learning applications.
Up to now, only two radar-camera fusion datasets provide raw ADC signal data: RaDICaL \cite{lim2021radical} and RADIal \cite{rebut2022raw}. 
RaDICaL \cite{lim2021radical} is the first dataset providing raw ADC signal data, specialized for object detection tasks involving pedestrians and vehicles. The authors encouraged researchers to further design their own processing methods by providing the raw radar measurements. 
RADIal \cite{rebut2022raw} is the richest dataset regarding radar data representations, offering not only ADC signals, but also processed data after ADC signals, including radar tensors and point clouds.

\subsubsection{Radar Tensor}

After multiple FFTs, radar tensors are obtained from ADC signals. They can be classified into three categories: 2D tensors (e.g., RADIATE \cite{sheeny2021radiate}, CRUW \cite{wang2021rethinking}, FloW \cite{cheng2021flow}), 3D tensors (e.g., CARRADA \cite{ouaknine2021carrada}, Zendar \cite{mostajabi2020high}, RADDet \cite{zhang2021raddet}, RADIal \cite{rebut2022raw}) and 4D tensors (e.g., K-Radar \cite {paek2022k}). 
Both RADIATE \cite{sheeny2021radiate} and CRUW \cite{wang2021rethinking} are in range-azimuth coordinates, presenting the BEV position of objects, while the FloW \cite{cheng2021flow} dataset is in range-Doppler coordinates, illustrating the relationship between range and Doppler velocity of each object. 
CARRADA \cite{ouaknine2021carrada} is the first dataset that combines synchronized stereo RGB images and 3D radar RAD tensors in autonomous driving.
As far as we are aware, K-Radar \cite {paek2022k} is the only dataset containing 4D radar tensors, with full information on range, Doppler, azimuth and elevation.

\subsubsection{Point Cloud}

Compared to radar tensors, point clouds serve as a lighter and more intuitive representation of objects. They are also the format of data output from commercial radars. 
Conventional 3D radars produce sparse point clouds, such as data in nuScenes \cite{caesar2020nuscenes}, Zender \cite{mostajabi2020high}, SeeingThroughFog \cite{bijelic2020seeing}, HawkEye \cite{guan2020through}, AIODrive \cite{weng2023all}, RADIal \cite{rebut2022raw}, FloW \cite{cheng2021flow}, RadarScenes \cite{schumann2021radarscenes} and aiMotive \cite{matuszka2022aimotive} dataset.
In recent years, the radar sensor has advanced from 3D to 4D with improvements in resolution and elevation measurement capabilities. Consequently, public 4D radar-camera fusion datasets are emerging, with examples such as Astyx \cite{meyer2019deep}, VoD \cite{palffy2022multi}, TJ4DRadSet \cite{zheng2022tj4dradset} and WaterScenes \cite{yao2023waterscenes}.
Although Astyx \cite{meyer2019deep} is the first 4D point cloud dataset, it is limited by the data size, containing only 500 frames. VoD \cite{palffy2022multi} and TJ4DRadSet \cite{zheng2022tj4dradset} datasets are improved in terms of data categories and data size, with the former consisting of 13 types and 8,693 frames, and the latter containing eight types and 40k frames. Meanwhile, these two datasets also contain simultaneous LiDAR data, facilitating comparison between the 4D radar point clouds and LiDAR point clouds.

\subsection{Dataset Categories}


For autonomous driving, it is critical to identify Vulnerable Road Users (VRU) on roads. Therefore, the most common categories in these datasets are pedestrians, bicycles, and cars \cite{meyer2019deep, ouaknine2021carrada, sheeny2021radiate, weng2023all, wang2021rethinking, schumann2021radarscenes, palffy2022multi}.
Datasets such as nuScenes \cite{caesar2020nuscenes}, AIODrive \cite{weng2023all}, RadarScenes \cite{schumann2021radarscenes}, VoD \cite{palffy2022multi} and aiMotive \cite{matuszka2022aimotive} have studied more than ten categories. nuScenes \cite{caesar2020nuscenes} provides precision in its classifications, with 23 object categories refining certain ambiguous categories. For instance, the pedestrian category is sub-categorized into groups such as adult and child, while the vehicle category is subclassed into the car, ambulance, police, motorcycle, trailer and truck. 
 Except for the category of pedestrian, RADIATE \cite{sheeny2021radiate} and RadarScenes \cite{schumann2021radarscenes} include a class called pedestrian group.
 AIODrive \cite{weng2023all} and VoD \cite{palffy2022multi} classify stationary objects on the roadside, such as building, road, wall, traffic sign, unused bicycle and bicycle rack.
 Furthermore, apart from objects on road surfaces, FloW \cite{cheng2021flow} is a floating waste dataset containing the category of bottle that can be utilized for Unmanned Surface Vehicles (USVs) on water surfaces. WaterScenes \cite{yao2023waterscenes} contains more objects of interest on water surfaces, including static objects such as piers and buoys, and dynamic objects such as ships, boats, vessels, kayaks, and sailors aboard these surface vehicles.

In addition to the primary object categories, some specific attributes of the object are also labeled in some datasets. Examples can be found in nuScenes \cite{caesar2020nuscenes}, in which vehicles are labeled as moving, stopped or parked, while pedestrians are marked as moving or standing.  
Besides, in VoD \cite{palffy2022multi} dataset, two types of occlusions (``spatial" and ``lighting") and attributes related to an object's activity (``stopped", ``moving", ``parked", ``pushed" and ``sitting") are also annotated. All these specific attributes are essential for scene understanding. 
 
\subsection{Dataset Size}


The reviewed datasets differ significantly in size, ranging from 500 to 1.4 million frames. nuScenes \cite{caesar2020nuscenes} is the largest dataset with 1.4 million images, radar frames and object bounding boxes in 40k keyframes. These data frames are split from 15 hours and 242 kilometers of driving data. 
On the other hand, Astyx \cite{meyer2019deep} provides only 500 frames, containing around 3k labeled 3D object annotations. 
Others like CRUW \cite{wang2021rethinking}, CARRADA \cite{ouaknine2021carrada}, RADIATE \cite{sheeny2021radiate}, AIODrive  \cite{weng2023all}, SeeingThroughFog  \cite{bijelic2020seeing}, CRUW  \cite{wang2021rethinking}, RADDet \cite{zhang2021raddet} and RADIal \cite{rebut2022raw} all contribute hundreds of thousands of frames.

In addition to data in frames, some datasets deliver videos for researchers to split keyframes and conduct further research on videos. For example, RadarScenes \cite{schumann2021radarscenes} offers 158 individual sequences with a total length of over four hours. Similarly, CRUW \cite{wang2021rethinking} and CARRADA \cite{ouaknine2021carrada} datasets provide videos of 3.5 hours and 21.2 minutes, respectively. 

\subsection{Recording Scenarios}


A rich data collection environment is crucial for training robust models in autonomous driving. Generally, datasets for autonomous driving are collected in road environments like urban streets, country roads, highways and parking lots, which are all represented in datasets like CARRADA \cite{ouaknine2021carrada}, RADIATE \cite{sheeny2021radiate}, RadarScenes \cite{schumann2021radarscenes}, RADlal \cite{rebut2022raw}, and K-Radar \cite{paek2022k}. 
However, it is not enough to collect data in common areas. nuScenes \cite{caesar2020nuscenes}, Zendar \cite{mostajabi2020high}, SeeingThroughFog \cite{bijelic2020seeing}, AIODrive \cite{weng2023all} and CRUW \cite{wang2021rethinking} involve dense traffic and challenging driving situations, including urban roads, residential areas and industrial areas. Moreover, RadarScenes \cite{schumann2021radarscenes} offers data for selected particular scenarios, such as T-junctions, commercial areas and road works. 
All the datasets mentioned above are from outdoor environments, and as to indoor scenarios, HawkEye \cite{guan2020through} and RaDICaL \cite{lim2021radical} are collected in indoor parking garages. These indoor environments present unique challenges and can help advance research in indoor autonomous vehicle navigation.

In terms of weather and lighting conditions, related data can be found in nuScenes \cite{caesar2020nuscenes}, SeeingThroughFog \cite{bijelic2020seeing}, CARRADA \cite{ouaknine2021carrada}, RADIATE \cite{sheeny2021radiate}, K-Radar \cite{paek2022k}, AIODrive \cite{weng2023all}, CRUW \cite{wang2021rethinking}, TJ4DRadSet \cite{zheng2022tj4dradset}, aiMotive \cite{matuszka2022aimotive} and WaterScenes \cite{yao2023waterscenes}. 
In particular, SeeingThroughFog \cite{bijelic2020seeing} focuses on extreme weather conditions, like fog, snow and rain. This dataset highlights the importance of data fusion and the redundancy of multiple sensors in adverse weather environments.
In addition to adverse weather conditions, RADIATE \cite{sheeny2021radiate}, K-Radar \cite{paek2022k} and aiMotive \cite{matuszka2022aimotive} involve conditions for the night. AIODrive \cite{weng2023all}, CRUW \cite{wang2021rethinking} and TJ4DRadSet \cite{zheng2022tj4dradset} supply data in specific vision-fail scenarios, such as darkness, bright light and blur, where the images are pretty bad in quality. These datasets provide valuable information about how autonomous driving technology operates in low visibility and low light scenarios.
Boreas \cite{burnett2023boreas} is a dataset involving data taken from specific routes through one-year repeat collection. It can be leveraged to study the effects of seasonal variation on self-localization and object detection.

Since it is time-cost and resource-cost for data collection on roads, some researchers have adopted simulated data to generate datasets.
In HawkEye \cite{guan2020through}, raw low-resolution heatmaps are transformed into high-frequency shapes. Additionally, the authors developed a data synthesizer to simulate radar signals from the created 3D point reflector models of cars. Meanwhile, Weng~\etal~\cite{weng2023all} used Carla \cite{dosovitskiy2017carla} simulator to create different driving scenarios with various sensors. With the annotation data generated by combining and post-processing Carla outputs, they presented AIODrive \cite{weng2023all}, a large-scale synthetic dataset for all mainstream perception tasks.

\section{Fusion Methodologies} \label{Fusion Methodologies}

\begin{figure*}[htbp]
\begin{center}
\includegraphics[width=1\linewidth]{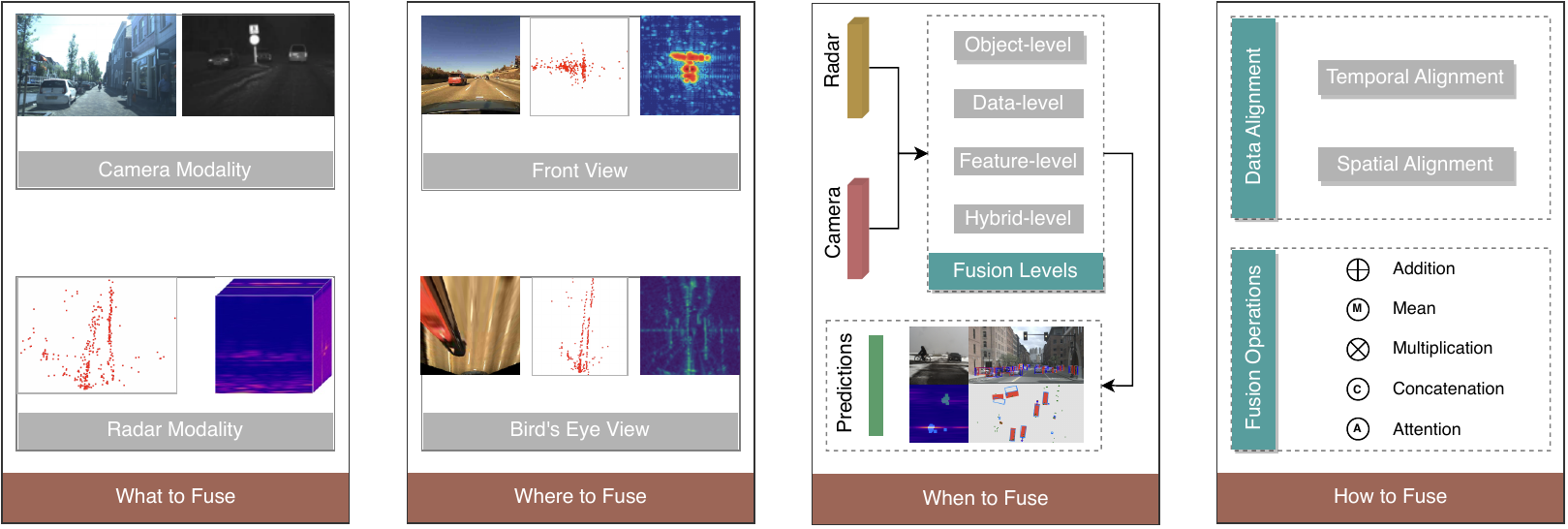}
\end{center}
\caption{Methodologies of radar-camera fusion. (a) ``What to fuse'': the input modalities for radar-camera fusion, such as RGB images, near-infrared images, point clouds, and radar tensors. (b) ``Where to fuse'': the position transformation for radar-camera fusion, including two perspectives: front view and bird's eye view. (c) ``When to fuse'': at which stage the two modalities are fused, encompassing the fusion of final objects, raw data, extracted features and hybrid representations. (d) ``How to fuse'': the data alignment and fusion operations, comprising of two types of alignment, namely temporal alignment and spatial alignment, and five types of fusion operations, namely addition, mean, multiplication, concatenation and attention.}
\label{fig:methodology}
\end{figure*}


In this section, we delve into the methodologies of radar-camera fusion related to object detection and semantic segmentation tasks, starting with ``why to fuse'', that is, the purpose and advantages of fusion. 
Subsequently, we analyze ``what to fuse'', covering diverse representations of both radar and camera modalities implicated in fusion. 
Next, we investigate ``where to fuse'', describing the coordinate relations between the two modalities before fusion. 
In the section on ``when to fuse'', we categorize the fusion levels and illustrate their differences.
After that, we explore the specifics of ``how to fuse'', including temporal-spatial synchronization and fusion operations.
Regarding network architectures for fusion, we categorize them into two architectures: point-based and tensor-based, followed by a more detailed classification of each category and the main ideas in these architectures. 
Finally, in the model evaluations section, we review various evaluation metrics and assess the performance of popular methods.
An overview of radar-camera fusion methodologies containing the questions and answers is demonstrated in Fig. \ref{fig:methodology}.

\subsection{Why to Fuse}


The integration of radar and camera sensors for object detection and semantic segmentation is intended to enhance the perception outcomes by capitalizing on the advantages of both sensing modalities.
As illustrated in Fig. \ref{fig:Comparison of characteristics of radar and camera} in Section \ref{sec:Comparison of Radar and Camera}, the combination of a radar sensor and a camera sensor enables the measurement of rich object attributes such as color, shape, range, and velocity.
In addition, with the ability to perceive in darkness and adverse weather conditions, the fusion of radar and camera can work all day for autonomous driving vehicles. 
For autonomous driving tasks, object detection and segmentation results obtained from radar-camera fusion can also assist in object tracking \cite{liu2021robust, nabati2021cftrack}, providing accurate environment perception information for the decision-making and control systems.
For other downstream tasks, such as trajectory prediction \cite{benterki2020artificial} and vehicle navigation \cite{cai2020probabilistic}, radar-camera fusion has successfully demonstrated excellent driving performance in both unseen urban and heavy traffic scenarios.

Numerous studies have also demonstrated that radar-camera fusion improves the accuracy and robustness of the network.
As it is difficult for image-based detectors to detect distant objects, Chadwick~\etal~\cite{chadwick2019distant} combined a radar sensor and sets of camera sensors in their experiments. Results exceed the performance of the camera detector, as the radar sensor persists in delivering a potent indication of motion for faraway objects.
Major~\etal~\cite{major2019vehicle} also proved that the velocity dimension derived from the radar sensor could be leveraged to increase detection performance.
Additionally, Nabati and Qi~\cite{nabati2021centerfusion} utilized radar features (e.g., depth, rotation, velocity) to complement the image features, resulting in an improvement of the overall nuScenes Detection Score (NDS) by more than 12\% compared to the SOTA camera-based algorithm including OFT \cite{roddick2018orthographic}, MonoDIS \cite{simonelli2019disentangling} and CenterNet \cite{zhou2019objects}. 
In noisy circumstances, Yadav~\etal~\cite{yadav2020radar+} discovered that radar data exhibit robustness in detection, and integration of radar data could enhance performance in these challenging scenarios.

\subsection{What to Fuse}


The objective of radar-camera fusion is the output data from the radar sensor and camera sensor, which are presented in different modalities at various fusion levels and via different fusion techniques.
For the camera sensor, the output data is typically presented as 2D images. In radar-camera fusion, there are mainly two kinds of images. One type is the RGB image with rich color information, such as images in nuScenes \cite{caesar2020nuscenes} dataset. The other is the infrared image captured with infrared cameras (including Far Infrared (FIR) and Near Infrared (NIR)), as illustrated in the images from SeeingThroughFog \cite{bijelic2020seeing} dataset. Though in lower resolutions, these images contain specific advantages in temperature differences and night visibility.
The data structure of an image is relatively simple, with low data dimensionality and high correlations between neighboring pixels. The simplicity of this structure allows deep neural networks to learn the fundamental representations of images, thus enabling them to detect objects within the images \cite{paek2022k}. 

As mentioned in Section \ref{Radar Data Representations}, radar data can be classified into different representations depending on the level of processing. ADC Signals, the underlying digital signals of the radar, cannot be marked with the location information of an object. MDS, a Time-Frequency representation, consists of consecutive radar frames and does not correspond to a single image frame. As a result, ADC Signals and MDS are commonly used for identifying the presence of objects and discriminating between different objects \cite{zhao2023cubelearn, stadelmayer2021data, chen2022attention, kim2015human}. With the ability to describe the shape of an object, radar tensors and point clouds are commonly leveraged for object detection and semantic segmentation tasks.

\subsection{Where to Fuse}


\subsubsection{Front View}


Fusion at the Front View (FV) involves projecting the radar data onto an image plane, where the radar data can be 3D point clouds, partial point cloud information or radar tensors.
Around the projected area, proposals that indicate potential objects are generated \cite{nabati2019rrpn, nabati2020radar, yadav2020radar+, cui20213d, li2020feature}. In this way, a large number of non-object regions are excluded, thus reducing the computational burden and increasing recognition speed.
Radar data mapped to the image plane can also be utilized to create feature maps for complementing the image-based features \cite{chadwick2019distant, kowol2020yodar, nabati2021centerfusion, cheng2021robust}. These methods improve the detection accuracy by leveraging the additional input, including ranges, velocities and RCS values.
Moreover, some researchers project radar point clouds onto the image plane to form a radar pseudo-image \cite{john2019rvnet, john2020so, dong2021radar, song2022ms}. For example, In RVNet \cite{john2019rvnet} and SO-Net \cite{john2020so}, a pseudo-image named ``Sparse Radar Image" is generated from radar data, containing information regarding depth, lateral velocity and longitudinal velocity. Besides, Dong~\etal~\cite{dong2021radar} projected both radar point clouds and 2D bounding boxes onto the image plane, forming new pseudo-images from camera RGB images. MS-YOLO \cite{song2022ms} generates radar mask maps by a mapping transformation neural network. In each mask map, the boxed area represents the presence of an object, and the gray value of each box indicates the velocity information of that object.

Projecting radar data onto the image plane assists in providing proposals and features. However, due to the low resolution in the azimuth angle provided by the radar as well as camera calibration errors, projected radar point clouds may deviate from the object. While increasing the Region of Interest (RoI) could potentially address the issue, it results in multiple objects within the same region, and consequently being detected repeatedly, causing confusion in object matching. Moreover, due to the occlusion of objects, the projection of radar data onto the image perspective may be limited.

\subsubsection{Bird's Eye View}


Another fusion position is to convert radar data or camera images into BEV coordinates.
For example, radar point clouds from each frame generate a BEV image of six height maps and one density map in \cite{meyer2019deep}. 
Besides, Cui~\etal~\cite{cui20213d} projected radar point clouds to both FV and BEV, and proposed a 3D region proposal network to generate proposals from both camera images and radar BEV images. Compared with generating proposals directly from point clouds, the CNN-based proposal generation approach increases the quality of proposals by leveraging the network's ability to extract deeper and richer information.
Problems come that BEV images discretize the sensing space into grids, which may lead to the loss of valuable information necessary to refine bounding boxes. To address this issue, Bansal~\etal~\cite{bansal2022radsegnet} added additional point-based features (e.g., velocities, RCS values) to the BEV map. 
Simple-BEV~\cite{harley2023simple} converts all radar point clouds from multiple radar sensors into BEV coordinates to yield high-dimensional BEV feature maps.

Apart from projecting radar data into BEV, Inverse Projection Mapping (IPM) \cite{bertozz1998stereo, oliveira2015multimodal} method can be utilized to convert camera images from FV to BEV with a homography matrix. 
For instance, Lim~\etal~\cite{lim2019radar} transformed the camera images into Cartesian coordinates using IPM and then combined them with 2D radar RA tensors.
In addition, both radar point clouds and camera images are projected onto BEV in \cite{zhou2023bridging}, where the independent feature extractors learn shared features.
Consequently, projecting data on BEV offers several advantages over FV, particularly in the case of occlusion \cite{wang2019pseudo}. Nonetheless, since IPM is based on an assumption of flat road surfaces, it often produces distortions of dynamic objects when applied to real-world scenarios \cite{zhou2023bridging}. 

\subsection{When to Fuse}


\begin{figure*}[t]
\begin{center}
\includegraphics[width=0.8\linewidth]{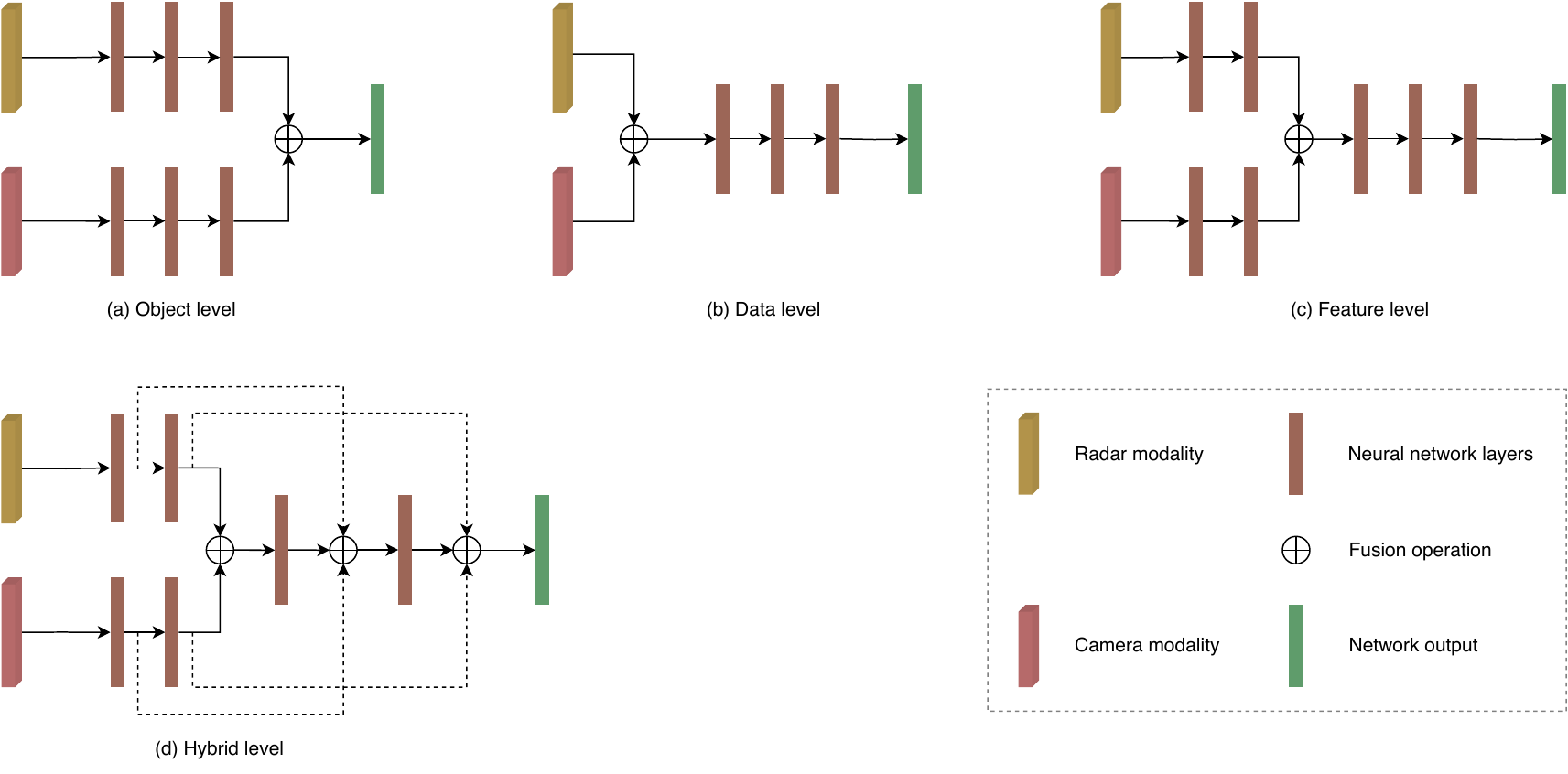}
\end{center}
\caption{Taxonomy of radar-camera fusion levels. (a) Object-level fusion. (b) Data-level fusion. (c) Feature-level fusion. (d) Hybrid-level fusion.}
\label{fig:fusion-level}
\end{figure*}

When to fuse refers to at which stage the radar and camera data are fused in the network. Based on the occasion of the fusion process, we classify radar-camera fusion levels into object-level, data-level, feature-level and hybrid-level. Fig. \ref{fig:fusion-level} illustrates the overview and difference between the four fusion levels.

\subsubsection{Object-level Fusion}


For object-level fusion (also known as decision-level fusion or late-level fusion), the independent objects acquired from the radar and camera sensors are fused at a later stage of the network to obtain the final integrated results, as demonstrated in Fig. \ref{fig:fusion-level}(a). 
In object-level fusion, how to match the results from the two different modalities is worth considering.
One way is to calculate the similarity (e.g., location, size, category) and then employ methods such as the Kalman filter, Bayesian theory, Hungarian algorithm and Bipartite Matching to match the outputs.
Another approach involves utilizing the transformation matrix between the radar and camera to determine the position relationships between the two modalities. For example, Jha~\etal~\cite{jha2019object} projected the radar detections onto the image plane using the transformation matrix, and then aligned independent detection objects from the two sensors.
Moreover, after completing the association of radar point clouds with camera images, Dong~\etal~\cite{dong2021radar} proposed AssociationNet for learning the semantic representation information from the two sensors. This network improves the accuracy of association by calculating and minimizing Euclidean distance between the representations from the pair of radar point clouds and image bounding boxes. 

Object-level fusion is commonly used in conventional radars and cameras, which offers high flexibility and modularity \cite{kim2020low}.
However, it also relies heavily on the accuracy of outputs from individual modules. For example, in scenarios where the camera sensor is obstructed, object-level fusion exclusively depends on the final objects detected by the radar sensor. 
Besides, rich intermediate features are discarded due to the sensing modality's weaknesses or errors in the sensors. As a result, object-level fusion methods can only utilize limited information obtained from the detection results.

\subsubsection{Data-level Fusion}


For data-level fusion (also referred to as low-level fusion or early-level fusion), the raw data or pre-processing data from radar and camera sensors are fused at the early stage of deep learning models, illustrated in Fig. \ref{fig:fusion-level}(b). 
Nobis~\etal~\cite{nobis2019deep} fed the concatenated camera and radar point clouds into the network and then employed VGG \cite{simonyan2014very} to extract features from the combined data.
Moreover, Bansa~\etal~\cite{bansal2022radsegnet} created a Semantic-Point-Grid (SPG) representation from camera semantic maps, radar point clouds and radar BEV grid maps. In their method, the SPG representation is then fed into SPG encoding to extract semantic information from cameras, aiding in the identification of radar points associated with objects of interest.
Instead of fusing radar point clouds with camera images, Nabati and Qi~\cite{nabati2019rrpn} proposed an RRPN, which generates object proposals to narrow the scope of detection on the camera images. However, if there is no radar point on an object, this object will be ignored. 
To solve the difficulty of associating radar point clouds with image pixels, Long~\etal~\cite{long2021radar} presented Radar-Camera Pixel Depth Association (RC-PDA), a learned method that associates radar point clouds with nearby image pixels to enhance and densify the radar image.

With the input of raw data, it is possible to exploit complete characteristics and learn a joint representation from these two modalities. However, data-level fusion methods tend to be sensitive to temporal or spatial misalignment within the data. Precise external calibration of the two sensors is essential for data-level fusion. 
Besides, as radar data representations are not consistent with the object's shape, it is difficult to match the radar tensors or radar point clouds with the image pixels. 


\subsubsection{Feature-level Fusion}

\begin{figure*}[t]
\begin{center}
\includegraphics[width=0.8\linewidth]{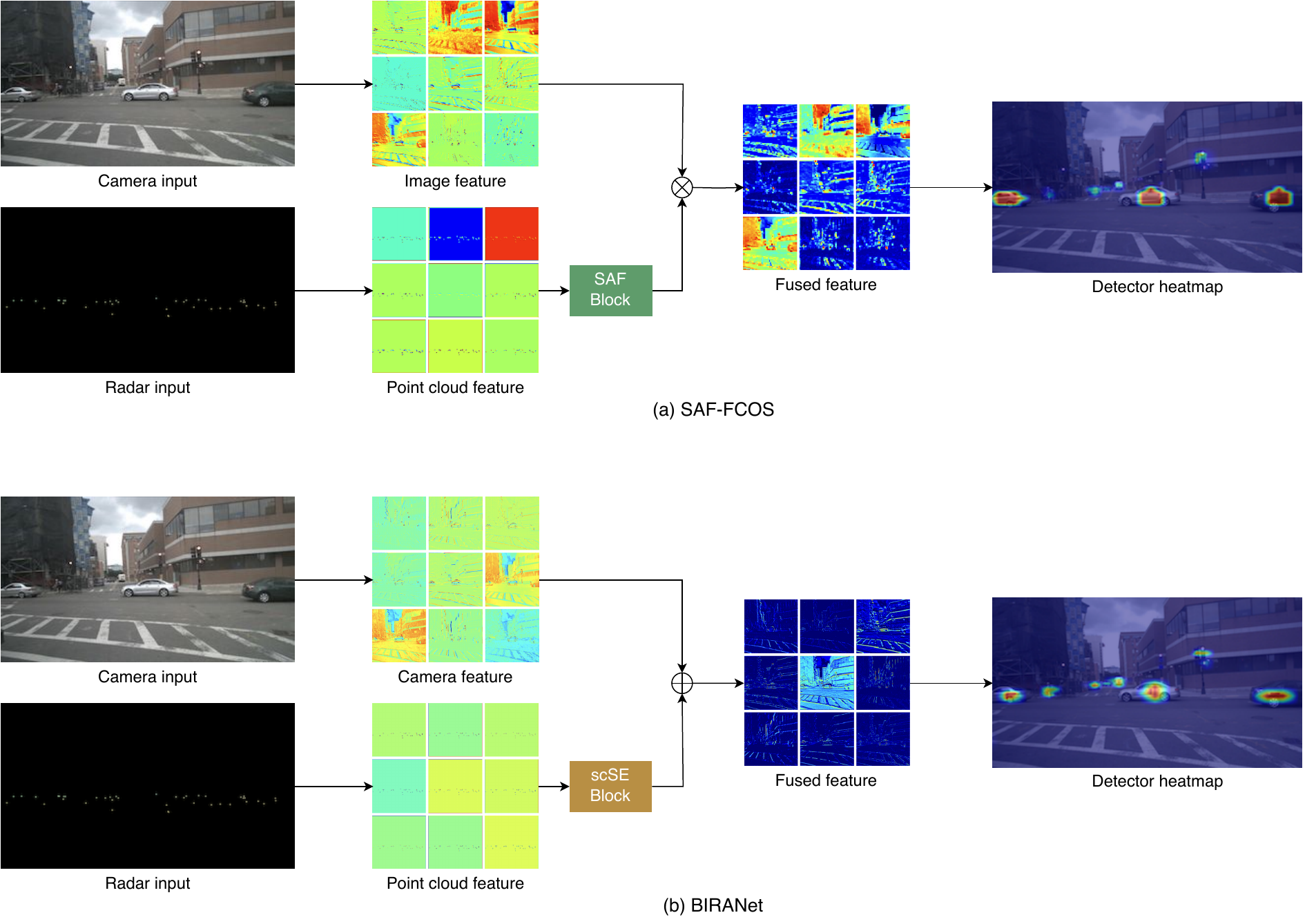}
\end{center}
\caption{Feature heatmaps of two feature-level fusion networks, namely (a) SAF-FCOS \cite{chang2020spatial} and (b) BIRANet \cite{yadav2020radar+}. Both networks take inputs from camera and radar branches, and extract image features and point cloud features using respective backbones. In the feature fusion stage, SAF-FCOS employs a multiplication operation, while BIRANet performs an element-wise addition operation. Detector heatmaps generated by the Grad-CAM \cite{selvaraju2017grad} are utilized for the result analysis of each method. To clearly show feature maps, only nine channels are selected for visualization.}
\label{fig:heatmap}
\end{figure*}


In feature-level fusion (also called middle-level fusion), features extracted from separate radar data and camera images are combined at an intermediate stage in deep learning-based fusion networks, as shown in Fig. \ref{fig:fusion-level}(c). 
In \cite{chadwick2019distant}, features from both radar and camera branches are generated by ResNet \cite{he2016deep} blocks and then fused by concatenation and addition operations. 
CenterFusion \cite{nabati2021centerfusion} detects objects by locating their center points in the image using CenterNet \cite{zhou2019objects}. After that, it utilizes a frustum-based association strategy to accurately match radar detections with objects in the image, generating radar-based feature maps to augment the image features. 
SAF-FCOS \cite{chang2020spatial} introduces the attention mechanism for weighting different positions of the feature maps. Specifically, it utilizes a Spatial Attention Fusion (SAF) block to merge the feature maps from radar and camera. In the SAF block, the radar image's feature maps are encoded as a spatial attention weight matrix, which is then applied to all channels to re-weight the feature maps extracted by the camera sensor.
BIRANet \cite{yadav2020radar+} uses Concurrent Spatial and Channel Squeeze \& Excitation (scSE) blocks \cite{roy2018concurrent} to highlight important spatial features and significant channels. The scSE block acts as attention and adaptively boosts activation of areas where radar point clouds are present while suppressing activation at other locations. This boosted feature map is then fused with the image feature map to improve the performance of the detection network. 
Considering feature maps that provide enhanced spatial and channel-wise information, BIRANet exhibits the capability to detect small objects that SAF fails to identify, as illustrated by the detector heatmaps shown in Figure \ref{fig:heatmap}.
In fact, attention maps can be generated from various sensors. Bijelic~\etal~\cite{bijelic2020seeing} extended the sensors to RGB camera, gated camera, LiDAR and radar by transforming all sensor data into uniform image coordinates. The feature maps of different sensors are then superimposed together by concatenation and multiplied with the sigmoid-processed entropy map for the final feature output.

For feature-level fusion, it is possible to design appropriate feature extractors for each modality according to its specific characteristics. Neural networks can also learn features jointly across modalities, making them complementary to each other.
However, it is worth noting that feature extraction and feature fusion do not address scenarios where camera sensors become unreliable \cite{kim2020low}.

\subsubsection{Hybrid-level Fusion}


Apart from equally fusing final objects, raw data or features from two modalities, some fusion methods combine different stages of data, which we define as hybrid-level fusion, shown in Fig. \ref{fig:fusion-level}(d). 
In \cite{nabati2020radar}, radar proposals are first generated from radar point clouds and 3D anchors derived from camera images. Then a Radar Proposal Refinement (RPR) network is proposed to fuse the radar proposals with camera image features, which enables the adjustment of the size and location of the radar proposals in the image. Besides, the RPR network also estimates an objectness score for each radar proposal, as some radar point clouds are caused by background noise.
Similarly, Cui~\etal~\cite{cui20213d} generated proposals based on camera images and radar BEV point clouds, followed by projecting the proposals onto three feature maps from camera images, radar BEV point clouds and radar FV point clouds. A Self-Supervised Model Adaptation (SSMA) block \cite{valada2020self} is utilized to fuse the proposals with features, which leverages an attention scheme for better correlation. 
Furthermore, HRFuser \cite{broedermann2022hrfuser} introduces ideas from HRNet \cite{yuan2020object} and HRFormer \cite{yuan2021hrformer}, adopting an asymmetric Multi-Window Cross-Attention (MWCA) to fuse the features captured by the RGB camera, LiDAR, radar and gated camera.

Compared with data-level and feature-level fusion, fusion from both proposals and features leads to more accurate proposals, producing better features for the two-stage network \cite{nabati2020radar, cui20213d}.
Generally, different modalities have different contributions to radar-camera fusion. One modality dominates, while the other provides supplementary information to refine the features. Thus, hybrid-level fusion takes advantage of different data levels and effectively preserves information at various stages. 
However, hybrid-level fusion should consider the importance of different modalities, which also pose implementation challenges. Since most implementations of hybrid-level fusion are based on experience and lack explainability to some extent, conducting numerous ablation experiments is needed to validate the efficiency of hybrid-level fusion. Moreover, models based on hybrid-level fusion typically have more branches in neural networks, dramatically slowing down the inference time.

\subsection{How to Fuse}

In this section, we present how to fuse radar data with camera images.
First of all, the primary consideration is the temporal and spatial alignment between the two sensors.
Then, in fusion operations, we compare five operations and analyze their advantages and disadvantages. 

\subsubsection{Data Alignment}
\paragraph{Temporal Alignment}


Temporal alignment in sensor fusion refers to synchronizing the temporal sequences of data from different sensors.
To obtain high-quality fusion results, the data collected by each sensor must be synchronized with the same time dimension. However, there may be time offsets between these sensors due to the differences in set-up time, crystal oscillator frequencies and measurement latency. 
Depending on the object of the temporal alignment methods, we categorize them into two types: estimating temporal latency between sensors and estimating temporal offset within the same frame.

\textbf{Estimating Temporal Latency:}
Generally, temporal latency consists of the measurement latency between sensors and the drift between different frames. 
Measurement latency mainly stems from computer scheduling, measurement acquisition, pre-processing, and communication transfer time. In aligning cycle time, drift is caused by the offset between the internal clock and the Coordinated Universal Time (UTC). 

A software-based technique for reducing temporal error is periodically estimating the maximum measurement latency and drift time \cite{olson2010passive, westenberger2011temporal, furgale2013unified}. Another alternative approach is to predict the future latency between sensors using Kalman filters \cite{huck2011precise} or Bayesian estimation \cite{guivant2012distributed} based on prior knowledge of the sensors' latency.
These methods improve the synchronization result and are suitable for most applications. Since trigger signals for sensors are not initiated simultaneously, there inevitably remains some degree of unknown latency, which can cause variations in acquisition times during data fusion. 
Thus, some researchers proposed solutions by combining a hardware controller trigger with the software strategy to reduce the execution time of activation threads in software \cite{steinbaeck2020hybrid, english2015triggersync, lu2019ieee}. These approaches communicate with hardware synchronization components at a low level to eliminate the data acquisition latency. However, standard commercial hardware often lacks hardware synchronization interfaces \cite{huber2009temporal}. When using such methods, the complexity and portability of the system design should be considered.

\textbf{Estimating Temporal Offset:}
As the temporal offset between sensors directly affects the fusion quality, some studies proposed temporal calibration strategies based on aligning the same objects from camera and radar sensors to extract timestamp offset. For example, Du~\etal~\cite{du2021novel} aligned the frames that a vehicle passes the detection line and then estimated the temporal offset between these two frames. Moreover, some researchers \cite{fu2020camera, liu2008immpda, nobis2019deep} suggested employing real-time pre-processing buffers that leverage algorithms like YOLOv3 \cite{redmon2018yolov3} and DBSCAN \cite{ester1996density} to reorganize the same frame.

\paragraph{Spatial Alignment}

Spatial alignment between radar and camera sensors involves transformation operations that map 3D or 2D radar point clouds to camera image pixels.
As the spatial calibration between radar and camera is a fundamental task for information fusion, several methods of joint calibration have been proposed. 
Among these approaches, whether a specially designed calibration target is needed in the calibration process is an important indicator, leading to different calibration design strategies. 

\textbf{Target-based Approaches:}
For target-based calibration approaches, specific calibration targets are utilized so that sensors can get precise locations of the target. These locations estimate the rigid transformation relations between the radar and camera sensors. The triangular corner reflector is the most common choice for radar calibration, reflecting specific RCS values for the positional information. 
Moreover, to obtain positions of the calibrated targets from both the radar sensor and camera sensor, some novel designed calibration boards are proposed. For example, a corner reflector and a styrofoam board are combined as the calibration target in \cite{wang2021roadside, pervsic2021spatiotemporal}. The styrofoam board is applied for visual recognition in camera sensors without affecting the radar signals. Wang~\etal~\cite{wang2021roadside} designed a calibration board consisting of a styrofoam board with four holes and a corner reflector in the center of these holes. The Perspective-n-Point (PnP) \cite{li2012robust} algorithm is then used to extract the 3D location of holes and estimate the location of the corner reflector. Moreover, Per\v{s}i\'{c}~\etal~\cite{pervsic2021spatiotemporal} introduced a calibration board consisting of a corner reflector and a styrofoam triangle adorned with a checkerboard, from which both radar and camera sensors can obtain accurate target positional readings. Based on the paired set of image pixels and radar points for the same targets in different locations, the transformation matrix between the radar coordinates and camera coordinates is calculated. 

\textbf{Target-less Approaches:}
On the other hand, target-less calibration approaches do not rely on specific checkerboards, thus improving the portability of calibration. However, the uncertainty of environmental factors when extracting the same features from multiple sensors is a common drawback in target-less calibration methods. 
Some researchers utilize precise radar velocity measurements based on the moving objects and the camera pose to implement radar-to-camera extrinsic calibration algorithms \cite{wise2021continuous, pervsic2021online, qiu2020real, pervsic2021spatiotemporal, guo2018pedestrian}.
Besides, machine learning algorithms are also utilized in \cite{iyer2018calibnet, scholler2019targetless, wang2021roadside, zhao2021calibdnn} to predict calibration parameters based on improving the consistency of radar point clouds and camera images. 

\begin{figure*}[ht]
\begin{center}
\includegraphics[width=1\linewidth]{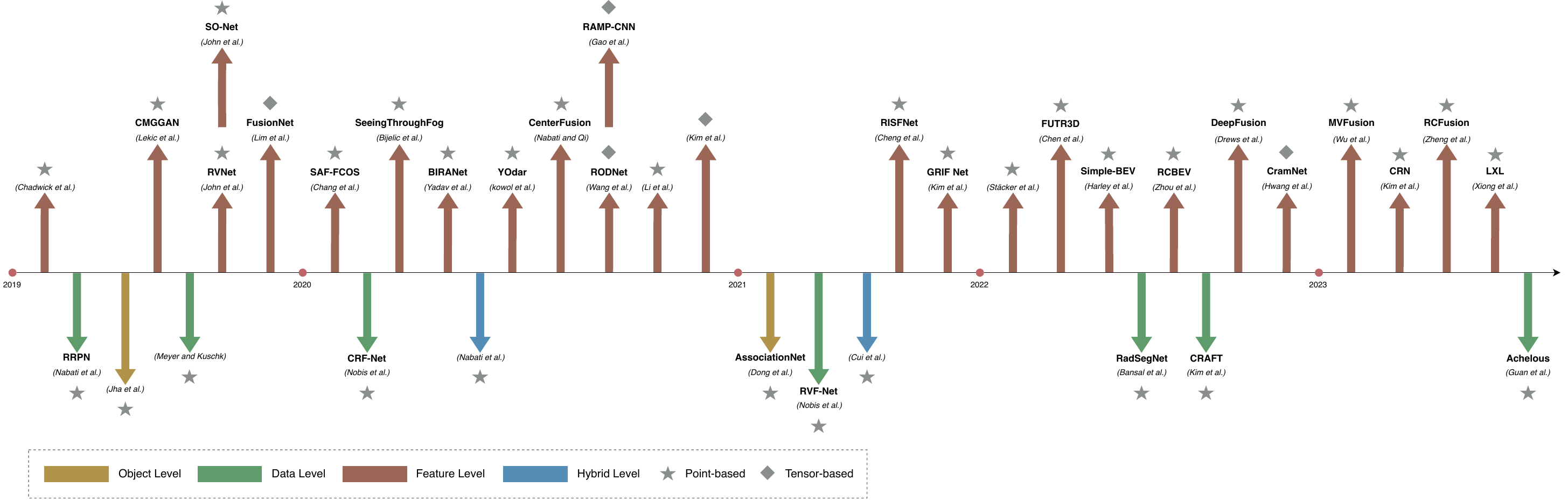}
\end{center}
\caption{Chronological overview of radar-camera fusion algorithms. Color of the arrow represents different fusion levels, and shape of the icon represents different network architectures.}
\label{fig:networks}
\end{figure*}

\subsubsection{Fusion Operations}

In radar-camera fusion, different fusion operations are used to fuse data from the two modalities.
Specifically, for object-level and data-level fusion networks, a transformation matrix is commonly used to align final objects or raw data \cite{jha2019object, nobis2021radar}.
In contrast, feature-level and hybrid-level fusion networks tend to utilize addition and concatenation operations.
In the \textbf{addition} operation, element-wise features in the feature maps are added. Thus, each channel in the feature map contains more feature information, making the classifier comprehend the feature details.
Similarly to the addition operation, the \textbf{mean} and \textbf{multiplication} operations calculate the average mean and multiplication of the element-wise feature maps, respectively.
In the \textbf{concatenation} operation, the feature maps are flattened into vectors and then concatenated along the rows. The primary objective of the concatenation operation is to enrich feature diversity, enabling the classifier to recognize objects with higher accuracy.

Given that the features of radars and cameras are heterogeneous, and the above fusion operations are sensitive to changes in the input data, the effectiveness of the modalities in particular scenarios is ignored. For example, the performance of the camera sensor tends to reduce in adverse weather conditions, while the radar sensor continues to work properly. Thus, the \textbf{attention} operation is proposed to re-calculate the weights of the feature maps from two modalities. One example of such an approach is Spatial Attention Fusion (SAF), proposed by Chang~\etal~\cite{chang2020spatial}. SAF extracts the spatial attention matrix from the radar images and then employs it to re-weight the feature maps from the image branch. 
Other approaches leverage the Mixture of Expert (MoE) \cite{jacobs1991adaptive, kim2020grif} to extract feature maps from respective expert networks and calculate the attention weights by a gating network. After that, based on these weights, the feature maps are re-assigned to optimize fusion performance.

\subsection{Network Architectures}

Generally, networks for radar-camera fusion are structured with dual input branches, where the data from radar and camera is input separately. Depending on the desired fusion stage, raw data, feature maps or final objects are fused in the designed network for fusion results. Based on the representations of radar data, we classify the radar-camera fusion networks in object detection and semantic segmentation tasks into point-based and tensor-based networks. We also provide a chronological overview of radar-camera fusion algorithms in Fig. \ref{fig:networks}, and summarize the comparable contents in Table \ref{tab:fusion methods}. 

\subsubsection{Point-based Networks}

Point-based Networks take radar point clouds as input. According to the different radar point cloud processing methods, we subdivide the point-based methods into projection-based, pseudo-image-based, voxel-based and BEV-based methods.


\textbf{Projection-based Methods:}
In point-based radar-camera fusion networks, radar point clouds are mostly projected onto the 2D image plane to provide proposals or features. Then networks such as VGG \cite{simonyan2014very}, ResNet \cite{he2016deep}, U-Net \cite{ronneberger2015u}, YOLOv3 \cite{redmon2018yolov3} and YOLOv4 \cite{bochkovskiy2020yolov4} are used for feature extraction.
Chadwick~\etal~\cite{chadwick2019distant} projected the radar point clouds onto the camera plane and generated two kinds of radar images: range image and range-rate image. Then they integrated an additional radar input branch upon the SSD \cite{liu2016ssd} network, and used both concatenation and element-wise addition operations to fuse the radar features after the image block. The branch structure exhibits potential flexibility in re-calculating weights between the camera image and radar representations. 
Besides, Meyer and Kuschk~\cite{meyer2019deep} generated a BEV image with six height maps and one density map from the point clouds of each frame. The authors also proposed a 3D region proposal network based on VGG \cite{simonyan2014very} to predict the position of boxes and the front angle of the detected object.
RVNet \cite{john2019rvnet}, a one-stage object detection network based on the YOLOv3 \cite{redmon2018yolov3}, contains two input branches for radar and camera, and two output branches for small obstacles and big obstacles. Specifically, radar point clouds are transformed into sparse radar images in the image coordinate system via the intrinsic matrix from the camera sensor. Each sparse radar image consists of three channels, namely depth, lateral velocity and longitudinal velocity.
Based on RVNet \cite{john2019rvnet}, SO-Net \cite{john2020so} is presented, focusing on multi-task learning within a single network. In RVNet, the two output branches are modified for vehicle detection and free-space segmentation.
CRF-Net~\cite{nobis2019deep} projects the radar point clouds onto the image plane and feeds the concatenated camera and radar data into a designed VGG-based network. This network enables learning which layer fusion would yield the best benefits by adjusting the weights to radar features on different layers.
In fact, particular objects in camera images tend to remain undetected in night-time scenarios, even when using standard object detection frameworks like YOLOv3 \cite{redmon2018yolov3}. YOdar~\cite{kowol2020yodar} involves lowering the score threshold and assigning radar point clouds to image slices, which are then combined through an aggregated output. Finally, a gradient-boosting classifier is employed to minimize the number of false positive predictions, improving the detection accuracy at night conditions.

In conclusion, projection-based methods in point-based processing leverage radar-to-image projection techniques and various deep learning networks for feature extraction. By leveraging techniques such as radar image generation, multi-stage fusion, and network adaptation, these methods enable robust perception and scene understanding in complex environments, thereby advancing the field of radar-camera fusion networks.

\textbf{Pseudo-Image-based Methods:}
Since image-based CNN networks cannot directly learn original radar point clouds, some studies convert radar point clouds into radar pseudo-images and then utilize image-based methods to extract features.
Based on the distant object detection method in \cite{chadwick2019distant}, Chang~\etal~\cite{chang2020spatial} proposed a radar pseudo-image generation model. Apart from transforming radar point clouds from 3D coordinates into 2D camera coordinates, they also converted the depth, longitudinal velocity and lateral velocity to a real pixel value in RGB channels. Then they introduced a Spatial Attention Fusion-based Fully Convolutional One-Stage (SAF-FCOS) network using a SAF block to merge feature maps derived from radar and camera sensors. In the SAF block, features of radar images are encoded as a spatial attention weight matrix, which is employed to re-weight the feature maps from the image branch. 
SeeingThroughFog~\cite{bijelic2020seeing} introduces a measurement entropy to fuse features from multiple sensors adaptively. Specifically, it applies convolution and sigmoid to the input entropy for a multiplication matrix. The matrix is then utilized to scale the concatenated features from different sensors. This approach adaptively fuses features in the feature extraction stack with the most accurate information.
In CenterFusion~\cite{nabati2021centerfusion}, a novel frustum-based radar association method is proposed to correlate radar detections with preliminary image results. Notably, the authors generated a heat map using depth and radial velocity channels to produce complementary features for the image. After that, they fed the concatenated features into the regression heads to refine the preliminary detection by re-calculating the object's depth, rotation, velocity and attributes. Finally, the results from the regression heads are decoded into 3D bounding boxes.

Overall, pseudo-image-based methods in point-based radar point cloud processing involve the transformation of radar point clouds into radar pseudo-images, which are then processed using image-based techniques. These methods utilize innovative approaches such as spatial attention mechanisms, adaptive feature fusion based on measurement entropy, and frustum-based radar association to enhance the accuracy and robustness of detection results.

\textbf{Voxel-based Methods:}
Apart from projecting radar point clouds onto the camera plane and transforming them into pseudo-images, some researchers extract features directly from 3D radar point clouds to complement the image features. This approach exploits the rich information from the radar point clouds, but requires more sophisticated processing techniques to handle the high-dimensional and unstructured nature of the data.
In GRIF Net \cite{kim2020grif}, an FPN \cite{lin2017feature} and a Sparse Block Network (SBNet) \cite{ren2018sbnet} are used as radar backbones to achieve superior performance with low computational resources. Specifically, in point cloud processing, GRIF Net converts point clouds into voxels. As the point clouds are sparse and most voxels are empty, it leverages SBNet to convolve only on masked areas, avoiding ineffective blank areas. In the fusion module, RoI features from image and radar feature maps are combined by convolutional MoE, demonstrating the effectiveness of radar sensors in detecting vehicles at longer distances than cameras.
In LXL \cite{xiong2023lxl}, the 4D radar branch produces 3D radar occupancy grids that indicate the occupancy status of radar point clouds. These 3D radar occupancy grids are leveraged together with predicted image depth maps to assist in the transformation of image perspective features to the BEV domain. This integration method effectively aligns image features with radar BEV representations, enabling effective fusion with radar features.

Above all, voxel-based methods in point-based radar point cloud processing extract features directly from 3D radar point clouds. These methods utilize techniques such as voxelization, sparse convolution, and occupancy grids to handle the high-dimensional and unstructured nature of radar data. 
By integrating radar features with image features, these methods demonstrate improved performance in detecting vehicles, especially at longer distances and in scenarios where camera data may be limited. The voxel-based approach allows for effective fusion and alignment of information between radar and camera modalities.

\textbf{BEV-based Methods:}
Recently, architectures utilizing BEV representations and transformer networks exhibited impressive performance. 
Simple-BEV~\cite{harley2023simple} focuses on BEV maps from multiple cameras and radars. This method generates a 3D volume with features by projecting 3D coordinates around the ego-vehicle camera images and bilinearly sampling features at projected locations. Later, a BEV feature map is produced by concatenating the 3D features with a rasterized radar image.
CRAFT \cite{kim2023craft} refines image proposals by radar point clouds via a Spatio-Contextual Fusion Transformer (SCFT). The SCFT aims to leverage cross-attention layers to exchange spatial and contextual information in BEV, enabling the fusion network to learn where and what information should be extracted from camera and radar modalities. 
MVFusion \cite{wu2023mvfusion} employs multi-view camera images to obtain semantic-aligned radar features, and subsequently integrates these features in a robust fusion transformer to optimize the cross-modal information interaction. 
CRN \cite{kim2023crn} introduces multi-modal deformable attention to tackle the spatial misalignment between radar and camera feature maps. With its aggregated semantic features and accurate BEV representations, CRN \cite{kim2023crn} currently ranks first among all radar-camera fusion detectors in the nuScenes \cite{caesar2020nuscenes} dataset, being the best approach in 3D radar-camera fusion.
RCFusion \cite{zheng2023rcfusion} achieves multi-modal feature fusion under a unified BEV perspective with the input of 4D radar and camera. The Interactive Attention Module (IAM), a key component of RCFusion, is utilized to weight the features of each modality, thus fully exploiting the advantages of both modalities.

In summary, BEV-based methods in point-based radar point cloud processing leverage BEV representations and transformer networks to achieve impressive performance in radar-camera fusion. These methods incorporate techniques such as refining proposals, cross-modal information interaction, semantic alignment, and attention mechanisms to optimize feature extraction and fusion between radar and camera data. With the ability to handle spatial misalignment and exploit the advantages of top-down perspective, BEV-based methods demonstrate high performance in 3D radar-camera fusion tasks.


\subsubsection{Tensor-based Networks}
Due to the potential loss of crucial information concerning objects or the surrounding environment during the processing of radar point clouds after CFAR detection, several researchers have put forward a fusion scheme that involves the fusion of radar tensors with camera images. We categorize these tensor-based networks into cross-supervised-based methods and projection-based methods.

\textbf{Cross-Supervised-based Methods:}
For radar data in the format of tensors, it is challenging to label the radar data as they are not spatially consistent compared to image data. Thus, some researchers propose cross-modal supervision methods to generate radar labels with the supervision of camera images. 
Specifically, RODNet \cite{wang2021rodnet, wang2021rethinking} is a radar object detection network using a camera-radar fusion strategy to cross-supervised 3D localization of detected objects during the training stage. It takes sequences of RA tensors as input and uses a neural network-based approach to extract the Doppler information. Specifically, to handle multi-chirp merging information and dynamic object motion, RODNet introduces two customized modules, namely M-Net and Temporal Deformable Convolution (TDC).
Moreover, Gao~\etal~\cite{gao2020ramp} fed sequences of RD tensor, RA tensor and AD tensor into convolutional autoencoders. They proposed a Radar Multiple-Perspectives Convolutional Neural Network (RAMP-CNN) that utilizes the temporal information in the chirps within a single frame, along with the change in spatial information between frames. In RAMP-CNN, features of these three tensors are then fused in a fusion module to generate new range-azimuth features. Compared to RODNet \cite{wang2021rodnet, wang2021rethinking}, RAMP-CNN achieves significant performance and maintains the same detection accuracy in night scenes as in the daytime.
Recently, Jin~\etal~\cite{jin2023cross} utilized the segmented camera image with radar customized adaption as the ground truth for training deep neural networks to perform panoptic segmentation on radar data. The proposed network utilizes panoptic segmentation to achieve radar-tailored sensing, including free-space segmentation and object detection, with only radar RA tensor in urban, rural, and highway scenarios. 

In conclusion, cross-supervised-based methods enable radar data to benefit from the rich spatial information available in camera data. Techniques like RODNet, RAMP-CNN, and cross-supervised panoptic segmentation have demonstrated the effectiveness of incorporating multi-modal supervision techniques to enhance object detection and segmentation tasks, thus showing performance improvements in handling the unique characteristics and challenges of radar data.


\textbf{Projection-based Methods:}
FusionNet~\cite{lim2019radar} converts both radar RA tensors and camera images into Cartesian coordinates, followed by projecting camera images onto the radar plane using a homography transformation. Upon passing through the independent feature extractor branches, features of the two modalities are passed through the additional fusion layers to form a unified feature map.
As it is challenging to fuse radar tensors and camera images in 3D coordinates, Hwang~\etal~\cite{hwang2022cramnet} proposed a radar-camera matching network named CramNet. CramNet overcomes the uncertainties in the geometric correspondences between the camera and radar through a ray-constrained cross-attention mechanism. Specifically, since a peak in the radar returns usually accompanies the optimal 3D position corresponding to the foreground pixel of an image, CramNet projects radar features along the pixel rays to estimate the depth and refine the 3D locations of camera pixels. Experiments on the RADIATE \cite{sheeny2021radiate} dataset demonstrate that the CramNet outperforms the baseline results from the Faster R-CNN \cite{ren2015faster} detector. Additionally, by conducting experiments on the filtering of RA tensors via varying intensity thresholds, radar RA tensors prove to contain more meaningful information for 3D object detection than sparse point clouds.

To sum up, projection-based fusion methods, such as FusionNet and CramNet, offer practical solutions for integrating radar and camera data by leveraging geometric transformations and novel attention mechanisms. These methods contribute to advancing the integration of multi-modal information and demonstrate promising results in object detection tasks, highlighting the significance of leveraging radar tensors in perception systems.

\subsection{Model Evaluations}

\subsubsection{Evaluation Metrics}
Various evaluation metrics are adopted or newly proposed to evaluate the performance of radar-camera fusion models, as summarized in Table \ref{tab:evaluation metrics}.
Similar to image-based object detection and semantic segmentation tasks, in radar-camera fusion, commonly used evaluation metrics are precision, recall, Average Precision (AP), Average Recall (AR), mean Average Precision (mAP) and mean Intersection over Union (mIoU). 
However, these metrics only calculate the prediction accuracy on a given test dataset. Attributes in multi-modal datasets, such as velocity, range, size and orientation, are ignored. Besides, for multi-modal networks, the IoU thresholds should depend on object distance and occlusion, as well as the type of sensors \cite{feng2020deep}.

To overcome these drawbacks, the nuScenes ~\cite{caesar2020nuscenes} dataset introduces mATE, mASE, mAOE, mAVE and mAAE, which stand for mean average translation, scale, orientation, velocity and attribute errors, respectively. Furthermore, they presented nuScenes Detection Score (NDS), half based on the mAP, half quantifying the previous five metrics.
To evaluate how well a detection result matches the ground truth, Wang~\etal~\cite{wang2021rodnet} defined Object Location Similarity (OLS) that quantifies the correlation between two detections concerning their distance, classes and scale information.
Additionally, some metrics designed for LiDAR object detection are also adopted in radar point clouds. For example, Cui~\etal~\cite{cui20213d} utilized Average Heading Similarity (AHS) to calculate the accuracy, which is formulated initially to calculate the average orientation angle in 3D LiDAR IoU defined in AVOD \cite{ku2018joint}.

\subsubsection{Performance Evaluation}
\begin{figure}[ht]
\begin{center}
\includegraphics[width=1\linewidth]{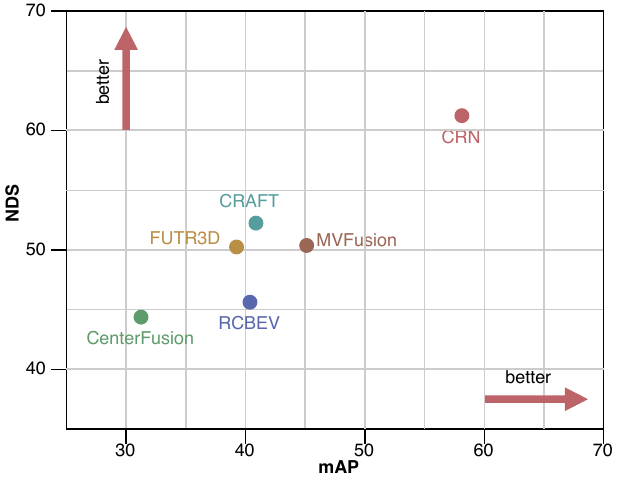}
\end{center}
   \caption{Performance of radar-camera fusion methods on the nuScenes \textit{test} set. The horizontal and vertical axes are mAP and NDS, respectively, and the larger their values, the better the performance.
   mAP and NDS are the metrics for evaluating the horizontal and vertical axes, respectively, where greater values indicate better performance.}
\label{fig:performance}
\end{figure}

Given that the majority of researchers have employed the nuScenes \cite{caesar2020nuscenes}, VoD \cite{palffy2022multi} and TJ4DRadSet \cite{zheng2022tj4dradset} datasets to evaluate the performance of their algorithms, we provide a comprehensive summary of the evaluation metrics and performance outcomes on these dataset in Table \ref{tab:evaluation}.
We also provide Fig. \ref{fig:performance} to clearly show the performance comparison of radar-camera fusion methods using the complete nuScenes dataset.
CenterFusion~\cite{nabati2021centerfusion}, the first radar-camera fusion algorithm to operate on the complete nuScenes dataset, achieves a performance outcome of 32.6\% on the mAP and 44.9\% on the NDS. CenterFusion solved the critical data association problem in radar-camera fusion by proposing a novel frustum-based radar association method, which generates a RoI frustum around objects in 3D space using preliminary detection results, and maps the radar detection to the center of objects on the image. In comparison to CenterNet \cite{duan2019centernet}, which solely relies on image input, CenterFusion delivers a relative increase of 38.1\% and 62.1\% on the NDS and velocity error metrics, respectively, demonstrating the effectiveness of using radar features and robustness of radar-camera fusion in challenging environments.

After that, numerous radar-camera fusion algorithms employ CenterFusion as the baseline. 
For example, RCBEV \cite{zhou2023bridging}, a feature-level fusion approach, extracts radar features using a temporal-spatial encoder and transforms image features into BEV representations. Experimental results demonstrate superior feature representation and more accurate 3D object detection outcomes, receiving the mAP and NDS of 40.6\% and 48.6\%, respectively.
CRAFT \cite{kim2023craft} achieves 41.1\% mAP and 52.3\% NDS on the nuScenes, where most of the gain in performance originates from the improved localization and velocity estimation with the assistance of the spatio-contextual fusion transformer. This transformer exploits both spatial and contextual properties of camera and radar data to detect objects in 3D space more accurately. 
CRN \cite{kim2023crn} currently emerges as the top-performing detector among all radar-camera fusion algorithms on the nuScenes dataset with 57.5\% mAP and 62.4\% NDS, being the best approach in 3D radar-camera fusion. The performance gain of the proposed CRN framework comes from its Radar-assisted View Transformation (RVT), which overcomes the lack of spatial information in an image and transforms perspective view image features to BEV with the help of sparse but accurate radar points. The transformed image features in BEV are then used in the Multi-modal Feature Aggregation (MFA) layers to generate a semantically rich and spatially accurate BEV representation. 
For 4D radar-camera fusion, LXL \cite{xiong2023lxl} has 56.31\% mAP and 36.32\% mAP, ranks first in the VoD and TJ4DRadSet dataset, respectively, being the best approach for 4D radar-camera fusion.

In general, substantial progress has been made by various algorithms operating on radar-camera fusion datasets. The mAP has improved by 24.9\%, and the NDS has increased by 17.5\% on the complete nuScenes dataset. On the other hand, the mAP has improved by 18.31\% on the VoD dataset. The incorporation of transformer architectures, attention mechanisms, and BEV features are crucial factors that have significantly contributed to enhancing performance outcomes.

\subsection{Summary}

Above all, we present the methodologies of radar-camera fusion related to object detection and semantic segmentation tasks. Through in-depth analysis of the five questions revolving around ``why to fuse", ``what to fuse", ``where to fuse", ``when to fuse" and ``how to fuse", we gain insight into the positive benefits of radar-camera fusion when applied on autonomous driving vehicles, including improved accuracy, robustness and redundancy.  
As indicated by the summary information in Table \ref{tab:fusion methods}, the number of methods designed for semantic segmentation tasks is fewer than that for the object detection task.
On the one hand, as enumerated in Section \ref{Dataset Tasks} and Table \ref{tab:datasets}, there are fewer public datasets with segmentation annotations than those with detection annotations available for radar-camera fusion.
On the other hand, the intrinsic characteristics of radar data render it more suitable for detection tasks. For example, the capacity for long-distance detection and velocity measurement confers a distinct advantage upon radar data for the effective detection of moving obstacles \cite{chadwick2019distant}. Conversely, the sparse and noisy point cloud structure presents significant limitations for semantic segmentation tasks \cite{schumann2018semantic}.

It is important to note that although radar is not conventionally used for semantic segmentation tasks, it can effectively improve the accuracy and reliability of the fused data \cite{bansal2022radsegnet}. By providing complementary information, such as depth information, radar data can enhance the performance of semantic segmentation algorithms. Also, radar data can be employed to localize objects in 3D space, which can validate the 2D output generated by semantic segmentation algorithms. However, simple fusion operations of image and radar (e.g., concatenation) for semantic segmentation can distort the semantic structure of the image, essentially adding noise to the image. As a result, this can detrimentally impact convergence rates, causing slow and suboptimal learning of the segmentation model.

%
%
%


\section{Challenges and Research Directions} \label{Challenges}

It is a challenging problem to balance the performance of different modalities so that all of them can perform at their best level and thus improve the overall performance. 
As described in Section \ref{Introduction}, radar-camera fusion faces numerous challenges. 
If these challenges are not properly addressed in sensor perception, they may also affect the subsequent tasks such as localization, prediction, planning and control.
In this section, we focus on improving the accuracy and robustness of radar-camera fusion, with discussions on the critical challenges and possible research directions from two aspects: multi-modal data and multi-modal fusion.

\begin{table*}[htbp]
\caption{Overview of challenges and research directions.}
\setlength\tabcolsep{5pt}
\footnotesize
\begin{tabular*}{1\linewidth}{
p{0.6cm}<{\centering}
p{2.45cm}<{}
p{5.5cm}<{\centering}
p{8.2cm}<{\centering}
}
\toprule
\centering
\bf{Topic} & \bf{Sub Topic} & \bf{Challenges} & \bf{Research Directions} \\\midrule
\multirow{14}{*}{\rotatebox{90}{Multi-modal Data}} 
& Data Quality & 
\begin{minipage}[t]{0.3\textwidth}
\begin{itemize}
  \item Sparsity
  \item Inaccuracy
  \item Noise
\end{itemize}
\end{minipage}
&
\begin{minipage}[t]{0.55\textwidth}
\begin{itemize}
  \item Leveraging multiple radar frames to enhance the density
  \item Applying 4D radar sensors with higher resolution
  \item Studying distribution of point clouds (e.g., Gaussian distribution)
  \item Denoising based on physical characteristics of radar data
\end{itemize}
\end{minipage}
\\\cmidrule{2-4}

& Data Diversity 
&
\begin{minipage}[t]{0.3\textwidth} 
\begin{itemize}
  \item Small Size
  \item Insufficient Conditions
\end{itemize} 
\end{minipage}
& 
\begin{minipage}[t]{0.45\textwidth}
\begin{itemize}
  \item Collecting data from adverse conditions
  \item Integrating synthetic data with real-world data
  \item Domain adaptation for model generalization with limited data
\end{itemize}
\end{minipage}
\\\cmidrule{2-4}

& Data Synchronization &
\begin{minipage}[t]{0.3\textwidth}
\begin{itemize}
  \item High Calibration Requirements
  \item Difficulties in Labeling
\end{itemize}
\end{minipage} 
&
\begin{minipage}[t]{0.45\textwidth}
\begin{itemize}
  \item 4D radar-camera calibration
  \item Real-time online calibration and correction
  \item Auto-labeling to reduce manual labeling
  \item Improving labeling efficiency via active learning, domain adaptation, transfer learning, semi-supervised learning
\end{itemize}
\end{minipage}\\
\midrule

\multirow{26}{*}{\rotatebox{90}{Multi-modal Fusion}} 
& Feature Extraction &

\begin{minipage}[t]{0.3\textwidth} 
\begin{itemize}
  \item Less effective using LiDAR-based or image-based algorithms
\end{itemize}
\end{minipage}
&
\begin{minipage}[t]{0.45\textwidth}
\begin{itemize}
  \item Introducing attention mechanisms
  \item Using Graph Neural Networks to dig deeper into the relationship between sparse point clouds
  \item Applying neural networks to extract radar information instead of traditional FFT operations
\end{itemize}  
\end{minipage}
\\\cmidrule{2-4}

& Data Association &
\begin{minipage}[t]{0.3\textwidth} 
\begin{itemize}
  \item Ambiguity in associating radar data with image data
  \item Poor association using calibration matrix 
\end{itemize} 
\end{minipage}
& 
\begin{minipage}[t]{0.45\textwidth}
\begin{itemize}
  \item Attention-based association with adaptive thresholds
  \item Joint state estimation
  \item Uncertainty estimation of object tracks
\end{itemize} 
\end{minipage} 
\\\cmidrule{2-4}

& Data Augmentation &
\begin{minipage}[t]{0.3\textwidth} 
\begin{itemize}
  \item Correlation and interdependence between radar and camera modalities
\end{itemize} 
\end{minipage}
& 
\begin{minipage}[t]{0.45\textwidth}
\begin{itemize}
  \item Joint data augmentation rather than augmenting each modality separately
\end{itemize}  
\end{minipage}
\\\cmidrule{2-4}

& Training Strategies &
\begin{minipage}[t]{0.3\textwidth} 
\begin{itemize}
  \item Difficulties in training
  \item Overfitting for multi-modal model
\end{itemize} 
\end{minipage}
& 
\begin{minipage}[t]{0.45\textwidth}
\begin{itemize}
  \item Weighting operations on loss functions
  \item Weights freezing strategies
  \item Knowledge distillation on uni-modal features for the multi-modal networks
\end{itemize} 
\end{minipage} 
\\\cmidrule{2-4}

& Model Robustness &
\begin{minipage}[t]{0.3\textwidth} 
\begin{itemize}
  \item Sensor degradation or failure in adverse conditions 
  \item Unseen driving scenarios
\end{itemize}
\end{minipage} 
& 
\begin{minipage}[t]{0.45\textwidth}
\begin{itemize}
  \item Attention mechanisms
  \item Uncertainty estimation
  \item Generative models for sensor defects or new scenarios
\end{itemize}  
\end{minipage}
\\\cmidrule{2-4}

& Model Evaluations &
\begin{minipage}[t]{0.3\textwidth} 
\begin{itemize}
  \item Different selected sub-dataset
  \item Unknown objects in the open world
\end{itemize} 
\end{minipage}
& 
\begin{minipage}[t]{0.45\textwidth}
\begin{itemize}
  \item Metrics related to uncertainties
  \item Developing visual toolkits for analyzing and optimizing fusion networks
\end{itemize}  
\end{minipage}
\\\cmidrule{2-4}

& Model Deployment &
\begin{minipage}[t]{0.3\textwidth} 
\begin{itemize}
  \item Edge devices with limited computational resources
  \item Balancing the importance of different tasks
\end{itemize} 
\end{minipage}
& 
\begin{minipage}[t]{0.45\textwidth} 
\begin{itemize}
  \item Lightweight models and acceleration schemes (e.g., pruning and quantization)
  \item Fusion-based multi-task perception and panoptic perception
\end{itemize} 
\end{minipage} 
\\\bottomrule

\end{tabular*}
\vspace{1mm}
\label{tab:challenges}
\end{table*}

\subsection{Multi-modal Data}

\subsubsection{Data Quality}

Unlike uni-modal data, multi-modal data requires consideration of each modality's native characteristics.
The information within an image is structured and regular, with partial information being associated with the whole image. In contrast, the spatial information embodied in radar point clouds tends to be disordered.
As a result, handling radar data poses a more significant challenge in the context of radar-camera fusion. We categorize these challenges related to data quality into three directions: sparsity, inaccuracy, and noise. 


\paragraph{Sparsity}
The sparsity of radar point clouds poses a challenge for neural networks to learn features effectively. Besides, as these point clouds do not comprehensively represent an object's shape, a fixed-size bounding box approach would be impractical.
To address the issue of sparsity, researchers usually combine multiple frames (from 0.25 seconds to 1 second) of radar data to get denser point clouds, making it conducive to improving accuracy \cite{schumann2018semantic, nobis2019deep, nabati2021centerfusion, kim2020grif, long2021radar, palffy2022multi, harley2023simple}. However, the time-dependent approach also causes system delays.
Nowadays, the 4D radar sensor is a potential research direction as it produces denser point clouds, reaching hundreds of points on a car. The spatial distribution of objects is effectively represented in 4D radar datasets, including Astyx \cite{meyer2019deep}, VoD \cite{palffy2022multi} and TJ4DRadSet \cite{zheng2022tj4dradset}. 
Experiments also indicate that the 4D radar is helpful in radar detection. For example, Zheng~\etal~\cite{zheng2022tj4dradset} proved that 4D radar has potential in 3D perception as the points get dense. Palffy~\etal~\cite{palffy2022multi} pointed out that the additional elevation data increases object detection performance (from 31.9\% to 38.0\% in mAP) in their VoD \cite{palffy2022multi} dataset.

\paragraph{Inaccuracy}

Aside from the sparsity of radar point clouds, the points may not be located at the object's center, but may be at any corner of the object or even outside of it \cite{meyer2019deep}.
To make the radar points located on or close to the object, Chadwick~\etal~\cite{chadwick2019distant} marked each radar point on the image as a small circle instead of a single pixel. Nabati and Qi~\cite{nabati2019rrpn} proposed RRPN to generate several anchors with different sizes and aspect ratios centered at the points of interest. These translated anchors are employed to achieve more precise bounding boxes where the points of interest are mapped to the object's right, left, or bottom.

Researchers have employed column and pillar expansion techniques to improve the accuracy of radar point clouds in the vertical dimension. 
For example, Nobis~\etal~\cite{nobis2019deep} assumed a height extension of three meters on each radar detection to associate camera pixels with radar data.
Nabati and Qi~\cite{nabati2021centerfusion} used pillar expansion to expand each radar point to a fixed-size pillar. In their experiment, the size of a pillar is set to $[0.2, 0.2, 1.5]$ meters along the $[x, y, z]$ directions. 
Notably, the column size should be different for different types of objects.
In \cite{li2020feature}, the authors utilized a clustering approach to group ground truth bounding boxes of vehicles into three distinct height categories. Following this, radar points are assigned a scale based on boundary edge values in each category.

In our opinion, column or pillar expansion is effective but still hardly convincing. The distribution of point clouds is a direction worth investigating.
For example, St\"{a}cker~\etal~\cite{stacker2022fusion} assumed a Gaussian distribution to measure the azimuth angle. According to the resulting Gaussian density curve, they generated denser radar point clouds, horizontally distributed over multiple pixels.

\paragraph{Noise}

Actually, the radar sensor returns noisy data from irrelevant objects, including ghost objects, ground detections and even multi-radar mutual interference \cite{lu2020see}. The noise from radars could cause the detection of fake targets, thereby limiting the accuracy of radar-based detection or segmentation. 

Conventional methods \cite{wang2021cfar, zhang2018support, jin2019automotive, alhumaidi2021interference} for automotive radar denoising are typically based on CFAR and peak detection algorithms, which exhibit poor generalization capabilities. Recently, deep learning methods have provided a key solution to the challenges associated with automotive radar data denoising. 
In \cite{schussler2022deep}, a deep neural network is proposed to enhance the target peaks on RA tensors. Rock~\etal~\cite{rock2021resource} analyzed the quantization of CNN-based denoising autoencoder for radar interference mitigation on radar RD tensors to guarantee real-time inference on low-performance equipment. 
Dubey~\etal~\cite{dubey2020region} realized the multi-radar mutual interference and object detection on RD tensors simultaneously with one one-stage CNN-based neural network. 
Moreover, in \cite{ristea2020fully, fuchs2020automotive, chen2021dnn, chen2022two, de2020deep}, fully-convolution networks are widely proposed and applied to interference mitigation on RD tensors.

However, conventional radars generally do not provide access to the radar tensor, highlighting the importance of noise mitigation techniques at the level of radar point clouds. 
For example, Nobis~\etal~\cite{nobis2019deep} designed a ground-truth filter to remove radar detections outside of the 3D ground truth bounding boxes. Cheng~\etal~\cite{cheng2021new} proposed a cross-modal radar point detector through the assistance of LiDAR, which could also remove the noisy points. 
Essentially, noise mitigation at the point cloud level is semantic segmentation of the point cloud, whereby a semantic segmentation model is leveraged to assign a category to each point. 
Numerous studies focus on point cloud segmentation, including PointMLP \cite{ma2022rethinking}, PointNeXt \cite{qian2022pointnext}, Point Transformer \cite{zhao2021point} and Point Cloud Transformer \cite{guo2021pct}. Notably, Point Transformer and Point Cloud Transformer introduce the self-attention mechanism within their point cloud processing networks to capture contextual features. However, it is worth noting that these approaches still rely on the advantages provided by two essential modules derived from PointNet++: Set Abstraction (SA) and Multi-Scale Grouping (MSG).
Recently, a revolutionary non-parametric point cloud segmentation model called Point-NN has been proposed. Point-NN \cite{zhang2023parameter} elegantly assembles farthest point sampling, K nearest neighbor, pooling, trigonometric position encodings and similarity measurement, thereby achieving SOTA performances on several benchmarks with superior performance to any other point cloud processing models.

In all, regardless of the stage of denoising in radar tensors or point clouds, a dataset with high-quality annotations is necessary. For the noise removal on radar tensors, researchers may adopt image denoising and restoration principles. For removing noisy point clouds, constructing features based on the physical characteristics of point clouds to guide models separate targets from clutters makes sense. As radar point clouds are sparse and inaccurate, modeling the inherent uncertainty is an open question, which can aid in effectively distinguishing targets from noise.

\subsubsection{Data Diversity}


\paragraph{Small Size}
Deep learning models rely on large amounts of training data to achieve high levels of accuracy.
However, multi-modal datasets consisting of both radar and camera data are much smaller than uni-modal image data. For instance, compared to the ImageNet \cite {deng2009imagenet, russakovsky2015imagenet} dataset with over 14 million images and over 20k classes, the largest radar-camera fusion dataset to date named CRUW \cite{wang2021rethinking} has only 400k frames and 260k objects.
Furthermore, regarding category distribution, most labels are vehicles, while pedestrians and bicycles are far less prevalent. The imbalance in these categories' distribution may result in overfitting designed deep learning networks \cite{nobis2021kernel}.

\paragraph{Insufficient Conditions}
In real scenarios, 360-degree perception of the surrounding environment is critical in autonomous driving, requiring multiple cameras and radars to work together. Besides, the multi-modal dataset also needs to consider complex weather conditions (e.g., rain, fog, snow) and complex road conditions (e.g., blocked roads, rural paths, intersections), all of which are time-consuming and labor-consuming tasks.

Some studies \cite{weng2023all, guan2020through} generate synthetic data via simulation tools (e.g., Carla \cite{dosovitskiy2017carla}). 
Researchers can freely match different sensors and generate different driving conditions with these tools, especially in complex and dangerous scenarios. 
However, it is also worth noting that although simulators can generate a variety of virtual datasets, the simulated data cannot completely replace the data from real scenarios. Moreover, exploring the appropriate methodology for integrating synthetic data with real-world data is a critical area of inquiry warranting further investigation \cite{nowruzi2019much}.

Domain adaptation is also a valuable research direction that aims to leverage knowledge learned from a related domain with adequate labeled data. Although domain adaptation has been applied in radar data, including radar tensor reconstruction \cite{stephan2021radar}, human sensing \cite{li2022unsupervised}, human activity recognition \cite{li2021semisupervised} and gesture recognition \cite{li2022towards}, it has not been employed in radar-camera fusion till now. 

\subsubsection{Data Synchronization}


\paragraph{High Calibration Requirements}

For radar-camera fusion systems, well-calibrated sensors are the prerequisite.
In multi-sensor calibration, LiDAR sensors are typically employed as an essential intermediary component. The LiDAR sensor is calibrated separately from the camera sensor and radar sensor, and then a transformation matrix between the radar and camera can be calculated \cite{zheng2022tj4dradset, burnett2023boreas}. 
Although numerous approaches (e.g., \cite{domhof2021joint, cui20213d, zhang20223dradar2thermalcalib}) are proposed for calibration between radars and the cameras, the accuracy of the calibration remains a challenge due to the inaccurate and vulnerable radar returns.
Besides, as far as we know, \cite{cui20213d} and \cite{paek2022k} are the only methods for 4D radar and camera calibration. As 4D radar technologies are developing rapidly, we believe that 4D radar calibration is a potential direction, and more finds will be proposed in the future.

In real scenarios, extrinsic calibration parameters between radar and camera sensors may change from vehicle vibration. Besides, different sampling frequencies of the radar and camera may produce a particular temporal difference between the data from each sensor. The temporal difference would cause data inconsistency, especially when the ego-car or targets move at high speed. Therefore, real-time online calibration and correction are essential research directions in the future.


\paragraph{Difficulties in Labeling}

The process of labeling data is labor-intensive and time-consuming, especially when dealing with multi-modal data. This is particularly true for radar-camera fusion, where the physical shapes of objects cannot be discerned directly from the radar data representation. 
Auto-labeling radar data is a potential research direction to address the challenge of laborious data labeling. Actually, labels for radar data can be calculated based on the corresponding ground truth from camera images and the extrinsic matrix between the radar sensor and the camera sensor. But the problem is that applying this labeling approach for radar data is not perfect, as radar targets may not always be located in the ground truth from images.
Sengupta~\etal~\cite{sengupta2022automatic} proposed a camera-aided method for automatically labeling radar point clouds, leveraging a pre-trained YOLOv3 \cite{redmon2018yolov3} network and the Hungarian algorithm for enhanced accuracy and efficiency. 
However, despite the potential advantages of auto-labeling radar data, filtering out noisy data around the object of interest is still challenging.

For camera image labeling, it is worth considering how to select appropriate labeling data for reducing labor costs.
Active learning is a supervised learning method that aims to select the smallest possible training set to achieve the desired data efficiency \cite{ren2021survey, zhan2022comparative}. The active learning network iteratively queries the most informative samples from the human labelers in an unlabeled data pool and then updates the weights for the network. This approach leads to equivalent performance with less labeled training data, reducing human labeling efforts.
Experiments from \cite{winterling2017optimizing} indicate that using only about 40\% of the data in the training set leads to the same classification results as the completely supervised reference experiment. 
Furthermore, many other methods would also be used to reduce the burden of data labeling, such as domain adaptation \cite{patel2015visual, chen2019learning, stephan2021radar}, transfer learning \cite{pan2009survey}, semi-supervised learning \cite{kingma2014semi} and lifelong learning \cite{parisi2019continual}. 

\subsection{Multi-modal Fusion}




\subsubsection{Feature Extraction}


Applying LiDAR-based feature extraction algorithms to radar modality is less effective due to the inherent sparsity of radar point clouds.
As an example, PointPillars \cite{lang2019pointpillars} algorithm converts LiDAR point clouds into pillars and then extracts features from each pillar. When this algorithm is adapted to radar point clouds, there may be few or even no points in a pillar, which makes it hard to extract features. In fact, results in \cite{scheiner2020seeing} and \cite{scheiner2021object} also indicate that the average precision of radar point clouds using PointPillars \cite{lang2019pointpillars} is much lower than using SSD \cite{liu2016ssd} and YOLOv3 \cite{redmon2018yolov3} detectors.

As PointPillars \cite{lang2019pointpillars} focus on local features, the attention mechanism is a potential research direction to extract global features to improve accuracy. For example, Radar Transformer \cite{bai2021radar} incorporates both vector attention and scalar attention mechanisms to effectively leverage spatial information, Doppler information, and reflection intensity information from radar point clouds. By integrating local attention features and global attention features, Radar Transformer achieves deep integration of radar information.
RPFA-Net \cite{xu2021rpfa} leverages the self-attention mechanism to extract global features (e.g., orientation) from point clouds. These global features enable the network to perform more efficient and effective regression of key object parameters (e.g., heading angle), thereby enhancing the accuracy and reliability of object detection. 
In addition, Gaussian Radar Transformer \cite{zeller2022gaussian} employs attentive upsampling and downsampling modules to enlarge the receptive field and capture distinctive spatial correlations, effectively addressing the challenge of capturing long-range dependencies in radar data. 
The attention-based techniques and multi-task learning used in HARadNet \cite{dubey2022haradnet} also lead to a significant performance improvement in the classification and detection.

To dig deeper into the relationship between sparse point clouds, GNN \cite{scarselli2008graph} is a promising research direction in which each point is considered as a node, and edges are the relationship between the points.
In Radar-PointGNN \cite{svenningsson2021radar}, GNN adopted for feature extraction of radar point clouds demonstrates that the graph representation produces more effective object proposals than other point cloud encoders by mapping radar point clouds to contextual representations. RadarGNN \cite{fent2023radargnn} indicates that GNNs can operate on unstructured and unordered data, obtaining both point features and point-pair features embedded in the edges of the graph. Thus, compared to voxelization operations, GNN eliminates the information loss from the sparse radar point clouds.
GNN also shows its advantages in detection from RA tensors. The Graph Tensor Radar Network (GTR-Net) \cite{meyer2021graph} architecture utilizes graph convolutional operations to aggregate information across the point cloud nodes. The process involves weighting the features of connected nodes based on their respective edge weights. In this way, it improves the defective sparse points by aggregating relevant information and thus leads to better performance.

Another potential research opportunity is using neural networks to extract radar information instead of traditional FFT operations, which can reduce the computational requirements that consume most of the operations and simplify the data flow in the embedded implementation.
For example, in RODNet \cite{wang2021rodnet}, FFT operations are only performed in sample and antenna dimensions, while the chirp dimension remains to get the range-azimuth-chirp tensor. Then a neural network is employed to process the chirp dimension for extracting Doppler features, enabling end-to-end training of radar features in-depth within the deep learning framework.

\subsubsection{Data Association}


Another significant challenge is the ambiguity in associating radar data with image data, as they are heterogeneous. The typical way is to project radar data onto the image plane and then bind the data in the same position through a calibration matrix \cite{bai2021robust}.
However, direct projection results in poor association with the objects' centers. As aforementioned, radar data is sparse, inaccurate and noisy, making poor association at either the object-level or data-level fusion. 

Thus, associating image data with radar data is an open question. 
Nabati and Qi~\cite{nabati2020radar} proposed a Radar Proposal Refinement (RPR) network to match proposals from radars and cameras.
Later, they integrated the detection boxes and pillar expansion through frustum association in CenterFusion \cite{nabati2021centerfusion}, allowing for the mapping of radar detections to the centers of objects and mitigating the issue of overlapping.
Dong~\etal~\cite{dong2021radar} used AssociationNet to learn the semantic representation information and associate radar point clouds and image bounding boxes by Euclidean distance.
For associating the semantics to radar point clouds, Bansal~\etal~\cite{bansal2022radsegnet} proposed a representation named Semantic-Point-Grid (SPG), which encodes semantic information from camera images into radar point clouds to identify camera pixel correspondences for each radar point.

In our opinion, a potential approach to associate radar data with image data is the attention-based association with adaptive thresholds. For example, Radar-Camera Pixel Depth Association (RC-PDA) is proposed to filter out occluded radar returns and enhance the projected radar depth map by generating a confidence measure for these associations in \cite{long2021radar}. 
Soft Polar Association (SPA) is proposed to associate radar point clouds around the image proposals in polar coordinates \cite{kim2023craft}. In order to overcome background clutter, it utilized consecutive cross attention-based encoder layers to integrate image proposal features and radar point features.

\subsubsection{Data Augmentation}

Numerous data augmentation methods have been proposed to increase the quantity and diversity of data samples, thus preventing network overfitting and enhancing model generalization.
For radar data in the form of point clouds, data augmentations such as random rotation, scaling, and flipping shifting are applied to enrich the diversity of samples in \cite{nabati2021centerfusion, zheng2022tj4dradset, zhou2023bridging}.
In addition, since radar tensors can be treated as images, existing image-based data augmentation algorithms (e.g., horizontal flipping, translating in range, interpolating, mixing) are tested in experiments and proved to be effective \cite{gao2020ramp}. 

However, all these data augmentation methods above are based only on the radar modality. In radar-camera fusion perception, designing effective data augmentation methods needs to consider the correlation and interdependence between radar and camera modalities, which means joint data augmentation methods are necessary rather than augmenting each modality separately. Otherwise, the model will learn from incorrect data, whose physical properties are unreliable. For example, when the coordinates of radar data and image are aligned, applying Cutmix \cite{yun2019cutmix} on the image and radar feature maps will undoubtedly destroy the target features (e.g., azimuth and elevation) obtained by the radar sensor, leading to incorrect model inferences. Therefore, designing joint data augmentation algorithms for the unique radar representations combined with image modality remains a significant challenge.

\subsubsection{Training Strategies}


Since a multi-modal network has additional input information, it should match or outperform the uni-modal network. However, this is not always the case. A multi-modal network is often prone to overfitting and tends to learn to ignore one branch if the hyper-parameters set for training are more suitable for the other branch.
Wang~\etal~\cite{wang2020makes} argued that the rates of overfitting and generalization vary across different modalities, and training a multi-modal network using the uni-modal training strategy may not be optimal for the overall network. 

A feasible approach to balance the performance is to add loss functions for each modality. In this way, after one modality converges, the remaining modality can still be generalized.
Besides, weighting operations on loss functions could be more beneficial to adapt to the learning rates of each modality. 
In recent studies, Wang~\etal~\cite{wang2020makes} proposed Gradient-Blending, which computes an optimal blending of modalities based on their overfitting behaviors. Although this method achieves SOTA accuracy on audio and vision benchmarks, the idea has yet to be applied in radar and camera modalities.
Moreover, dropout operation helps overcome overfitting. Nobis~\etal~\cite{nobis2019deep} introduced BlackIn by deactivating camera image data. The lack of camera input data forces the network to rely more on sparse radar data for specific potential values. 

Fine-tuning a multi-modal network over pre-trained uni-modal encoders can also outperform fusion from scratch. Lim~\etal~\cite{lim2019radar} utilized the weights freezing strategy to train a single branch network using the optimal training hyper-parameters. These weights were subsequently loaded into the corresponding branches to train the fusion network. Experimental results indicate that the best strategy is to train the camera branch in advance and then train the entire network with the gradient propagation disabled through the camera branch.
Recently, knowledge distillation has shown its performance in multi-modal networks by distilling the pre-trained uni-modal features to the multi-modal networks \cite{dou2020unpaired, du2021improving}. It could also be a potential research direction in radar-camera fusion.

\subsubsection{Model Robustness}


Another challenge is how to guarantee the model's robustness when the sensors are degraded, or the autonomous driving vehicles enter into adverse or unseen driving scenarios.
Most reviewed methods focus on the accuracy of public datasets, while only a few consider sensor failure with only one modality as the input data.
In RadSegNet \cite{bansal2022radsegnet}, the SPG encoding independently extracts information from cameras and radar, as well as encodes semantic information from camera images into radar point clouds. Thus in scenarios where the camera input becomes unreliable, the SPG encoding method maintains reliable operation using radar data alone.
Bijelic~\etal~\cite{bijelic2020seeing} introduced an entropy channel for each sensor stream and a feature fusion architecture to exchange features, which still work in unseen weather conditions and sensor failures.
Moreover, the attention mechanism is also an effective choice for guiding mixed information from different sensors to fuse features of multiple modalities, as well as handle original features from a single modality. For example, attention maps leverage features learned from different sensors to predict the importance of specific parameters in \cite{zhang2019robust}.

It is essential to focus not only on the accuracy of the predicted outcomes, but also on the degree of certainty the model has about them. 
Uncertainty is a potential direction that can be used to handle unseen driving scenarios. Specifically, a multi-modal network should present higher uncertainty against unseen objects. The Bayesian neural network is a valuable choice for calculating uncertainty. It utilizes a prior distribution of network weights to infer the posterior distribution, thereby estimating the probability associated with a given prediction \cite{feng2020deep, mackay1992practical}.
In radar-camera fusion, YOdar \cite{kowol2020yodar} is an uncertainty-based method in which uncertainty combines outputs of radar and camera networks with a gradient-boosting classifier. Experimental results show that YOdar increases performance significantly at night scenes.

Another way that may be useful to increase the networks' robustness is generative models. They can detect sensor defects or new scenarios an autonomous vehicle has never entered. 
Wheeler~\etal~\cite{wheeler2017deep} described a methodology for constructing stochastic automotive radar models based on deep learning with adversarial loss connected to real-world data. The resulting model exhibits fundamental radar effects while maintaining real-time capability.
Lekic and Babic ~\etal~\cite{lekic2019automotive} introduced a Conditional Multi-Generator Generative Adversarial Network (CMGGAN) to generate pseudo-images containing all the surrounding objects detected by the radar sensor.
In our opinion, designing specific deep generative models for radar-camera fusion is an interesting open question.

%

\subsubsection{Model Evaluations}


Most researchers utilize the nuScenes \cite{caesar2020nuscenes} dataset to evaluate the performance of their algorithms.
However, the selected sub-dataset and the evaluation metrics are different, leading to a lack of direct comparisons. As summarized in Table \ref{tab:evaluation}, some methods \cite{nabati2019rrpn, john2020so, chang2020spatial, kowol2020yodar, nobis2021radar} use portions of the nuScenes dataset for training, validating and testing, while some others \cite{john2019rvnet, nabati2020radar, kim2020grif, stacker2022fusion} exploit data collected by part of sensors within the nuScenes dataset. In addition, some researchers \cite{nobis2019deep, yadav2020radar+, li2020feature} do not clarify which section of the nuScenes dataset they utilized in their experiments. 

In terms of evaluation metrics, even though some studies provide results using AP and mAP metrics, the type and value of the threshold are different. Besides, only a few works provide information on the inference time, which is also calculated by authors on their own devices and lacks uniform hardware measurements.
In our opinion, since the nuScenes \cite{caesar2020nuscenes} dataset has been used to evaluate the performance of major algorithms, researchers should validate the performance of their algorithms on the same IoU, metrics and sub-dataset. This would enable a more meaningful and direct comparison of the results obtained from various studies.

Moreover, standard evaluation metrics are not specifically designed for situations where sensors are defective. Metrics related to uncertainties, such as Probability-based Detection Quality (PDQ) \cite{hall2020probabilistic}, may be helpful in radar-camera fusion to compare the robustness of different algorithms.  
Radar-camera fusion also faces the challenges of unknown objects in the open world. In such scenarios, evaluation metrics proposed in \cite{dhamija2020overlooked, joseph2021towards} can be utilized for open-set objects in radar-camera fusion.

Furthermore, visualization evaluation techniques are a potential research direction for analyzing and optimizing radar-camera fusion networks. Several methods \cite{montavon2018methods, hohman2018visual} have been proposed for interpreting and understanding deep neural networks. However, to the best of our knowledge, there has yet to be an investigation of visual analytics in radar-camera fusion. How to design the visual toolkits for radar-camera fusion networks is still an open and challenging question.

\subsubsection{Model Deployment}


Radar-camera fusion holds significant potential in practical autonomous driving vehicles, where models of radar-camera fusion are deployed on edge devices.
Compared with high computational servers, edge devices are often equipped with limited computational resources in memory, bandwidth, Graphics Processing Unit (GPU) and Central Processing Unit (CPU). Nevertheless, they still need to meet the low-latency and high-performance requirements. Currently,\cite{yu2020autonomous} is the only work that reports on fusion output speed, reaching 11 Hz on an NVIDIA Jetson AGX TX2.
The results of fusion algorithms on edge devices are an open question, and how to improve the computational efficiency is worth considering. Some network acceleration schemes (e.g., pruning and quantization \cite {han2015deep, cheng2017survey}) are good choices to be applied to radar-camera fusion models.

It is valuable to implement multiple tasks in a uni-model for real applications. In multi-task learning, the knowledge learned during training for one task can be shared and used to improve performance on the other tasks \cite{liang2019multi}. Besides, by sharing model features between multiple tasks, the overall number of parameters and computations can be reduced, making it more efficient in real-time autonomous driving applications \cite{nabati2020radar, liu2023bevfusion}. 
Multi-task in radar-camera fusion is still in the preliminary stage, and we believe the multi-task approach in radar-camera fusion is a potential research direction. Nonetheless, combining multiple tasks into a unified optimization objective results in a complex optimization problem, especially when the tasks are related but have different performance metrics. Finding a set of hyper-parameters that can effectively balance the importance of different tasks is challenging.


\section{Conclusion} \label{Conclusion}

With the rapid development of autonomous driving, radar-camera fusion, a multi-modal and all-weather solution, is gaining more attention in both academic research and industrial applications. This review investigates and discusses radar-camera fusion studies on object detection and semantic segmentation tasks.
Starting with the working principles of radar and camera sensors, we gradually introduce the importance of radar-camera fusion in autonomous driving perception. 
Through the analysis of radar signal processing, we gain a deep understanding of radar representations, which also provides fundamental support for the radar-camera fusion datasets. 
As to fusion methodologies, we delve into various fusion methods and explore questions about ``why to fuse'', ``what to fuse'', ``where to fuse'', ``when to fuse'' and ``how to fuse''.


Based on the current radar-camera fusion datasets and methods, we discuss the critical challenges and raise possible research directions involving multi-modal data and multi-modal fusion.
In general, radar-camera fusion is moving towards data representations containing rich information. On the one hand, representations such as ADC signals and radar tensors provide more potential information, which is valuable for multi-modal fusion. On the other hand, the new 4D radar sensors provide denser point clouds and higher resolutions, which will become a new trend in autonomous driving.
Fusion approaches are evolving towards customizing radar algorithms based on particular radar characteristics. Additionally, methods on multi-frames and multi-tasks in radar-camera fusion are expected in future works. 
Above all, we hope that our survey serves as a comprehensive reference for researchers and practitioners in developing robust perception in radar-camera fusion.


\section*{Acknowledgment}
This research was funded by the Suzhou Municipal Key Laboratory for Intelligent Virtual Engineering (SZS2022004), the Key Programme Special Fund of XJTLU (KSF-A-19), the Suzhou Science and Technology Project (SYG202122), the Research Development Fund of XJTLU (RDF-19-02-23) and Jiangsu Engineering Research Center for Data Science and Cognitive Computing. 
This work received financial support from Jiangsu Industrial Technology Research Institute (JITRI) and Wuxi National Hi-Tech District (WND).

\bibliographystyle{IEEEtran}
\bibliography{IEEEabrv,weblink,citations}

\begin{IEEEbiography}[{\includegraphics[width=1in,height=1.25in,clip,keepaspectratio]{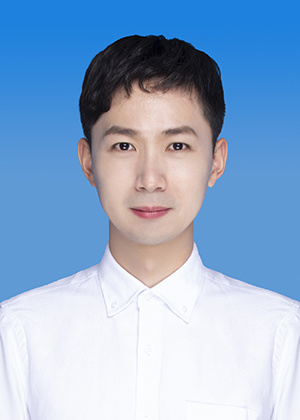}}]{Shanliang Yao} (Student Member, IEEE) received the B.E. degree in 2016 from the School of Computer Science and Technology, Soochow University, Suzhou, China, and the M.S. degree in 2021 from the Faculty of Science and Engineering, University of Liverpool, Liverpool, U.K. He is currently a joint Ph.D. student of University of Liverpool, Xi'an Jiaotong-Liverpool University and Institute of Deep Perception Technology, Jiangsu Industrial Technology Research Institute. His current research is centered on multi-modal perception using deep learning approach for autonomous driving. He is also interested in robotics, autonomous vehicles and intelligent transportation systems.
\end{IEEEbiography}
\vskip -1cm
\begin{IEEEbiography}
[{\includegraphics[width=1in,height=1.25in,clip,keepaspectratio]{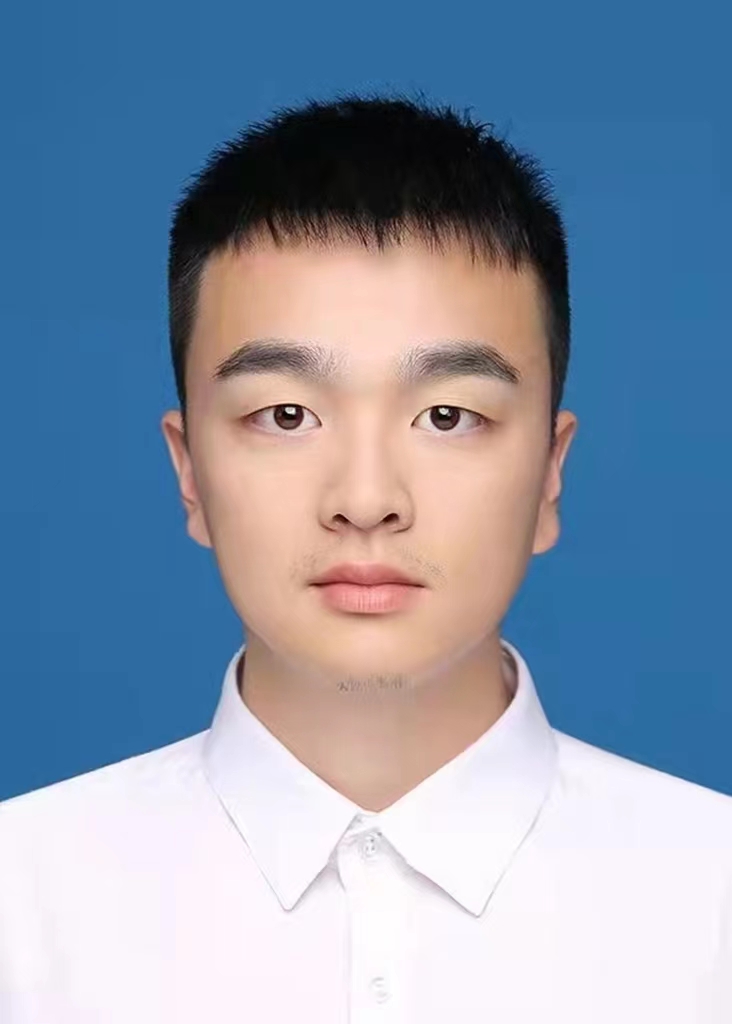}}]{Runwei Guan} (Student Member, IEEE) received his M.S. degree in Data Science from University of Southampton, Southampton, United Kingdom, in 2021. He is currently a joint Ph.D. student of University of Liverpool, Xi'an Jiaotong-Liverpool University and Institute of Deep Perception Technology, Jiangsu Industrial Technology Research Institute. His research interests include visual grounding, panoptic perception based on the fusion of radar and camera, lightweight neural network, multi-task learning and statistical machine learning. He serves as the peer reviewer of IEEE TRANSACTIONS ON NEURAL NETWORKS AND LEARNING SYSTEMS, Engineering Applications of Artificial Intelligence, Journal of Supercomputing, IJCNN, etc.
\end{IEEEbiography}
\vskip -1cm
\begin{IEEEbiography}
[{\includegraphics[width=1in,height=1.25in,clip,keepaspectratio]{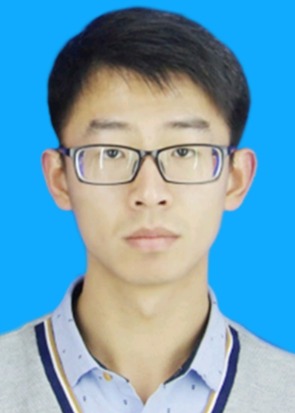}}]{Xiaoyu Huang} (Student Member, IEEE) received the B.S. degree in electronic science and technology from Nanjing University of Information Science \& Technology in 2019. In 2022, he received the M.S. degree from the University of Liverpool and Xi'an Jiaotong-Liverpool University joint program, where he is currently working toward the Ph.D. degree with the School of Advanced Technology. His research interests include computer vision, pattern recognition, deep learning, and image processing.
\end{IEEEbiography}
\vskip -1cm
\begin{IEEEbiography}
[{\includegraphics[width=1in,height=1.25in,clip,keepaspectratio]{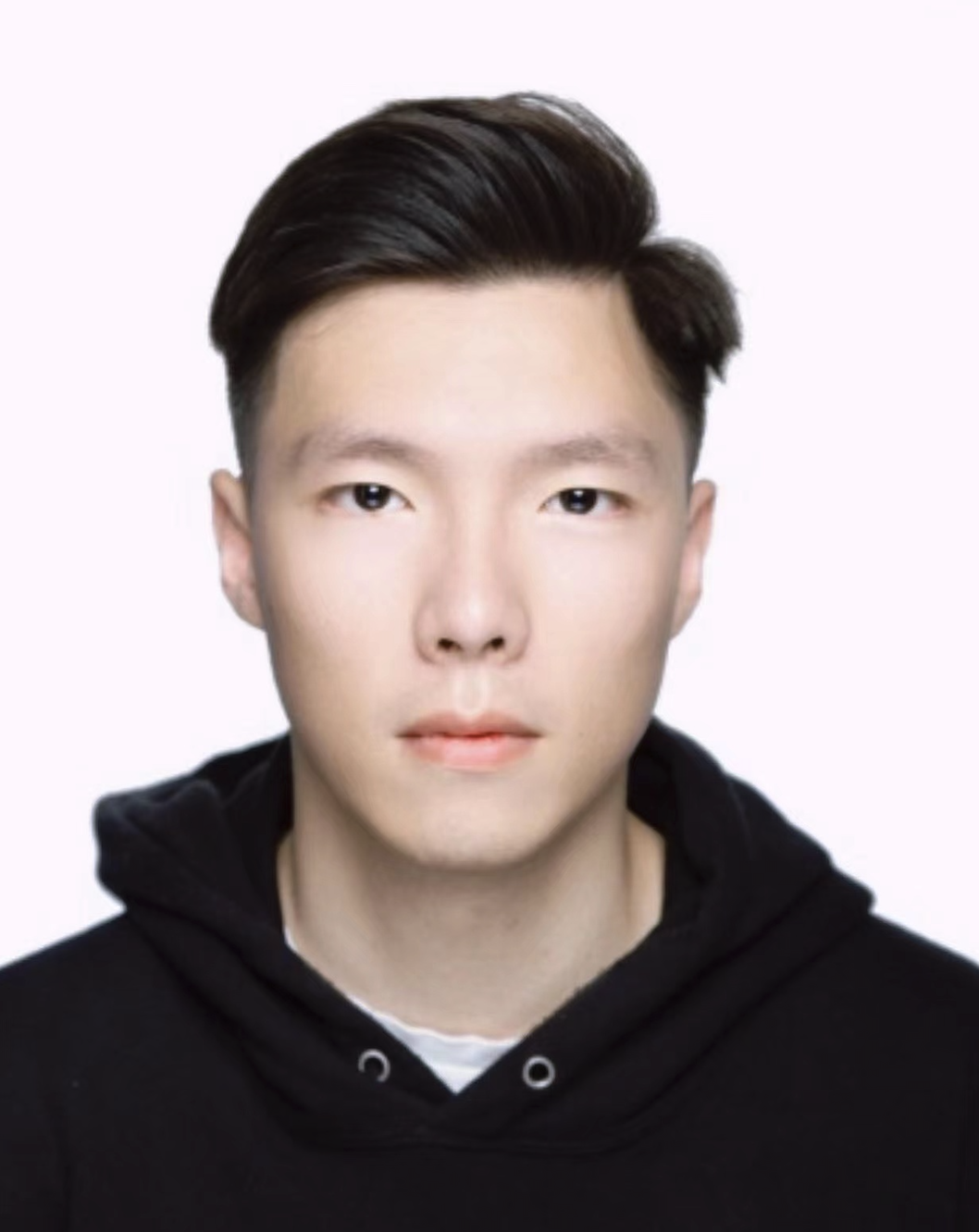}}]{Zhuoxiao Li} (Student Member, IEEE) received the MSc degree in Information Systems (2020)  from the Information School, University of Sheffield, Sheffield, UK. After that, he joined the VR+ Culture Lab of the School of Information Management, Sun Yat-sen University as a research assistant. He is currently a Ph.D student at the Department of Computer Science at the University of Liverpool. His current main research direction is the combination of unmanned surface vehicle and virtual reality/augmented reality. He is also interested in deep learning applications in geographic information systems and remote sensing.
\end{IEEEbiography}
\vskip -1cm
\begin{IEEEbiography}
[{\includegraphics[width=1in,height=1.25in,clip,keepaspectratio]{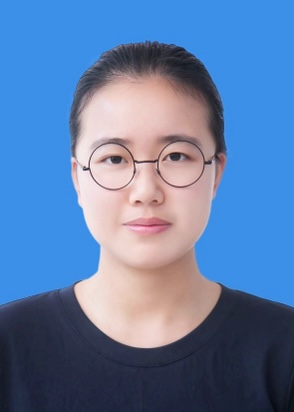}}]{Xiangyu Sha} (Student Member, IEEE) is currently working toward the B.S. degree in computer science from University of Liverpool, United Kingdom. She will complete her undergraduate study in 2024. Her research interests include computer vision, virtual reality, extended reality, robotics simulation and sensor fusion in autonomous driving vehicles. She has won full scholarships for two consecutive years and participated the Programme and Poster Competition as a Summer Undergraduate Research Fellow in 2022.
\end{IEEEbiography}
\vskip -1cm
\begin{IEEEbiography}
[{\includegraphics[width=1in,height=1.25in,clip,keepaspectratio]{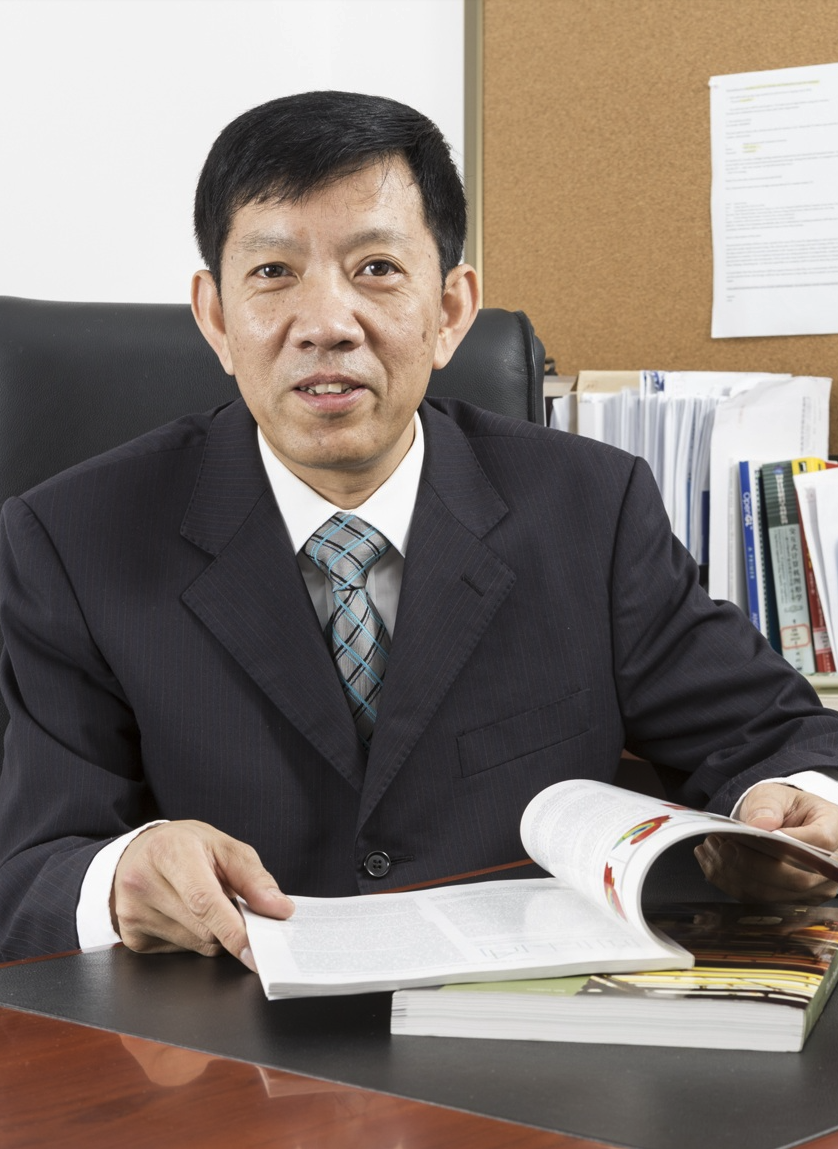}}]{Yong Yue}
Fellow of Institution of Engineering and Technology (FIET), received the B.Eng. degree in mechanical engineering from Northeastern University, Shenyang, China, in 1982, and the Ph.D. degree in computer aided design from Heriot-Watt University, Edinburgh, U.K., in 1994. He worked in the industry for eight years and followed experience in academia with the University of Nottingham, Cardiff University, and the University of Bedfordshire, U.K. He is currently a Professor and Director with the Virtual Engineering Centre, Xi'an Jiaotong-Liverpool University, Suzhou, China. His current research interests include computer graphics, virtual reality, and robot navigation.
\end{IEEEbiography}
\vspace{-1cm}
\begin{IEEEbiography}
[{\includegraphics[width=1in,height=1.25in,clip,keepaspectratio]{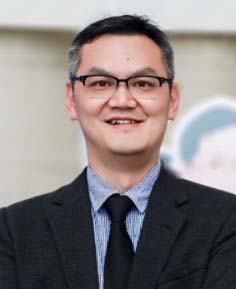}}]{Eng Gee Lim}
(Senior Member, IEEE) received the B.Eng. (Hons.) and Ph.D. degrees in Electrical and Electronic Engineering (EEE) from Northumbria University, Newcastle, U.K., in 1998 and 2002,
respectively. He worked for Andrew Ltd., Coventry, U.K., a leading communications systems company from 2002 to 2007. Since 2007, he has been with Xi'an Jiaotong–Liverpool University, Suzhou, China, where he was the Head of the EEE Department, and the University Dean of research and graduate studies. He is currently the School Dean of Advanced Technology, the Director of the AI University Research Centre, and a Professor with the Department of EEE. He has authored or coauthored over 100 refereed international journals and conference papers. His research interests are artificial intelligence (AI), robotics, AI+ health care, international standard (ISO/IEC) in robotics, antennas, RF/microwave engineering, EM measurements/simulations, energy harvesting, power/energy transfer, smart-grid communication, and wireless communication networks for smart and green cities. He is a Charted Engineer and a fellow of The Institution of Engineering and Technology (IET) and Engineers Australia. He is also a Senior Fellow of Higher Education Academy (HEA).
\end{IEEEbiography}
\vspace{-1cm}
\begin{IEEEbiography}[{\includegraphics[width=1in,height=1.25in,clip,keepaspectratio]{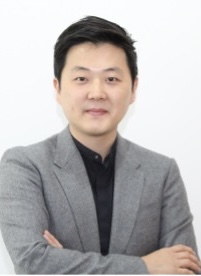}}]{Hyungjoon Seo} (Member, IEEE), received the bachelor's degree in civil engineering from Korea University, Seoul, South Korea, in 2007, and the Ph.D. degree in geotechnical engineering from Korea University in 2013. In 2013, he worked as a research professor in Korea University. He served as a visiting scholar at University of Cambridge, Cambridge, UK, and he worked for engineering department in University of Cambridge as a research associate from 2014 to 2016. In August 2016, he got an assistant professor position in the Department of Civil Engineering at the Xi'an Jiaotong Liverpool University (XJTLU), China. He has been an assistant professor at the University of Liverpool, UK, from 2020. His research interests are monitoring using artificial intelligence and SMART monitoring system for infrastructure, soil-structure interaction (tunneling, slope stability, pile), Antarctic survey and freezing ground. 
Hyungjoon is the director of the CSMI (Centre for SMART Monitoring Infrastructure), CSMI is collaborating with University of Cambridge, University of Oxford, University of Bath, UC Berkeley University, Nanjing University, and Tongji University on SMART monitoring. He presented a keynote speech at the 15th European Conference on Soil Mechanics and Geotechnical Engineering in 2015. He is currently appointed editor of the CivilEng journal and organized two international conferences. He has published more than 50 scientific papers including a book on Geotechnical Engineering and SMART monitoring. 
\end{IEEEbiography}
\vspace{-1cm}
\begin{IEEEbiography}[{\includegraphics[width=1in,height=1.25in,clip,keepaspectratio]{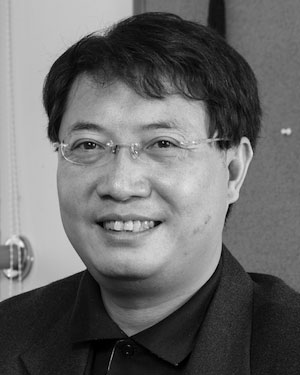}}]{Ka Lok Man}
(Member, IEEE), received the Dr.Eng. degree in electronic engineering from the Politecnico di Torino, Turin, Italy, in 1998, and the Ph.D. degree in computer science from Technische Universiteit Eindhoven, Eindhoven, The Netherlands, in 2006. He is currently a Professor in Computer Science and Software Engineering with Xi'an Jiaotong-Liverpool University, Suzhou, China. His research interests include formal methods and process algebras, embedded system design and testing, and photovoltaics.
\end{IEEEbiography}
\vspace{-1cm}
\begin{IEEEbiography}
[{\includegraphics[width=1in,height=1.25in,clip,keepaspectratio]{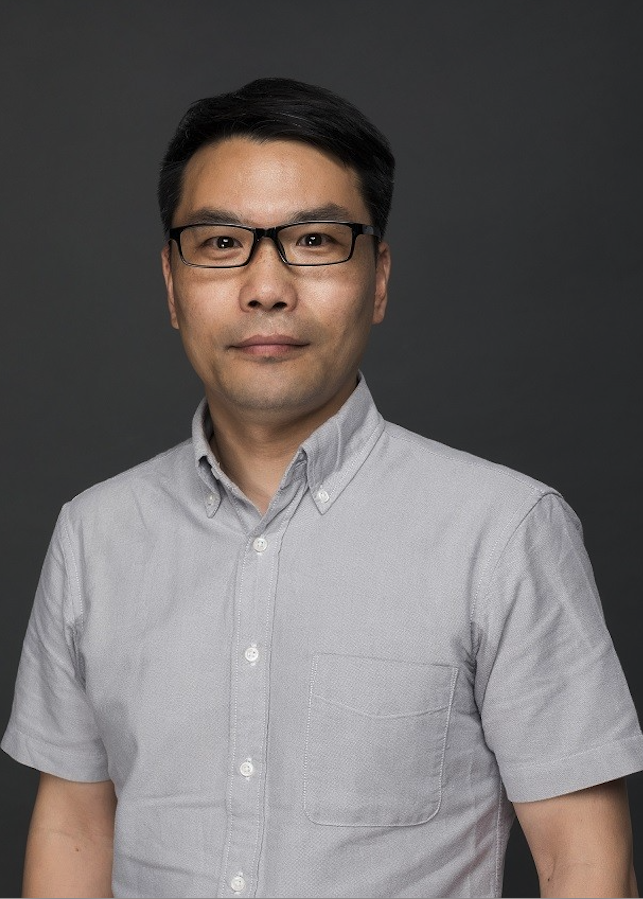}}]{Xiaohui Zhu}
(Member, IEEE) received his Ph.D. from the University of Liverpool, UK in 2019. He is currently an assistant professor, Ph.D. supervisor and Programme Director with the Department of Computing, School of Advanced Technology, Xi'an Jiaotong-Liverpool University. He focuses on advanced techniques related to autonomous driving, including sensor-fusion perception, fast path planning, autonomous navigation and multi-vehicle collaborative scheduling. 
\end{IEEEbiography}
\vspace{-1cm}
\begin{IEEEbiography}[{\includegraphics[width=1in,height=1.25in,clip,keepaspectratio]{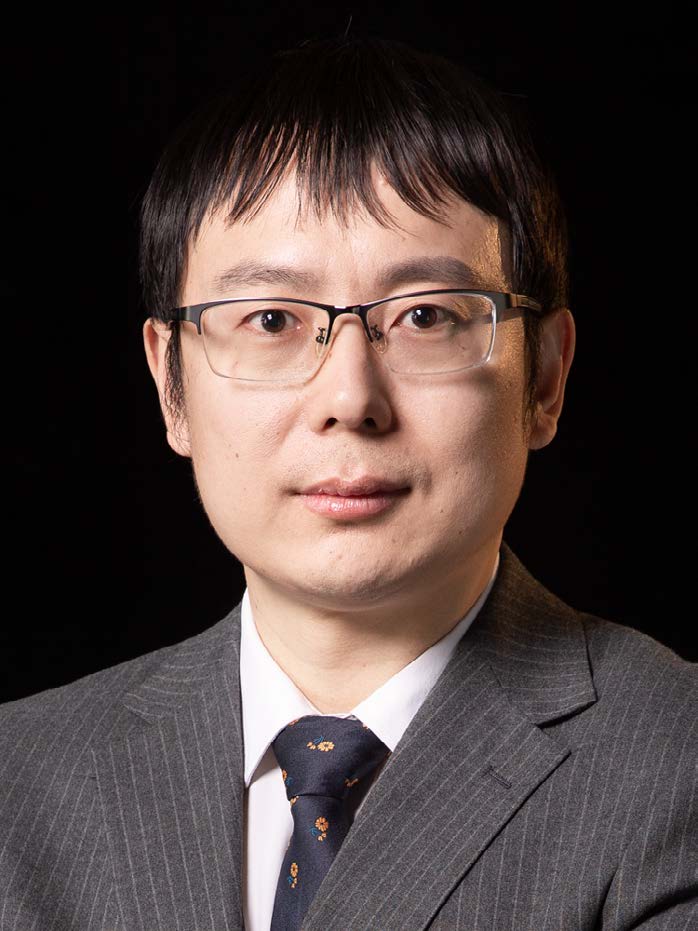}}]{Yutao Yue}
(Member, IEEE) was born in Qingzhou, Shandong, China, in 1982. He received the B.S. degree in applied physics from the University of Science and Technology of China, in 2004, and the M.S. and Ph.D. degrees in computational physics from Purdue University, USA, in 2006 and 2010, respectively. From 2011 to 2017, he worked as a Senior Scientist with the Shenzhen Kuang-Chi Institute of Advanced Technology and a Team Leader of the Guangdong ``Zhujiang Plan'' 3rd Introduced Innovation Scientific Research Team. From 2017 to 2018, he was a Research Associate Professor with the Southern University of Science and Technology, China. Since 2018, he has been the Founder and the Director of the Institute of Deep Perception Technology, JITRI, Jiangsu, China. Since 2020, he has been working as an Honorary Recognized Ph.D. Advisor of the University of Liverpool, U.K., and Xi'an Jiaotong-Liverpool University, China. He is the co-inventor of over 300 granted patents of China, USA, and Europe. He is also the author of over 20 journals and conference papers. His research interests include computational modeling, radar vision fusion, perception and cognition cooperation, artificial intelligence theory, and electromagnetic field modulation. Dr. Yue was a recipient of the Wu WenJun Artificial Intelligence Science and Technology Award in 2020.
\end{IEEEbiography}

\vfill

\onecolumn

\begin{landscape}
\captionsetup{font=normalsize}
\scriptsize
\begin{longtable}{p{1.8cm}p{0.6cm}p{2.2cm}p{1.9cm}p{3cm}p{4cm}p{4cm}p{3.1cm}}

\caption{\MakeUppercase{Overview of radar-camera fusion datasets}}\label{tab:datasets}\\

\toprule
\textbf{Name} & \textbf{Year} & \textbf{Tasks} & \textbf{Annotations} & \textbf{Data Representations} & \textbf{Categories}  & \textbf{Size}  & \textbf{Link} \\\midrule
\endfirsthead
\caption{\MakeUppercase{Overview of radar-camera fusion datasets}}\\
\toprule
\textbf{Name} & \textbf{Year} & \textbf{Tasks} & \textbf{Annotations} & \textbf{Data Representations} & \textbf{Categories}  & \textbf{Size}  & \textbf{Link} \\\midrule
\endhead

nuScenes \cite{caesar2020nuscenes} & 2019 &  
\begin{minipage}[t]{1\textwidth}
Object Detection,\\
Object Tracking
\end{minipage}
&
3D box-level 
& 
\begin{minipage}[t]{1\textwidth}
Camera: RGB image;\\
Radar: point cloud
\end{minipage}
& 23 classes (Vehicle, Pedestrian, Bicycle, Movable Object, Static Object, etc.)  & 1000 scenes, 1.4M boxes in 40k frames, 5.5 hours  & \url{https://www.nuscenes.org/nuscenes} \\\midrule

Astyx \cite{meyer2019automotive} & 2019 & Object Detection & 3D box-level & \begin{minipage}[t]{1\textwidth}
Camera: RGB image;\\
Radar: point cloud
\end{minipage} & 7 classes (Bus, Car, Cyclist, Motorcyclist, Person, Trailer, Truck) & 500 frames, around 3000 labeled objects & \url{http://www.astyx.net} \\\midrule

SeeingThroughFog \cite{bijelic2020seeing} & 2020 & Object Detection & 
\begin{minipage}[t]{0.4\textwidth}
2D box-level,\\
3D box-level
\end{minipage} &  
\begin{minipage}[t]{0.4\textwidth}
Camera: RGB image, \\ gated image, thermal image;\\
Radar: point cloud
\end{minipage}
& 4 classes (Passenger Car, Large Vehicle, Pedestrian, Ridable Vehicle) & 12k samples in real-world driving scenes and 1.5k samples in controlled weather conditions within a fog chamber, 100k objects  & \url{https://www.uni-ulm.de/en/in/driveu/projects/dense-datasets} \\\midrule

CARRADA \cite{ouaknine2021carrada} & 2020 & 
\begin{minipage}[t]{1\textwidth}
Object Detection,\\
Semantic Segmentation,\\
Object Tracking, \\
Scene Understanding
\end{minipage}
& \begin{minipage}[t]{1\textwidth}
2D box-level,\\
2D pixel-level \\
\end{minipage} & \begin{minipage}[t]{1\textwidth}
Camera: RGB image; \\
Radar: range-azimuth tensor,\\
range-Doppler tensor\\
\end{minipage} & 3 classes (Pedestrian, Car, Cyclist)  & 12,666 frames, 78 instances, 7,139 annotated frames with instances, 23GB synchronized camera and radar views & \url{https://arthurouaknine.github.io/codeanddata/carrada}\\\midrule

HawkEye \cite{guan2020through} & 2020 & 
\begin{minipage}[t]{1\textwidth}
Semantic Segmentation\\
\end{minipage} & \begin{minipage}[t]{1\textwidth}
3D point-level 
\end{minipage}  & 
\begin{minipage}[t]{1\textwidth}
Camera: RGB image; \\
Radar: point cloud 
\end{minipage} & 9 classes of cars (Sub-compact, Compact, Mid-sized, Full-sized, Sport, SUV, Jeep, Van, Truck) & 3k images, 4k scenes, 120 car models  & \url{https://jguan.page/HawkEye} \\\midrule

Zendar \cite{mostajabi2020high} & 2020 & 
\begin{minipage}[t]{1\textwidth}
Object Detection,\\
Mapping,\\
Localization \\
\end{minipage}
& 2D box-level & 
\begin{minipage}[t]{1\textwidth}
Camera: RGB image;\\
Radar: range-Doppler tensor, \\ range-azimuth tensor, \\point cloud \\
\end{minipage} & 1 class (Car) & Over 11k moving cars labeled in 27 diverse scenes with over 40k automatically generated labels & \url{http://zendar.io/dataset}\\\midrule

RADIATE \cite{sheeny2021radiate} & 2020 & Object Detection & 2D box-level & \begin{minipage}[t]{1\textwidth}
Camera: RGB image;\\
Radar: range-azimuth tensor \\
\end{minipage}
& 8 classes (Car, Van, Bus, Truck, Motorbike, Bicycle, Pedestrian, A group of pedestrians) & 200k bounding boxes over 44k radar frames & \url{http://pro.hw.ac.uk/radiate} \\\midrule

AIODrive \cite{weng2023all} & 2020 & \begin{minipage}[t]{1\textwidth}
Object Detection,\\
Semantic Segmentation,\\
Object Tracking,\\
Trajectory Prediction, \\
Depth Estimation
\end{minipage} & 
\begin{minipage}[t]{1\textwidth}
2D box-level,\\
3D box-level\\
\end{minipage}
& \begin{minipage}[t]{1\textwidth}
Camera: RGB image;\\
Radar: point cloud\\
\end{minipage} &
11 classes (Vehicle, Pedestrian, Vegetation, Building, Road, Sidewalk, Wall, Traffic Sign, Pole and Fence) & 500k annotated images for five camera viewpoints, 100k annotated frames for radar sensor  & \url{http://www.aiodrive.org}\\\midrule

CRUW \cite{wang2021sne} & 2021 & Object Detection & 2D box-level & 
\begin{minipage}[t]{1\textwidth}
Camera: RGB image;\\
Radar: range-azimuth tensor \\
\end{minipage} & 3 classes (Pedestrian, Cyclist, Car) & 400k frames, 260k objects, 3.5 hours & \url{https:/www.cruwdataset.org/}  \\\midrule

RaDICaL \cite{lim2021radical} & 2021 & Object Detection & 2D box-level & \begin{minipage}[t]{1\textwidth}
Camera: RGB image, \\ RGB-D image;\\
Radar: ADC signal \\
\end{minipage} & 2 classes (Car, Pedestrian) & 393k frames & \url{https://publish.illinois.edu/radicaldata}\\\midrule

RadarScenes \cite{schumann2021radarscenes} & 2021 & 
\begin{minipage}[t]{1\textwidth}
Semantic Segmentation, \\Object Tracking\\
\end{minipage} & \begin{minipage}[t]{1\textwidth}
2D point-level
\end{minipage}
&
\begin{minipage}[t]{1\textwidth}
Camera: RGB image;\\
Radar: point cloud \\
\end{minipage} & 11 classes (Car, Large Vehicle, Truck, Bus, Train, Bicycle, Motorized Two-wheeler, Pedestrian, Pedestrian Group, Animal, Other) & 40,208 frames, 158 individual sequences, 118.9M radar points &\url{https://radar-scenes.com}\\\midrule

RADDet \cite{zhang2021raddet} & 2021 & Object Detection & 
\begin{minipage}[t]{1\textwidth}
2D box-level,\\
3D box-level\\
\end{minipage} & 
\begin{minipage}[t]{1\textwidth}
Camera: RGB image;\\
Radar: range-azimuth-Doppler \\tensor \\
\end{minipage}
& 6 classes (Person, Bicycle, Car, Motorcycle, Bus, Truck) & 10,158 frames & \url{https://github.com/ZhangAoCanada/RADDet}\\\midrule

FloW \cite{cheng2021flow} & 2021 & Object Detection & 2D box-level &
\begin{minipage}[t]{1\textwidth}
Camera: RGB image;\\
Radar: range-Doppler tensor, \\point cloud \\
\end{minipage}
& 1 class (Bottle) & 4k frames & \url{https://github.com/ORCA-Uboat/FloW-Dataset}\\\midrule

RADIal \cite{rebut2022raw} & 2021 & 
\begin{minipage}[t]{1\textwidth}
Object Detection,\\
Semantic Segmentation\\
\end{minipage}
 & 2D box-level & \begin{minipage}[t]{1\textwidth}
Camera: RGB image;\\
Radar: ADC signal, \\  range-azimuth-Doppler tensor, \\ range-azimuth tensor, \\ range-Doppler tensor, \\ point cloud \\
\end{minipage} & 1 class (Vehicle)  & 8,252 frames are labelled with 9,550 vehicle & \url{
https://github.com/valeoai/RADIal} \\\midrule

VoD \cite{palffy2022multi} & 2022 & Object Detection & \begin{minipage}[t]{1\textwidth}
2D box-level,\\
3D box-level\\
\end{minipage} & 
\begin{minipage}[t]{1\textwidth}
Camera: RGB image;\\
Radar: point cloud \\
\end{minipage} &
13 classes (Car, Pedestrian, Cyclist, Rider, Unused Bicycle, Bicycle Rack, Human Depiction, Moped or Scooter, Motor, Ride Other, Vehicle Other, Truck, Ride Uncertain)  & 8693 frames, 123,106 annotations of both moving and static objects, including 26,587 pedestrian, 10,800 cyclist and 26,949 car labels  & \url{https://tudelft-iv.github.io/view-of-delft-dataset}\\\midrule

Boreas \cite{burnett2023boreas} & 2022 & 
\begin{minipage}[t]{1\textwidth}
Object Detection,\\
Localization,\\
Odometry
\end{minipage}
& 2D box-level &
\begin{minipage}[t]{1\textwidth}
Camera: RGB image;\\
Radar: range-azimuth tensor \\
\end{minipage}
& 4 classes (Car, Pedestrian, Cyclist, Misc)  & 7.1k frames for detection, over 350km of driving data, 326,180 unique 3D box annotations & \url{https://www.boreas.utias.utoronto.ca}\\\midrule

TJ4DRadSet \cite{zheng2022tj4dradset} & 2022 & \begin{minipage}[t]{1\textwidth}
Object Detection,\\
Object Tracking\\
\end{minipage} & 3D box-level & 
\begin{minipage}[t]{1\textwidth}
Camera: RGB image;\\
Radar: point cloud \\
\end{minipage} & 8 classes (Car, Pedestrian, Cyclist, Bus, Motorcyclist, Truck, Engineering Vehicle, Tricyclist) & 40k frames in total, 7757 frames within 44 consecutive sequences & \url{https://github.com/TJRadarLab/TJ4DRadSet} \\\midrule

K-Radar \cite{paek2022k} & 2022 & \begin{minipage}[t]{1\textwidth}
Object Detection,\\
Object Tracking,\\
SLAM
\end{minipage} & 3D box-level & 
\begin{minipage}[t]{1\textwidth}
Camera: RGB image;\\
Radar: range-azimuth-Doppler \\tensor \\
\end{minipage} & 5 classes (Pedestrian, Motorbike, Bicycle, Sedan, Bus or Truck) & 35k frames of 4D radar tensor & \url{https://github.com/kaist-avelab/k-radar} \\\midrule

aiMotive \cite{matuszka2022aimotive} & 2022 & Object Detection & 3D box-level 
& \begin{minipage}[t]{1\textwidth}
Camera: RGB image;\\
Radar: point cloud \\
\end{minipage}
& 14 classes (Pedestrian, Car, Bus, Truck, Van, Motorcycle, Pickup, Rider, Bicycle, Trailer, Train, Shopping Cart, Other Object) & 26,583 frames, 425k objects & \url{https://github.com/aimotive/aimotive_dataset} \\\midrule

WaterScenes \cite{yao2023waterscenes} & 2023 & \begin{minipage}[t]{1\textwidth}
Object Detection,\\
Instance Segmentation,\\
Semantic Segmentation,\\
Free-space Segmentation,\\
Waterline Segmentation,\\
Panoptic Perception
\end{minipage} & 2D box-level 
& \begin{minipage}[t]{1\textwidth}
Camera: RGB image;\\
Radar: point cloud \\
\end{minipage}
& 7 classes (Pier, Buoy, Sailor, Ship, Boat, Vessel, Kayak) & 54,120 frames, 202k objects & \url{https://waterscenes.github.io} \\

\bottomrule
\end{longtable}

\end{landscape}
\begin{landscape}

\scriptsize
\begin{longtable}{p{1.9cm}p{0.6cm}p{2.4cm}p{1.5cm}p{2.1cm}p{2.1cm}p{2.2cm}p{0.8cm}p{2.2cm}p{1.5cm}p{2cm}}
\caption{\MakeUppercase{Summary of radar-camera fusion methods}}\label{tab:fusion methods}\\

\toprule
\textbf{Reference} & \textbf{Year} & \textbf{Task} &\textbf{Annotations} & \textbf{Categories} & \textbf{Modalities} & \textbf{Network Architecture} & \textbf{Fusion Level} & \textbf{Fusion Operation} & \textbf{Dataset} & \textbf{Source Code}\\\midrule
\endfirsthead
\caption{\MakeUppercase{Summary of radar-camera fusion methods}}\\
\toprule
\textbf{Reference} & \textbf{Year} & \textbf{Task} & \textbf{Annotations} & \textbf{Categories} & \textbf{Modalities} & \textbf{Network Architecture} & \textbf{Fusion Level} & \textbf{Fusion Operation} & \textbf{Dataset} & \textbf{Source Code}\\\midrule
\endhead

Chadwick~\etal~\cite{chadwick2019distant} & 2019 & Object Detection & 2D box-level & Vehicle & 
\begin{minipage}[t]{1\textwidth}
Camera: RGB image;\\
Radar: point cloud
\end{minipage}
& One-stage network based on ResNet \cite{he2016deep} & Feature-level & \begin{minipage}[t]{1\textwidth}
Addition;\\ Concatenation
\end{minipage}& Self-recorded & - \\\midrule

\begin{minipage}[t]{1\textwidth}
RRPN \cite{nabati2019rrpn}
\end{minipage} & 2019 & Object Detection & 2D box-level & Car, Truck, Person, Motorcycle, Bicycle, Bus & 
\begin{minipage}[t]{1\textwidth}
Camera: RGB image;\\
Radar: point cloud
\end{minipage} 
& RRPN \cite{nabati2019rrpn} & Data-level & \begin{minipage}[t]{1\textwidth}Transformation matrix \end{minipage} & nuScenes \cite{caesar2020nuscenes} & \url{https://github.com/mrnabati/RRPN}\\\midrule

Jha~\etal~\cite{jha2019object} & 2019 & Object Detection & 2D box-level & Pedestrian, Lamp Post & 
\begin{minipage}[t]{1\textwidth}
Camera: RGB image;\\
Radar: point cloud
\end{minipage}
& YOLOv3 \cite{redmon2018yolov3} & Object-level &
\begin{minipage}[t]{1\textwidth}Transformation matrix \end{minipage}
& Self-recorded & - \\\midrule

\begin{minipage}[t]{1\textwidth}
CMGGAN \cite{lekic2019automotive}
\end{minipage} & 2019 & Semantic Segmentation & 2D point-level & Free Space & 
\begin{minipage}[t]{1\textwidth}
Camera: RGB image;\\
Radar: grid map
\end{minipage} & CMGGAN \cite{lekic2019automotive} & Feature-level & Addition & Self-recorded & - \\\midrule

Meyer and Kuschk \cite{meyer2019automotive} & 2019 & Object Detection & 3D box-level & Car & 
\begin{minipage}[t]{1\textwidth}
Camera: RGB image;\\
Radar: point cloud
\end{minipage} & A 3D region proposal network based on VGG \cite{simonyan2014very} & Data-Level & \begin{minipage}[t]{1\textwidth}Transformation matrix\end{minipage} & Astyx \cite{meyer2019automotive} & - \\\midrule

\begin{minipage}[t]{1\textwidth}
RVNet \cite{john2019rvnet}
\end{minipage} & 2019 & Object Detection & 2D box-level & Vehicle, Pedestrian, Two-wheelers, Objects (movable objects and debris) & 
\begin{minipage}[t]{1\textwidth}
Camera: RGB image;\\
Radar: point cloud
\end{minipage} & RVNet \cite{john2019rvnet} based on YOLOv3 \cite{redmon2018yolov3} & Feature-level & Concatenation & nuScenes \cite{caesar2020nuscenes} & -\\\midrule

\begin{minipage}[t]{1\textwidth}
FusionNet \cite{lim2019radar}
\end{minipage} & 2019 &
\begin{minipage}[t]{1\textwidth}
Object Detection,\\
Object Classification
\end{minipage}
& 2D box-level & Vehicle & 
\begin{minipage}[t]{1\textwidth}
Camera: RGB image;\\
Radar: range-azimuth \\tensor
\end{minipage} & FusionNet \cite{lim2019radar} inspired by SSD \cite{liu2016ssd} & Feature-level & Concatenation & nuScenes \cite{caesar2020nuscenes} & - \\\midrule

\begin{minipage}[t]{1\textwidth}
SO-Net \cite{john2020so}
\end{minipage} & 2020 & 
\begin{minipage}[t]{1\textwidth}
Object Detection,\\
Semantic Segmentation
\end{minipage}& 
\begin{minipage}[t]{1\textwidth}
2D box-level,\\
2D pixel-level
\end{minipage} & Vehicle, Free Space &
\begin{minipage}[t]{1\textwidth}
Camera: RGB image;\\
Radar: point cloud
\end{minipage} & SO-Net \cite{john2020so} based on the RVNet \cite{john2019rvnet} & Feature-level & Concatenation & nuScenes \cite{caesar2020nuscenes} & - \\\midrule

\begin{minipage}[t]{1\textwidth}
SAF-FCOS \cite{chang2020spatial}
\end{minipage} & 2020 & Object Detection & 2D box-level & Bicycle, Car, Motorcycle, Bus, Train, Truck &
\begin{minipage}[t]{1\textwidth}
Camera: RGB image;\\
Radar: point cloud
\end{minipage} & SAF-FCOS \cite{chang2020spatial} based on FCOS \cite{tian2019fcos} & Feature-level & \begin{minipage}[t]{1\textwidth}
Addition; \\ Multiplication \end{minipage}  & nuScenes \cite{caesar2020nuscenes} & \url{https://github.com/Singingkettle/SAF-FCOS} \\\midrule

\begin{minipage}[t]{1\textwidth}
CRF-Net \cite{nobis2019deep}
\end{minipage} & 2020 & Object Detection & 2D box-level & Car, Bus, Motorcycle, Truck, Trailer, Bicycle, Human & 
\begin{minipage}[t]{1\textwidth}
Camera: RGB image;\\
Radar: point cloud
\end{minipage} & CRF-Net \cite{nobis2019deep} based on RetinaNet \cite{lin2017focal} & Data-level & Concatenation  & nuScenes \cite{caesar2020nuscenes}, Self-recorded & \url{https://github.com/TUMFTM/CameraRadarFusionNet} \\\midrule

Bijelic~\etal~\cite{bijelic2020seeing} & 2020 & Object Detection & 2D box-level & Vehicle & 
\begin{minipage}[t]{1\textwidth}
Camera: RGB image;\\
Radar: point cloud
\end{minipage} & A modiﬁed VGG \cite{simonyan2014very} backbone and SSD \cite{liu2016ssd} blocks & Feature-level & \begin{minipage}[t]{1\textwidth}
Concatenation; \\ Attention \end{minipage}& DENSE \cite{bijelic2020seeing}& \url{https://github.com/princeton-computational-imaging/SeeingThroughFog} \\\midrule

\begin{minipage}[t]{1\textwidth}
BIRANet \cite{yadav2020radar+}
\end{minipage} & 2020 & Object Detection & 2D box-level & Car, Truck, Person, Motorcycle, Bicycle,  Bus & 
\begin{minipage}[t]{1\textwidth}
Camera: RGB image;\\
Radar: point cloud
\end{minipage} & RANet and BIRANet \cite{yadav2020radar+} based on ResNet \cite{he2016deep} & Feature-level & Addition & nuScenes \cite{caesar2020nuscenes} & \url{https://github.com/RituYadav92/Radar-RGB-Attentive-Multimodal-Object-Detection} \\\midrule

Nabati and Qi~\cite{nabati2020radar} & 2020 & \begin{minipage}[t]{1\textwidth}
Object Detection,\\
Depth Estimation
\end{minipage} & 2D box-level & Car, Truck, Person, Motorcycle, Bicycle, Bus & 
\begin{minipage}[t]{1\textwidth}
Camera: RGB image;\\
Radar: point cloud
\end{minipage} & FPN \cite{lin2017feature} with ResNet \cite{he2016deep} as backbone, and RPN in Faster R-CNN \cite{ren2015faster} & Hybrid-level & - & nuScenes \cite{caesar2020nuscenes}& -\\\midrule

\begin{minipage}[t]{1\textwidth}
YOdar \cite{kowol2020yodar}
\end{minipage} & 2020 & 
\begin{minipage}[t]{1\textwidth}
Object Detection
\end{minipage}
& 2D box-level & Vehicle & 
\begin{minipage}[t]{1\textwidth}
Camera: RGB image;\\
Radar: point cloud
\end{minipage} & YOdar \cite{kowol2020yodar} based on YOLOv3 \cite{redmon2018yolov3} & Feature-level & Concatenation & nuScenes \cite{caesar2020nuscenes}& -\\\midrule

\begin{minipage}[t]{1\textwidth}
CenterFusion \cite{nabati2021centerfusion}
\end{minipage} & 2020 & Object Detection & 3D box-level & Car, Truck, Bus, Trailer, Pedestrian, Barrier, Motorcycle, Bicycle, Traffic Cone & 
\begin{minipage}[t]{1\textwidth}
Camera: RGB image;\\
Radar: point cloud
\end{minipage} & CenterNet \cite{zhou2019objects} with the DLA \cite{yu2018deep} backbone & Feature-level & Concatenation & nuScenes \cite{caesar2020nuscenes} & \url{https://github.com/mrnabati/CenterFusion} \\\midrule

\begin{minipage}[t]{1\textwidth}
RODNet \cite{wang2021rodnet}
\end{minipage} & 2021 & Object Detection & 2D box-level & Pedestrian, Cyclist, Car &
\begin{minipage}[t]{1\textwidth}
Camera: RGB image;\\
Radar: range-azimuth \\tensor
\end{minipage}
& RODNet \cite{wang2021rodnet} & Feature-level & - & CRUW \cite{wang2021rethinking} & \url{https://github.com/yizhou-wang/RODNet} \\\midrule

\begin{minipage}[t]{1\textwidth}
RAMP-CNN \cite{gao2020ramp}
\end{minipage} & 2021 & Object Detection & 2D box-level & Pedestrian, Cyclist, Car &
\begin{minipage}[t]{1\textwidth}
Camera: RGB image;\\
Radar: \\ range-azimuth-Doppler \\tensor
\end{minipage}
& RAMP-CNN \cite{gao2020ramp} & Feature-level & Concatenation & CRUW \cite{wang2021rethinking} & - \\\midrule

Li and Xie~\cite{li2020feature} & 2021 & Object Detection & 3D box-level & Vehicle & 
\begin{minipage}[t]{1\textwidth}
Camera: RGB image;\\
Radar: point cloud
\end{minipage} & A network based on YOLOv3 \cite{redmon2018yolov3} & Feature-level & 
\begin{minipage}[t]{1\textwidth}
Concatenation; \\
Multiplication
\end{minipage} & nuScenes \cite{caesar2020nuscenes} & -\\\midrule

Kim~\etal~\cite{kim2020low} & 2021 & Object Detection & 3D box-level & Vehicle & 
\begin{minipage}[t]{1\textwidth}
Camera: RGB image;\\
Radar: range-azimuth \\tensor
\end{minipage} & A network based on VGG \cite{simonyan2014very} and FPN \cite{lin2017feature} & Feature-level & Concatenation & Self-recorded & -\\\midrule

\begin{minipage}[t]{1\textwidth}
AssociationNet \cite{dong2021radar}
\end{minipage} & 2021 & Object Detection & 2D box-level & Vehicle & 
\begin{minipage}[t]{1\textwidth}
Camera: RGB image;\\
Radar: point cloud
\end{minipage} & AssociationNet \cite{dong2021radar} & Object-level & \begin{minipage}[t]{1\textwidth}Transformation matrix;\\ Concatenation \end{minipage} & Self-recorded & - \\\midrule

\begin{minipage}[t]{1\textwidth}
RVF-Net \cite{nobis2021radar}
\end{minipage} & 2021 & Object Detection & 3D box-level & Car & 
\begin{minipage}[t]{1\textwidth}
Camera: RGB image;\\
Radar: point cloud
\end{minipage} & RVF-Net \cite{nobis2021radar} based on VoxelNet \cite{zhou2018voxelnet} & Data-level & Concatenation & nuScenes \cite{caesar2020nuscenes} & \url{https://github.com/TUMFTM/RadarVoxelFusionNet}\\\midrule

Cui~\etal~\cite{cui20213d} & 2021 & Object Detection & 3D box-level & Car & 
\begin{minipage}[t]{1\textwidth}
Camera: RGB image;\\
Radar: point cloud
\end{minipage} & CNN with SSMA \cite{valada2020self} block & Hybrid-level & Concatenation & Astyx \cite{meyer2019automotive} & - \\\midrule

\begin{minipage}[t]{1\textwidth}
RISFNet \cite{cheng2021robust}
\end{minipage} & 2021 & Object Detection & 2D box-level & Bottle & 
\begin{minipage}[t]{1\textwidth}
Camera: RGB image;\\
Radar: point cloud
\end{minipage} & RISFNet \cite{cheng2021robust} based on CSPdarknet53 \cite{bochkovskiy2020yolov4} and VGG \cite{simonyan2014very} & Feature-level & \begin{minipage}[t]{1\textwidth}
Concatenation; \\Addition; \\Multiplication\\
\end{minipage} & FloW \cite{cheng2021flow} & - \\\midrule

\begin{minipage}[t]{1\textwidth}
GRIF Net \cite{kim2020grif}
\end{minipage} & 2021 & Object Detection & 3D box-level & Vehicle & 
\begin{minipage}[t]{1\textwidth}
Camera: RGB image;\\
Radar: point cloud
\end{minipage} & GRIF Net \cite{kim2020grif} based on FPN \cite{lin2017feature} and SBNet \cite{ren2018sbnet} & Feature level & Attention & nuScenes \cite{caesar2020nuscenes} & - \\\midrule

St\"{a}cker ~\etal~\cite{stacker2022fusion} & 2022 & Object Detection & 2D box-level & Car, Person, Truck, Bicycle, Motorcycle & 
\begin{minipage}[t]{1\textwidth}
Camera: RGB image;\\
Radar: point cloud
\end{minipage} & A network based on RetinaNet \cite{lin2017focal} architecture with a ResNet \cite{he2016deep} backbone & Feature level & 
\begin{minipage}[t]{1\textwidth}
Addition,\\
Concatenation
\end{minipage} & nuScenes \cite{caesar2020nuscenes} & - \\\midrule

\begin{minipage}[t]{1\textwidth}
FUTR3D \cite{chen2023futr3d}
\end{minipage} & 2022 & Object Detection & 3D box-level &  Car, Truck, Bus, Pedestrian, Barrier, Trailer, Construction Vehicle, Motorcycle, Bicycle, Traffic cone & \begin{minipage}[t]{1\textwidth}
Camera: RGB image;\\
Radar: point cloud
\end{minipage} & FUTR3D \cite{zhou2023bridging} with ResNet-101 \cite{he2016deep} as backbone and FPN \cite{lin2017feature} as neck & Feature-level & Concatenation & nuScenes \cite{caesar2020nuscenes} & \url{https://github.com/Tsinghua-MARS-Lab/futr3d} \\\midrule

Simple-BEV~\cite{harley2023simple} & 2022 & Semantic Segmentation & 2D pixel-level & Vehicle, Background & 
\begin{minipage}[t]{1\textwidth}
Camera: RGB image;\\
Radar: point cloud
\end{minipage} & A network with a ResNet \cite{he2016deep} backbone & Feature-level & Concatenation & nuScenes \cite{caesar2020nuscenes} & - \\\midrule

\begin{minipage}[t]{1\textwidth}
RadSegNet \cite{bansal2022radsegnet}
\end{minipage} & 2022 & 
\begin{minipage}[t]{1\textwidth}
Object Detection,\\
Semantic Segmentation
\end{minipage} & 
\begin{minipage}[t]{1\textwidth}
2D box-level,\\
2D pixel-level
\end{minipage} & Car, Truck & 
\begin{minipage}[t]{1\textwidth}
Camera: RGB image;\\
Radar: point cloud, \\ range-azimuth \\ tensor
\end{minipage} & RadSegNet \cite{bansal2022radsegnet} & Data-level & Concatenation & Astyx \cite{meyer2019automotive}, RADIATE \cite{sheeny2021radiate} & - \\\midrule

\begin{minipage}[t]{1\textwidth}
RCBEV \cite{zhou2023bridging}
\end{minipage} & 2022 & Object Detection & 3D box-level &  Car, Truck, Bus, Pedestrian, Barrier, Trailer, Construction Vehicle, Motorcycle, Bicycle, Traffic cone & \begin{minipage}[t]{1\textwidth}
Camera: RGB image;\\
Radar: point cloud
\end{minipage} & RCBEV \cite{zhou2023bridging} with Swin Transformer \cite{liu2021swin} as backbone and FPN \cite{lin2017feature} as neck & Feature-level & Concatenation & nuScenes \cite{caesar2020nuscenes} & - \\\midrule

\begin{minipage}[t]{1\textwidth}
CRAFT~\cite{kim2023craft}
\end{minipage}  & 2022 & Object Detection & 3D box-level & Car, Truck, Bus, Pedestrian, Barrier, Trailer, Construction Vehicle, Motorcycle, Bicycle, Traffic cone & \begin{minipage}[t]{1\textwidth}
Camera: RGB image;\\
Radar: point cloud
\end{minipage} & CRAFT \cite{kim2023craft} based on DLA \cite{yu2018deep} & Data-level & Concatenation & nuScenes \cite{caesar2020nuscenes} & - \\\midrule

\begin{minipage}[t]{1\textwidth}
DeepFusion~\cite{drews2022deepfusion}
\end{minipage}  & 2022 & Object Detection & 3D box-level & Car, Truck, Bus, Pedestrian, Barrier, Trailer, Construction Vehicle, Motorcycle, Bicycle, Traffic cone & \begin{minipage}[t]{1\textwidth}
Camera: RGB image;\\
Radar: point cloud
\end{minipage} & DeepFusion \cite{drews2022deepfusion} & Feature-level  & Concatenation & Self-recorded, nuScenes \cite{caesar2020nuscenes} & - \\\midrule

\begin{minipage}[t]{1\textwidth}
CramNet~\cite{hwang2022cramnet}
\end{minipage} & 2022 & Object Detection & 3D box-level & Car, Van, Truck, Bus, Motorbike, Bicycle & 
\begin{minipage}[t]{1\textwidth}
Camera: RGB image;\\
Radar: \\ range-azimuth tensor
\end{minipage} & CramNet \cite{hwang2022cramnet} & Feature-level & Attention & RADIATE \cite{sheeny2021radiate} & - \\\midrule

\begin{minipage}[t]{1\textwidth}
MVFusion~\cite{wu2023mvfusion}
\end{minipage} & 2023 & Object Detection & 3D box-level & Car, Truck, Bus, Pedestrian, Barrier, Trailer, Construction Vehicle, Motorcycle, Bicycle, Traffic cone & 
\begin{minipage}[t]{1\textwidth}
Camera: RGB image;\\
Radar: point cloud
\end{minipage} & MVFusion \cite{wu2023mvfusion} & Feature-level & 
\begin{minipage}[t]{1\textwidth}
Addition;\\
Concatenation
\end{minipage}& nuScenes \cite{caesar2020nuscenes} & - \\\midrule

\begin{minipage}[t]{1\textwidth}
CRN~\cite{kim2023crn}
\end{minipage} & 2023 & Object Detection & 3D box-level & Car, Truck, Bus, Pedestrian, Barrier, Trailer, Construction Vehicle, Motorcycle, Bicycle, Traffic cone & 
\begin{minipage}[t]{1\textwidth}
Camera: RGB image;\\
Radar: point cloud
\end{minipage} & CRN \cite{kim2023crn} based on ResNet \cite{he2016deep}, ConvNeXt \cite{liu2022convnet} and FPN \cite{lin2017feature} & Feature-level & Concatenation & nuScenes \cite{caesar2020nuscenes} & - \\\midrule

\begin{minipage}[t]{1\textwidth}
RCFusion~\cite{zheng2023rcfusion}
\end{minipage} & 2023 & Object Detection & 3D box-level & Car, Truck, Pedestrian, Cyclist & 
\begin{minipage}[t]{1\textwidth}
Camera: RGB image;\\
Radar: point cloud
\end{minipage} & RCFusion~\cite{zheng2023rcfusion} & Feature-level & 
\begin{minipage}[t]{1\textwidth}
Concatenation;\\
Multiplication;\\
Attention
\end{minipage} & 
VoD \cite{palffy2022multi},
TJ4DRadSet \cite{zheng2022tj4dradset} & - \\\midrule

\begin{minipage}[t]{1\textwidth}
LXL~\cite{xiong2023lxl}
\end{minipage} & 2023 & Object Detection & 3D box-level & Car, Pedestrian, Cyclist & 
\begin{minipage}[t]{1\textwidth}
Camera: RGB image;\\
Radar: point cloud
\end{minipage} & LXL~\cite{xiong2023lxl} & Feature-level & 
\begin{minipage}[t]{1\textwidth}
Concatenation;\\
Multiplication
\end{minipage} & 
VoD \cite{palffy2022multi},
TJ4DRadSet \cite{zheng2022tj4dradset}
 & - \\\midrule

\begin{minipage}[t]{1\textwidth}
Achelous~\cite{guan2023achelous}
\end{minipage} & 2023 & 
\begin{minipage}[t]{1\textwidth}
Object Detection,\\
Semantic Segmentation,\\
Free-space Segmentation,\\
Waterline Segmentation,\\
Point Cloud Segmentation
\end{minipage} 
& \begin{minipage}[t]{1\textwidth}
2D box-level,\\
2D pixel-level,\\
3D point-level
\end{minipage}
& Pier, Buoy, Sailor, Ship, Boat, Vessel, Kayak & 
\begin{minipage}[t]{1\textwidth}
Camera: RGB image;\\
Radar: point cloud
\end{minipage} & Achelous~\cite{guan2023achelous} & Data-level & Concatenation & WaterScenes \cite{yao2023waterscenes} & \url{https://github.com/GuanRunwei/Achelous} \\


\bottomrule
\end{longtable}

\end{landscape}
\begin{landscape}

\scriptsize
\begin{longtable}{p{3.6cm}p{8cm}p{11cm}}
\caption{\MakeUppercase{Summary of radar-camera fusion evaluation metrics}}\label{tab:evaluation metrics}\\

\toprule
\multicolumn{1}{l}{\textbf{Metric}} & \multicolumn{1}{c}{\textbf{Formula}} &  \multicolumn{1}{c}{\textbf{Definition}} \\\midrule
\endfirsthead

\caption{\MakeUppercase{Summary of radar-camera fusion evaluation metrics}}\\
\toprule
\multicolumn{1}{l}{\textbf{Metric}} & \multicolumn{1}{c}{\textbf{Formula}} & \multicolumn{1}{c}{\textbf{Definition}} \\\midrule
\endhead

Accuracy
& 
\begin{equation}
\mathrm{Accuracy} = \frac{\mathrm{TP + TN}}{\mathrm{TP + TN + FP + FN}}
\end{equation}
& \hspace{1cm}
Accuracy is the number of correct predictions over all predictions.
\\\midrule

{\vspace{0.2cm}}
Precision
& 
\begin{equation}
\mathrm{Precision} = \frac{\mathrm{TP}}{\mathrm{TP} + \mathrm{FP}}
\end{equation}
& 
{\vspace{0.08cm}}
\hspace{1cm} Precision is the fraction of true positive among total predicted positive.
\\\midrule

{\vspace{0.2cm}}
Recall
& 
\begin{equation}
\mathrm{Recall} = \frac{\mathrm{TP}}{\mathrm{TP + FN}} 
\end{equation}
& 
{\vspace{0.08cm}}\hspace{1cm}
Recall is the fraction of true positive over all actual positive in the dataset.
\\\midrule

{\vspace{0.2cm}}
F1-Score
& 
\begin{equation}
\text{F1-Score} = 2 * \frac{\mathrm{Precision * Recall}}{\mathrm{Precision + Recall} } 
\end{equation}
& \hspace{1cm}
\begin{minipage}[t]{0.55\textwidth}
F1-Score is the harmonic mean of precision and recall, describing a balance between precision and recall.
\end{minipage}
\\\midrule

{\vspace{0.35cm}}
Average Precision (AP)  
&
{\vspace{0.05cm}}
\begin{equation}
\mathrm{AP} = {\int_{0}^{1}\mathrm{precision}(r)\mathrm{d}r}
\end{equation}
& \hspace{1cm}
\begin{minipage}[t]{0.55\textwidth}
AP is the precision averaged over all recall values between 0 and 1 for a single class. It is the area under the Precision-Recall curve.
\begin{itemize}
\setlength{\parsep}{0pt}
\setlength{\parskip}{0pt}
\setlength{\leftmargin}{0pt}
  \item $r$: recall value
  \item precision($r$): the precision at recall value of $r$
\end{itemize}
\end{minipage}
\\\midrule

{\vspace{0.2cm}}
Average Recall (AR)
& 
\begin{equation}
\mathrm{AR}=2 \int_{0.5}^1 \operatorname{recall}(o) \mathrm{d} o
\end{equation}
& \hspace{1cm}
\begin{minipage}[t]{0.55\textwidth}
AR is the average of all recalls at IoU thresholds from 0.5 to 1.0. It is twice the area under the Recall-IoU curve.
\begin{itemize}
\setlength{\parsep}{0pt}
\setlength{\parskip}{0pt}
\setlength{\leftmargin}{0pt}
  \item $o$: the IoU overlap
  \item recall($o$): the recall at IoU value of $o$
\end{itemize}
\end{minipage}
\\\midrule

{\vspace{0.2cm}}
Frame Per Second (FPS)
& 
\begin{equation}
\mathrm{FPS}=\frac{\mathrm{m}}{\mathrm{s}} 
\end{equation}
& \hspace{1cm}
\begin{minipage}[t]{0.55\textwidth}
FPS is a measure of how many images the model processes per second.
\begin{itemize}
\setlength{\parsep}{0pt}
\setlength{\parskip}{0pt}
\setlength{\leftmargin}{0pt}
  \item $m$: the number of images
  \item $s$: total seconds consumed
\end{itemize}
\end{minipage}
\\\midrule

{\vspace{0.2cm}}
Mean Average Precision (mAP) &
\begin{equation}
\mathrm{mAP} = {\frac{1}{N}{\sum _{i=1}^{N}\operatorname {AP_i} }}
\end{equation}
& \hspace{1cm}
\begin{minipage}[t]{0.55\textwidth} 
mAP is the average value of AP, that is, the average of the area under the Precision-Recall curve of each category.
\begin{itemize}
\setlength{\parsep}{0pt}
\setlength{\parskip}{0pt}
\setlength{\leftmargin}{0pt}
\item $N$: the number of classes
\item $AP_i$: AP value of the $i$th class
\end{itemize} \end{minipage}
\\\midrule


Mean Intersection over Union (mIoU) &
\begin{equation}
\mathrm{mIoU}=\frac{1}{N+1} \sum_{i=0}^N \frac{\mathrm{TP}}{\mathrm{FN + FP + TP}}
\end{equation}
&
{\vspace{0.05cm}}\hspace{1cm}
\begin{minipage}[t]{0.55\textwidth} 
IoU is the overlap between the predicted value and the ground truth divided by the area of union.
Then, mIoU is the average value of IoU over all classes.
\end{minipage}
\\\midrule

{\vspace{0.4cm}}
nuScenes Detection Score (NDS) \cite{caesar2020nuscenes}
& 
\begin{equation}
\mathrm{NDS}=\frac{1}{10}\left[5 \mathrm{mAP}+\sum_{\mathrm{mTP} \in \mathbb{T P}}(1-\min (1, \mathrm{mTP}))\right]
\end{equation}
\begin{equation}
  \mathrm{mTP}=\frac{1}{|\mathbb{C}|} \sum_{c \in \mathbb{C}} \mathrm{TP}_{c}
\end{equation}
& 
{\vspace{0.1cm}} \hspace{1cm}
\begin{minipage}[t]{0.55\textwidth}
NDS is a weighted sum of mAP and five TP metrics.
\begin{itemize}
\setlength{\parsep}{0pt}
\setlength{\parskip}{0pt}
\setlength{\leftmargin}{0pt}
  \item mAP: mean Average Precision over all classes
  \item $\mathbb{T P}$: the set of the five mean True Positive metrics, including box location, size, orientation, attributes, and velocity
  \item $\mathrm{mTP}$: the mean True Positive over all classes
  \item $\mathbb{C}$: the set of classes
\end{itemize}
\end{minipage}
\\\midrule

{\vspace{0.2cm}}
Object Location Similarity (OLS) \cite{wang2021rodnet} &
\begin{equation}
\mathrm{OLS}=\exp \left\{\frac{-d^2}{2\left(s \kappa_{cls}\right)^2}\right\}
\end{equation}
&
{\vspace{0.05cm}} \hspace{1cm}
\begin{minipage}[t]{0.55\textwidth} 
OLS describes the correlation between two detections related to distance, classes and scale information.
\begin{itemize}
\setlength{\parsep}{0pt}
\setlength{\parskip}{0pt}
\setlength{\leftmargin}{0pt}
\item $d$: distance between two points in an RA tensor
\item $s$: the distance between the object and the radar sensor, indicating object scale information
\item $\kappa_{cls}$: a constant value that donates the error tolerance for each class $cls$, which can be calculated based on the average object size of that class
\end{itemize} 
\end{minipage}
\\\midrule

{\vspace{0.2cm}}
Average Heading Similarity (AHS) \cite{geiger2012we, ku2018joint} &
\begin{equation}
\mathrm{AHS}=\frac{1}{11} \sum_{r \in\{0,0.1, \ldots, 1\}} \max _{\tilde{r}: \tilde{r} \geq r} \mathrm{s}(\tilde{r})
\end{equation}
\begin{equation}
\mathrm{s}(r)=\frac{1}{|\mathcal{D}(r)|} \sum_{i \in \mathcal{D}(r)} \frac{1+\cos \Delta_\theta^{(i)}}{2} \delta_i
\end{equation}
&
{\vspace{0.05cm}} \hspace{1cm}
\begin{minipage}[t]{0.55\textwidth} 
AHS is the average orientation accuracy in 3D IOU and global orientation angle.
\begin{itemize}
\setlength{\parsep}{0pt}
\setlength{\parskip}{0pt}
\setlength{\leftmargin}{0pt}
\item $r$: recall value
\item $\mathcal{D}(r)$: all object detections at recall rate $r$
\item $\Delta_\theta^{(i)}$: difference in global orientation of detection $i$ as determined by the estimated and ground truth orientation
\item $\delta_i$: whether detection $i$ is assigned to a ground truth bounding box
\end{itemize} 
\end{minipage}
\\

\bottomrule
\end{longtable}

\end{landscape}
\begin{landscape}

\scriptsize
\setlength{\LTcapwidth}{\textwidth} 
\begin{longtable}{p{2.1cm}p{1.3cm}p{2.6cm}p{1.6cm}p{1.8cm}p{1.3cm}p{2.8cm}p{3.6cm}p{3.4cm}}

\caption{\MakeUppercase{performance overview of radar-camera methods}}\label{tab:evaluation}\\

\toprule

\multicolumn{1}{l}{\multirow{2}[2]{*}{\textbf{Method}}} & \multicolumn{1}{l}{\multirow{2}[2]{*}{\textbf{Dataset}}} & \multicolumn{4}{c}{\textbf{Metrics}} & \multicolumn{1}{l}{\multirow{2}[2]{*}{\textbf{Threshold}}} & \multirow{2}[2]{*}{\textbf{Sub-dataset}} & \multirow{2}[2]{*}{\textbf{Hardware}} \\ \cmidrule{3-6}
\multicolumn{1}{c}{} & \multicolumn{1}{c}{} & \textbf{AP/AR} & \textbf{mAP} & \textbf{Others} & \textbf{Inference Time}
\\\midrule

\endfirsthead
\caption{\MakeUppercase{performance overview of radar-camera methods}}\\
\toprule
\multicolumn{1}{l}{\multirow{2}[2]{*}{\textbf{Method}}} & \multicolumn{1}{l}{\multirow{2}[2]{*}{\textbf{Dataset}}} & \multicolumn{4}{c}{\textbf{Metrics}} & \multicolumn{1}{l}{\multirow{2}[2]{*}{\textbf{Threshold}}} & \multirow{2}[2]{*}{\textbf{Sub-dataset}} & \multirow{2}[2]{*}{\textbf{Hardware}} \\ \cmidrule{3-6}
\multicolumn{1}{c}{} & \multicolumn{1}{c}{} & \textbf{AP/AR} & \textbf{mAP} & \textbf{Others} & \textbf{Inference Time}
\\\midrule
\endhead

RRPN \cite{nabati2019rrpn} & nuScenes &
\begin{minipage}[t]{1\textwidth}
AP(NS-F): 43.0 \\ AP$^{.50}$(NS-F): 64.9 \\ AP$^{.75}$(NS-F): 48.5 \\ AP (NS-FB): 35.5 \\ AP$^{.50}$(NS-FB): 59.0 \\ AP$^{.75}$(NS-FB): 37.0 \\ AR(NS-F): 48.6 \\ AR(NS-FB): 42.1 
\end{minipage} \vspace{0.025cm} & - & - & - & IoU=\{0.5, 0.75\} & 
\begin{minipage}[t]{0.16\textwidth}
a) NS-F sub-dataset: from front camera and front radar only, with 23k samples \\b) NS-FB sub-dataset: from the rear camera and two rear radars, with 45k samples
\end{minipage}
 & - \\\midrule

RVNet \cite{john2019rvnet} & nuScenes &
\begin{minipage}[t]{1\textwidth}AP(Cycle): 20.0 \\ AP(Pedestrian): 14.0 \\ AP(Vehicle): 59.0 \\ AP(Obstacle): 26.0\end{minipage} \vspace{0.025cm} & 25.0 & - & 17 ms & IoU=\{0.5\} & Samples from front camera and front radar & One NVIDIA GeForce 1080 GPU\\\midrule

SO-Net \cite{john2020so} & nuScenes &
\begin{minipage}[t]{1\textwidth}
AP(detection): 42.3 \\ AP(segmentation): 99.1\end{minipage} \vspace{0.025cm}& - & - & 25 ms & IoU=\{0.5\} &  Samples under rainy and nighttime conditions (308 pairs for training and 114 pairs for testing) & One NVIDIA GeForce 1080 GPU\\\midrule

SAF-FCOS \cite{chang2020spatial} & nuScenes &
 \begin{minipage}[t]{1\textwidth}AP: 72.4 \\AP$^{.50}$: 90.0 \\ AP$^{.75}$: 79.3  \\ AR: 79.0\end{minipage} \vspace{0.025cm} & - &  & - & IoU=\{0.5, 0.75, 0.5-0.95\} & A total of 34,149 radar-camera pairs & Eight NVIDIA GeForce GTX 1080Ti GPUs\\\midrule

CRF-Net \cite{nobis2019deep} & nuScenes &
- & 55.23 & - & 43 ms & - & Merge the original 23 object classes into seven classes & One NVIDIA Titan XP GPU\\\midrule

BIRANet \cite{yadav2020radar+} & nuScenes &
\begin{minipage}[t]{1\textwidth}AP: 72.3 \\ AP$^{.50}$: 88.9 \\ AP$^{.75}$: 84.3 \\ AP$^{.85}$: 65.7 \\ AR: 75.3\end{minipage} \vspace{0.025cm}& - &- &- & IoU=\{0.5, 0.75, 0.85\} & Merged relevant classes into six classes & One NVIDIA Titan Pascal GPU\\\midrule

Nabati and Qi~\cite{nabati2020radar} & nuScenes &
\begin{minipage}[t]{1\textwidth}AP: 35.6 \\ AP$^{.50}$: 60.53 \\ AP$^{.75}$: 37.38 \\ AR: 42.1\end{minipage} \vspace{0.025cm} & -& MAE: 2.65 &  & IoU=\{0.5, 0.75\} & Samples from front and rear cameras together with all radars & Two NVIDIA Quadro P6000 GPUs\\\midrule

YOdar \cite{kowol2020yodar} & nuScenes &
AP: 43.1 & 39.4 & -& -& - & Samples from nuScenes (29,853 frames for training, 3,289 frames for validation and 1,006 frames for testing) & One NVIDIA Quadro P6000 GPU\\\midrule

CenterFusion \cite{nabati2021centerfusion} & nuScenes &
- & 32.6 & \begin{minipage}[t]{1\textwidth}NDS: 44.9 \\ mATE: 63.1 \\ mASE: 26.1 \\ mAOE: 51.6 \\ mAVE: 61.4 \\ mAAE: 11.5\end{minipage} \vspace{0.035cm} & - & Distance=\{0.5, 1, 2, 4\} & Complete nuScenes & Two NVIDIA P5000 GPUs \\\midrule

Li and Xie~\cite{li2020feature} & nuScenes &
\begin{minipage}[t]{1\textwidth}AP: 24.3 \\ AP$^{.50}$: 48.4 \\ AP$^{.75}$: 22.3 \\ AR: 33.7\end{minipage}  \vspace{0.025cm}& 48.4 & - & - & IoU=\{0.5, 0.75\} & Dataset is randomly divided into a training set, a validation set, and a testing set according to the ratio of 6:2:2 & One NVIDIA GeForce GTX 1080Ti GPU\\\midrule

RVF-Net \cite{nobis2021radar} & nuScenes &
AP: 54.86  & - & - & 44 ms & Distance=\{0.5\} & Samples under rainy and nighttime conditions & One NVIDIA Titan XP GPU\\\midrule

GRIF Net \cite{kim2020grif} & nuScenes &
\begin{minipage}[t]{1\textwidth}AP$^{0.5m}$: 44.1 \\ AP$^{1m}$: 66.5 \\AP$^{2m}$: 71.9 \\ AP$^{4m}$: 74.9\end{minipage} \vspace{0.025cm}& - & - & - & Distance=\{0.5, 1, 2, 4\} & One front camera and 3 front radars & One NVIDIA GeForce GTX 1080Ti GPU\\\midrule

St\"{a}cker~\etal~\cite{stacker2022fusion} & nuScenes &
 - & 36.78 &  & 36.7 ms & IoU=\{0.5\} & Sample from front camera and radar & One NVIDIA GeForce RTX 2080 GPU\\\midrule

FUTR3D \cite{chen2023futr3d} & nuScenes &
- & 39.9 & \begin{minipage}[t]{1\textwidth}NDS: 50.8\end{minipage} \vspace{0.025cm}& - & Distance=\{0.5, 1, 2, 4\} & Complete nuScenes & - \\\midrule

RCBEV \cite{zhou2023bridging} & nuScenes &
-& 40.6 & \begin{minipage}[t]{1\textwidth}NDS: 45.6 \\ mATE: 48.4 \\ mASE: 25.7 \\ mAOE: 58.7 \\ mAVE: 70.2 \\ mAAE: 14.0\end{minipage} \vspace{0.025cm}& - & Distance=\{0.5, 1, 2, 4\} & Complete nuScenes & Four NVIDIA GeForce GTX 3090 GPUs\\\midrule

CRAFT \cite{kim2023craft} & nuScenes &
-& 41.1 & \begin{minipage}[t]{1\textwidth}
NDS: 52.3 \\ mATE: 46.7 \\ mASE: 26.8 \\ mAOE: 45.3 \\ mAVE: 51.9 \\ mAAE: 11.4
\end{minipage} \vspace{0.025cm}& 4.1 FPS & Distance=\{0.5, 1, 2, 4\} & Complete nuScenes & \begin{minipage}[t]{1\textwidth}
Training: four NVIDIA GeForce \\ RTX 3090 GPUs; \\ Testing: one RTX 3090 GPU
\end{minipage}   \\\midrule

MVFusion \cite{wu2023mvfusion} & nuScenes &
- & 45.3 & \begin{minipage}[t]{1\textwidth}NDS: 51.7 \\ mATE: 56.9 \\ mASE: 24.6 \\ mAOE: 37.9 \\ mAVE: 78.1 \\ mAAE: 12.8\end{minipage} \vspace{0.025cm}& - & Distance=\{0.5, 1, 2, 4\} & Complete nuScenes & Eight NVIDIA RTX A6000 GPUs\\\midrule

CRN \cite{kim2023crn} & nuScenes &
- & 57.5 & \begin{minipage}[t]{1\textwidth}NDS: 62.4 \\ mATE: 46.0 \\ mASE: 27.3 \\ mAOE: 44.3 \\ mAVE: 35.2 \\ mAAE: 18.0\end{minipage} \vspace{0.025cm}& 7.2 FPS & Distance=\{0.5, 1, 2, 4\} & Complete nuScenes & -\\\midrule

\multirow{2}[2]{*}{RCFusion \cite{zheng2023rcfusion}} 
& VoD &
- & 49.65 & \begin{minipage}[t]{1\textwidth}-\end{minipage} \vspace{0.025cm}& - & 
IoU=\{0.25, 0.5\} & Complete VoD & \multirow{2}[2]{*}{NVIDIA GeForce RTX 3090 GPUs}\\\cmidrule{2-8}
& TJ4DRadSet &
- & 33.85 & \begin{minipage}[t]{1\textwidth}BEV mAP: 39.76\end{minipage} \vspace{0.025cm}& 10.8 FPS & IoU=\{0.25, 0.5\} & Complete TJ4DRadSet \\\midrule

\multirow{2}[2]{*}{LXL \cite{xiong2023lxl}} 
& VoD &
- & 56.31 & \begin{minipage}[t]{1\textwidth}-\end{minipage} \vspace{0.025cm}& - & IoU=\{0.25, 0.5\} & Complete VoD & \multirow{2}[2]{*}{-} \\\cmidrule{2-8}
& TJ4DRadSet &
- & 36.32 & \begin{minipage}[t]{1\textwidth}BEV mAP: 41.20\end{minipage} \vspace{0.025cm}& - & IoU=\{0.25, 0.5\} & Complete TJ4DRadSet & \\

\bottomrule
\end{longtable}

\end{landscape}

\twocolumn

\end{document}